\documentclass[lettersize,journal]{IEEEtran}
\usepackage{amsmath,amsfonts}
\usepackage{algorithmic}
\usepackage{algorithm}
\usepackage{array}
\usepackage[caption=false,font=normalsize,labelfont=sf,textfont=sf]{subfig}
\usepackage{textcomp}
\usepackage{stfloats}
\usepackage{url}
\usepackage{verbatim}
\usepackage{graphicx}
\usepackage{cite}
\usepackage{amssymb}
\usepackage{bbm}
\usepackage{tabularx,booktabs}
\usepackage{tikz}
\usepackage{xcolor}
\usepackage{titletoc}

\definecolor{darkblue}{rgb}{0.0,0.5,0.5}
\usepackage[
    colorlinks,
    hidelinks,
    bookmarks=false
]{hyperref}
\hypersetup{colorlinks,breaklinks,linkcolor=blue,urlcolor=blue,anchorcolor=blue,citecolor=blue}
\usepackage{booktabs,caption}
\usepackage{multirow}
\usepackage{lscape}
\usepackage{epstopdf}
\usepackage{lineno}
\usepackage{pifont}
\usepackage{multirow}

\newcommand{\mathcalbold}[1]{\boldsymbol{\mathcal{#1}}}

\DeclareMathOperator*{\argmin}{arg\,min\,}

\usepackage{amsthm}
\newtheorem{definition}{Definition}
\newtheorem{remark}{Remark}
\newtheorem{proposition}{Proposition}

\begin{document}

\title{Generalized Least Squares Kernelized Tensor Factorization}

\author{Mengying~Lei,~Lijun~Sun\textsuperscript{$\ast$}
\thanks{Mengying Lei and Lijun Sun are with McGill University, Montreal, Quebec H3A 0C3, Canada. E-mail: mengying.lei@mail.mcgill.ca (Mengying Lei), lijun.sun@mcgill.ca (Lijun Sun).}
\thanks{{$\ast$} Corresponding author.}
}



\maketitle

\begin{abstract}
Recovering incomplete multidimensional tensor-structured data is a fundamental task in many real-world applications. Smoothness-constrained low-rank tensor factorization effectively captures global and long-range correlations, but often struggles to characterize short-scale, high-frequency, or locally varying structures. We propose \emph{GLSKF}, a complementary \underline{G}eneralized \underline{L}east \underline{S}quares \underline{K}ernelized Tensor \underline{F}actorization framework, for multidimensional spatiotemporal data completion. GLSKF additively integrates a covariance-regularized low-rank global component with an explicitly modeled locally correlated residual component under a GLS objective, enabling effective modeling of both global dependencies and localized variations. A \textit{covariance norm} regularizer encodes spatiotemporal dependencies in both components: structured covariances are imposed on the latent factor columns to enforce smoothness in the global factorization, whereas compactly supported sparse kernels are used to model local correlations in the residual. We develop an alternating least squares algorithm with blockwise linear-system updates that exploit the Kronecker structure of the covariance matrices under missing data and facilitate fast conjugate gradient solves. Additional computational gains are obtained by exploiting the sparsity and Toeplitz structure of the local residual covariance matrices for efficient matrix--vector multiplications. We evaluate GLSKF on four real-world multidimensional data-completion tasks: traffic speed imputation, color image completion, digital video recovery, and MRI data reconstruction. Experimental results demonstrate that GLSKF achieves superior reconstruction performance and favorable scalability across a range of tensor completion tasks, supporting its broad applicability to multidimensional data completion.
\end{abstract}


\section{Introduction}
\IEEEPARstart{M}{ultidimensional} tensors provide a natural representation for real-world data with inherent spatial and temporal structures, such as traffic state measurements (speed or flow), color images, digital videos, functional MRI scans, and user-item rating records~\cite{lu2011survey,lei2025low,wangsubspace}. With the rapid advancement of data acquisition technologies, tensor-structured datasets have become increasingly prevalent and play a vital role across a wide range of applications~\cite{sidiropoulos2017tensor,xing2026lightweight,xing2025human}. However, a fundamental challenge in analyzing such data is the ubiquitous sparsity and incompleteness due to sensor failures, occlusions, limited sampling budgets, and privacy or storage constraints~\cite{lei2022bayesian,chen2021low}. As a result, tensor data recovery has emerged as a central research problem in various domains, including traffic state imputation~\cite{chen2021low,lei2022bayesian,chen2024nt}, image/video inpainting~\cite{panagakis2021tensor,ravishankar2019image}, signal processing~\cite{cichocki2015tensor}, and remote sensing~\cite{wang2023tensor}.

The key to effective tensor completion lies in leveraging the intrinsic relationships between observed and missing entries. Real-world multidimensional data often exhibit low-dimensional latent structures, motivating the development of low-rank modeling approaches. Representative methods include matrix/tensor factorization (MF/TF) with a pre-specified rank~\cite{acar2011scalable}, nuclear-norm minimization-based matrix/tensor completion (MC/TC)~\cite{liu2012tensor}, and more expressive tensor formats such as tensor ring and tensor train decompositions~\cite{yuan2019tensor}. Under a low-rank assumption, the missing entries in a $D$-dimensional tensor $\boldsymbol{\mathcal{Y}}\in\mathbb{R}^{I_1\times\cdots\times I_D}$ can be inferred using observed values by projecting the data onto a reduced-dimensional latent space of rank $R$ with $R\ll\left\{I_d\right\}_{d=1}^D$. However, the underlying limitation of standard low-rank models is the permutation invariance along each dimension, which restricts the incorporation of side information or additional consistency on the latent factors~\cite{lei2022bayesian,chen2021low}. In practice, this often leads to suboptimal recovery since real-world data generally exhibit local smoothness, e.g.,~spatial correlations among neighboring sensors or pixels and temporal continuity across adjacent time points.

To overcome the limitations of standard low-rank methods, a large body of work integrates smoothness constraints/regularizations. Examples include total variation (TV) and quadratic variation (QV)-regularized tensor decompositions~\cite{yokota2016smooth}, graph-constrained matrix and tensor learning methods~\cite{rao2015collaborative,bahadori2014fast,yu2016learning,wang2018traffic}, as well as autoregressive (AR)-regularized MF~\cite{yu2016temporal}, TF~\cite{xiong2010temporal}, and TC~\cite{chen2021low}. These models are typically optimized using gradient-based schemes, often with closed-form sub-updates embedded within gradient descent (GD) loops, resulting in a per-iteration complexity on the order of $\mathcal{O}\left(\sum_{d=1}^D\left(R I_d\right)^2\right)$. More efficient iterative solvers such as conjugate gradient (CG) have also been explored~\cite{rao2015collaborative} and often converge in fewer iterations.

Despite the effectiveness in modeling global and low-frequency/long-range structures, smoothness-constrained low-rank approaches often struggle to capture high-frequency, locally varying dependencies~\cite{sang2012full}. The residuals of low-rank approximations commonly exhibit short-scale, structured correlations that reflect rapid variations. Examples include sudden drops in traffic speed due to incidents and sharp edges in digital images. Accurately capturing such local variations using a single low-rank component model generally requires a substantial increase in rank $R$, which can be computationally prohibitive~\cite{yokota2016smooth}, as the per-iteration costs in most low-rank optimizers scale with $R$. Addressing this challenge demands frameworks that can jointly account for both global and local structures in a computationally efficient manner.

\begin{figure*}[!t]
\centering
\includegraphics[width=0.8\textwidth]{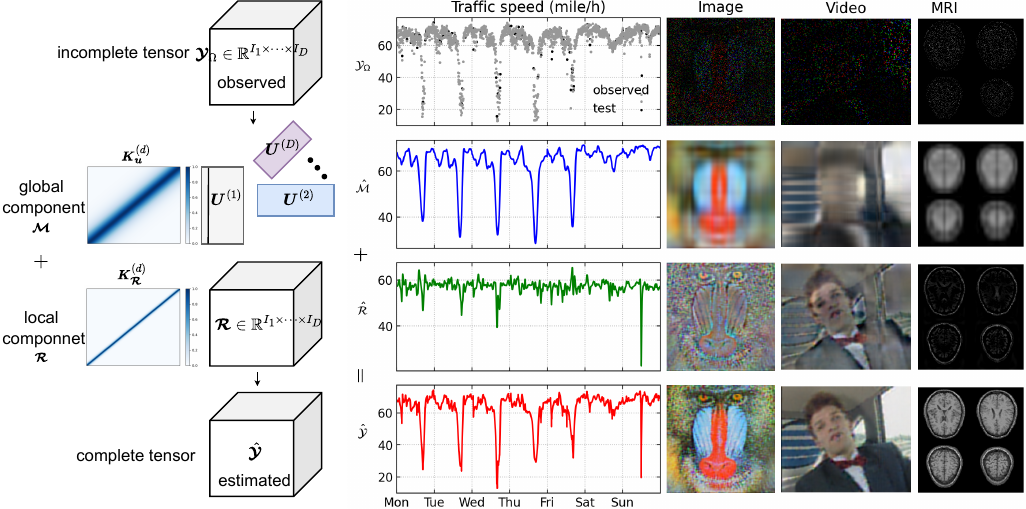}
\caption{Overview of GLSKF. Representative visualizations are shown for traffic speed imputation, image inpainting, video completion, and MRI reconstruction under 90\% random missingness. Rows correspond to the observed incomplete data $\mathcalbold{{Y}}_{\Omega}$, estimated global component $\hat{\mathcalbold{{M}}}$, estimated local residual component $\hat{\mathcalbold{{R}}}$, and final reconstruction $\hat{\mathcalbold{{Y}}}$.}
\label{fig:glskf}
\end{figure*}

In this paper, we propose a complementary \textbf{\underline{G}eneralized \underline{L}east \underline{S}quares \underline{K}ernelized Tensor \underline{F}actorization (GLSKF)} framework for recovering incomplete multidimensional tensor data. For a $D$th-order tensor $\boldsymbol{\mathcal{{Y}}}$, GLSKF additively integrates two complementary components through a GLS (generalized least squares) loss: {(i)} a {global} low-rank, covariance-constrained CANDECOMP/PARAFAC (CP) factorization $\mathcalbold{{M}}=\sum_{r=1}^R\boldsymbol{u}_r^{(1)}\circ\boldsymbol{u}_r^{(2)}\circ\cdots\circ\boldsymbol{u}_r^{(D)}$, modeling long-range, low-frequency structures; and {(ii)} a locally correlated residual $\mathcalbold{{R}}$, designed to capture short-scale, high-frequency variations. The overall model is $\mathcalbold{{Y}}\approx\mathcalbold{{M}}+\mathcalbold{{R}}$ (see Figure~\ref{fig:glskf} for illustration). To enforce smoothness and correlation structure, we introduce \emph{covariance norm} regularization, which admits a probabilistic interpretation as placing Gaussian process (GP) priors on the corresponding model components (cf. Remark~\ref{remark:GLSKF}). We impose structured covariances on the latent factors and use a product kernel with banded, sparse covariances for the residual. To handle missing data efficiently, we design projection operators that {preserve the Kronecker structure} of the covariances, allowing for fast matrix--vector multiplications (MVMs) within CG-based updates. By exploiting Kronecker MVMs together with sparse and Toeplitz computations for residual covariances, GLSKF enables matrix-free CG-based updates for which the dominant operations scale favorably with the number of observed entries and the total tensor size, making it suitable for large-scale applications. Moreover, explicitly modeling local variations through the residual component reduces the rank required for global factorization, further improving efficiency.
The main contributions of this work are summarized below:
\begin{itemize}
\item[(1)] We propose a complementary completion model that combines a covariance-constrained low-rank global component with a locally correlated residual, jointly modeling long-range/low-frequency dependencies and short-scale/high-frequency variations. We introduce a generalized {covariance norm} regularizer to constrain both components in a unified manner.

\item[(2)] We develop an efficient alternating least squares (ALS) algorithm with block-wise linear-system updates. By using selection/projection operators that maintain the Kronecker covariance structure under missingness, the resulting subproblems can be solved efficiently via CG with fast Kronecker MVMs, together with sparse and Toeplitz computations for the local module. The overall framework is computationally scalable with respect to the problem size and can be further accelerated on graphics processing units (GPUs).

\item[(3)] We conduct extensive experiments on four real-world multidimensional datasets, including traffic imputation, color image inpainting, color video completion, and MRI reconstruction. GLSKF outperforms state-of-the-art low-rank baselines in both recovery accuracy and computational efficiency.
\end{itemize}

The remainder of this paper is organized as follows. Section~\ref{sec:RelatedWork} surveys related work on kernelized tensor modeling and additive frameworks. Section~\ref{sec:Preliminaries} introduces notation and background on constrained factorization models. Section~\ref{sec:Methodology} describes the GLSKF methodology, including the global-local additive structure and scalable optimization algorithms. Section~\ref{sec:Experiments} presents experimental results on four real-world tensor completion tasks. Section~\ref{sec:Discussion} discusses component roles and model scalability, and Section~\ref{sec:Conclusion} concludes the paper with a summary and directions for future research.

\section{Related work} \label{sec:RelatedWork}
\paragraph*{Kernel-based methods}
The proposed GLSKF framework models correlations in the latent space by constraining the factor columns through covariance norms: covariance-induced quadratic norms defined by positive definite covariance matrices. For positive semidefinite kernels, a nugget term or the Moore–Penrose inverse is used, yielding a seminorm/generalized covariance norm. This regularization, which constitutes the ``kernelized'' aspect of GLSKF, can be equivalently interpreted as imposing GP priors on the latent factors in a probabilistic formulation. Kernelized or covariance-regularized low-rank modeling has been widely studied for spatiotemporal data in both probabilistic frameworks~\cite{zhou2012kernelized,lei2022bayesian,lei2024scalable} and optimization-based formulations. For example,~\cite{allen2014generalized} introduced covariance-based quadratic norms to encode structural information in generalized principal component analysis and matrix decomposition, while~\cite{larsen2024tensor} employed kernel-weighted norms to incorporate prior smoothness in tensor decomposition.

Recent studies have further extended kernel-based learning to structured tensor representations beyond conventional low-rank factorization~\cite{wu2024low,he2014dusk}. For example,~\cite{li2026support} proposed a support tensor ring kernel machine with dual-stage acceleration for high-dimensional tensor classification, and~\cite{wangsubspace} developed a subspace kernel learning framework for discriminative learning of tensor sequences. These studies highlight the broad utility of kernel methods for capturing tensor-mode dependencies. Nevertheless, their objectives mainly address supervised tensor learning, tensor classification, or discriminative representation learning. By contrast, GLSKF is developed for missing-data recovery in incomplete multidimensional arrays, where kernelized low-rank factors are coupled with a locally correlated residual module and estimated through a GLS formulation. This global-local construction allows GLSKF to exploit both smooth latent structures and residual covariance patterns, which is essential for heterogeneous data such as traffic, images, videos, and MRI tensors.

\paragraph*{Global-local frameworks}
A key feature of GLSKF is its additive integration of two complementary components: a global kernelized low-rank factorization and a locally correlated residual. The idea of decomposing data variation into global and local components has been studied in several related contexts. In large-scale spatial modeling, for example,~\cite{sang2012full,katzfuss2013bayesian} combined reduced-rank covariance approximations with tapered, compactly supported covariance functions to capture both long-range and short-range spatial dependencies. Similarly, in functional data analysis,~\cite{descary2019functional,masak2022random} incorporated banded covariance structures into matrix completion (MC) or process estimation to model local dependence. These methods provide important precedents for global-local covariance modeling, but they are not primarily designed for high-dimensional tensor completion with complex missing patterns. In particular, missing entries may disrupt the Kronecker or structured covariance operations that are typically exploited for scalable computation.

More recently, additive low-rank models have been combined with plug-and-play (PnP) neural modules to encode local priors, such as the global-local-nonlocal (GLON) model~\cite{zhao2021tensor}. While effective, these neural network-based components often lack the interpretability and structural transparency offered by covariance-based modeling. In contrast, GLSKF employs a {GLS} loss to model correlated residuals within a kernel-smoothed low-rank framework. A related use of GLS to impose correlations on model errors was proposed in~\cite{allen2014generalized}. However, their method integrates out the local component and applies constraints solely to the global low-rank space, which limits the ability to handle missing data efficiently.

One work closely related to GLSKF is the Bayesian complementary kernelized low-rank (BCKL) model~\cite{lei2022bckl}, which also adopts an additive global-local structure. BCKL, however, is developed under a fully Bayesian framework using Markov chain Monte Carlo (MCMC) inference, which typically requires thousands of samples to approximate the posterior distributions of latent factors and covariance hyperparameters. In contrast, GLSKF focuses on scalable deterministic estimation. It combines kernelized low-rank modeling with an explicit locally correlated residual component under a GLS optimization framework, leading to efficient block-wise updates and fast point estimation. To clarify the relation between GLSKF and existing global-local modeling strategies, Table~\ref{tab:ref} compares computational and structural properties of several representative approaches. A more detailed summary and comparison is provided in Supplementary Table~\ref{tab:ref-sup}.

\begin{table*}[!t]
\caption{Comparison of global-local modeling frameworks.}
\begin{tabular}{p{140pt}p{60pt}p{60pt}p{60pt}p{60pt}p{61pt}}
\toprule
Method & Local residual explicitly recovered & Supports tensor missing data & Covariance-based local model & Mode-wise kernel regularization & Scalable observed-entry updates \\
\midrule
Spatial full-scale approximation \cite{sang2012full,katzfuss2013bayesian} & \checkmark & \ding{55} & \checkmark & \ding{55} & \ding{55} \\
Functional MC \cite{descary2019functional,masak2022random} & Partially & Limited & \checkmark & \ding{55} & \ding{55} \\
Generalized low-rank model \cite{allen2014generalized} & \ding{55} & Partially & \checkmark & \checkmark & \ding{55} \\
GLON \cite{zhao2021tensor} & \checkmark & \checkmark & \ding{55} & \ding{55} & \ding{55} \\
BCKL \cite{lei2022bckl} & \checkmark & \checkmark & \checkmark & \checkmark & \ding{55} \\
GLSKF & \checkmark & \checkmark & \checkmark & \checkmark & \checkmark \\
\bottomrule
\end{tabular}
\label{tab:ref}
\end{table*}

\section{Preliminaries} \label{sec:Preliminaries}
\subsection{Notation}
Throughout this paper, we use lowercase letters to denote scalars (e.g.,~$x$), boldface lowercase letters to denote vectors (e.g.,~$\boldsymbol{x}\in\mathbb{R}^{M}$), and boldface uppercase letters to denote matrices (e.g.,~$\boldsymbol{X}\in\mathbb{R}^{M\times T}$). The $\ell_2$-norm of a vector $\boldsymbol{x}$ is defined as $\| \boldsymbol{x}\| _2 = \sqrt{\sum_{m} x_{m}^2}$. For a matrix $\boldsymbol{X}\in\mathbb{R}^{M\times T}$, the $(m,t)$th entry is denoted by $x_{mt}$. We denote the $N\times N$ identity matrix by $\boldsymbol{I}_N$. Given two matrices $\boldsymbol{A}\in\mathbb{R}^{M\times T}$ and $\boldsymbol{B}\in\mathbb{R}^{P\times Q}$, the {Kronecker product} is defined as $\boldsymbol{A}\otimes\boldsymbol{B} =
\begin{bmatrix}
  a_{11}\boldsymbol{B} & \cdots & a_{1T}\boldsymbol{B} \\
  \vdots & \ddots & \vdots \\
  a_{M1}\boldsymbol{B} & \cdots & a_{MT}\boldsymbol{B}
\end{bmatrix}
\in\mathbb{R}^{MP\times TQ}$. The {Hadamard product} (i.e.,~element-wise product) for two matrices $\boldsymbol{A}$ and $\boldsymbol{B}$ of the same size is denoted by $\boldsymbol{A}\circledast\boldsymbol{B}$, and the {outer product} of two vectors $\boldsymbol{a}$ and $\boldsymbol{b}$ is denoted by $\boldsymbol{a}\circ\boldsymbol{b}$. If $\boldsymbol{A} =\left[\boldsymbol{a}_{1},\dots,\boldsymbol{a}_{T}\right]\in\mathbb{R}^{M\times T}$ and $\boldsymbol{B} = \left[\boldsymbol{b}_{1},\dots,\boldsymbol{b}_{T}\right]\in\mathbb{R}^{P\times T}$ have the same number of columns, the {Khatri--Rao product} is defined as the column-wise Kronecker product: $\boldsymbol{A}\odot\boldsymbol{B} = [\boldsymbol{a}_{1}\otimes\boldsymbol{b}_{1},\dots,\boldsymbol{a}_{T}\otimes\boldsymbol{b}_{T}]\in\mathbb{R}^{MP\times T}$. The {vectorization} operator on $\boldsymbol{X}$, denoted by $\operatorname{vec}(\boldsymbol{X})$, stacks the columns of $\boldsymbol{X}$ into a single vector. We denote a $D$th-order tensor by $\mathcalbold{{X}}\in\mathbb{R}^{I_1\times\cdots\times I_D}$, and its $(i_1,\dots,i_D)$th entry by $x_{i_1\dots i_D}$. The mode-$d$ unfolding of $\mathcalbold{{X}}$ with $d=1,\dots,D$ is denoted by $\boldsymbol{X}_{(d)}$, which maps the tensor into a matrix along the $d$th mode. We define the vectorization of $\mathcalbold{{X}}$ as $\operatorname{vec}\left(\mathcalbold{{X}}\right)=\operatorname{vec}\left(\boldsymbol{X}_{(1)}\right)$. The projection operator $\mathcal{P}_{\Omega}:\mathbb{R}^{I_1\times\dots\times I_D}\rightarrow\mathbb{R}^{I_1\times\cdots\times I_D}$ maps a tensor onto the index set $\Omega$: $\left[\mathcal{P}_{\Omega}\left(\mathcalbold{{X}}\right)\right]_{i_1\dots i_D}=x_{i_1\dots i_D}$ if $(i_1,\ldots,i_D) \in \Omega$ and 0 otherwise. For simplicity, we alternatively write $\mathcalbold{{X}}_{\Omega}:= \mathcal{P}_{\Omega}\left(\mathcalbold{{X}}\right)$.

\subsection{Smoothness Constrained Low-rank Factorization}
Consider a spatiotemporal data matrix $\boldsymbol{Y}\in\mathbb{R}^{M \times T}$, e.g.,~measurements observed at $M$ spatial locations over $T$ time points. The standard MF approximates $\boldsymbol{Y}$ as $\boldsymbol{Y} \approx \boldsymbol{U}\boldsymbol{V}^{\top}$, where $\boldsymbol{U} \in \mathbb{R}^{M \times R}$ and $\boldsymbol{V} \in \mathbb{R}^{T \times R}$ are latent factor matrices, and $R \ll \min\{M, T\}$ represents a pre-specified rank. In graph-regularized MF~\cite{rao2015collaborative}, the optimization problem is formulated as:
\begin{equation} \label{Eq:GRALS}
\begin{aligned}
\min_{\boldsymbol{U},\boldsymbol{V}}\, &\frac{1}{2}\left\| \boldsymbol{Y}_{\Omega} - \left(\boldsymbol{U}\boldsymbol{V}^{\top}\right)_{\Omega}\right\| _F^2  \\
&+ \frac{1}{2}\left(\operatorname{tr}\left(\boldsymbol{U}^{\top}\boldsymbol{L}_{U}\boldsymbol{U}\right) + \operatorname{tr}\left(\boldsymbol{V}^{\top}\boldsymbol{L}_{V}\boldsymbol{V}\right)\right),
\end{aligned}
\end{equation}
where $\Omega$ denotes the set of observed entries. The Laplacian-based smoothness constraints are defined as $\boldsymbol{L}_U = \lambda_u\boldsymbol{Lap}_{\text{s}} + \eta_u\boldsymbol{I}_{M},~\boldsymbol{L}_V = \lambda_v\boldsymbol{Lap}_{\text{t}} + \eta_v\boldsymbol{I}_T$, with $\{\lambda_u, \eta_u, \lambda_v, \eta_v\}$ being regularization weights and $\bigl\{\boldsymbol{Lap}_{\text{s}}\in\mathbb{R}^{M\times M}, \boldsymbol{Lap}_{\text{t}}\in\mathbb{R}^{T\times T}\bigr\}$ denoting Laplacian matrices for the respective dimensions.

To capture sequential dependencies along the temporal mode, temporal autoregressive (AR) regularization is often applied~\cite{chen2021low}. Let $\boldsymbol{V}\in\mathbb{R}^{T\times R}$ denote the latent temporal factor matrix with $\{\boldsymbol{v}_t\in\mathbb{R}^R\}_{t=1}^T$ being the transposes of the rows. The temporal AR-regularized MF (TRMF)~\cite{yu2016temporal} assumes an order-$d$ AR model on $\boldsymbol{v}_t$ with dynamics $\boldsymbol{v}_{t} \approx \sum_{k=1}^d \boldsymbol{A}_{\ell_k} \boldsymbol{v}_{t-\ell_k}$, where $\mathcal{L}=\{\ell_k\}_{k=1}^d$ is the set of time lags and $\{\boldsymbol{A}_{\ell_k} \in \mathbb{R}^{R \times R} : \ell_k \in \mathcal{L}\}$ are coefficient matrices. The corresponding optimization problem is:
\begin{equation}
\begin{aligned}
\min_{\boldsymbol{U},\boldsymbol{V}}  \,  &\frac{1}{2}\left\| \boldsymbol{Y}_{\Omega} - \left(\boldsymbol{U}\boldsymbol{V}^{\top}\right)_{\Omega}\right\| _F^2 + \frac{\eta_v}{2}\left\| \boldsymbol{V}\right\| _F^2 + \frac{\eta_u}{2}\left\| \boldsymbol{U}\right\|_F^2 \\
& + \frac{\lambda_v}{2}\sum_{t=1+\max(\mathcal{L})}^{T}\left\| \boldsymbol{v}_{t} - \sum_{k=1}^d \boldsymbol{A}_{\ell_k}\boldsymbol{v}_{t-\ell_k}\right\| _2^2,
\end{aligned}
\end{equation}
where $\{\lambda_v, \eta_v, \eta_u\}$ are regularization parameters.

In higher-order tensor settings, TV (total variation) or QV (quadratic variation) regularization is widely applied~\cite{yokota2016smooth}. For an incomplete $D$th-order tensor $\mathcalbold{{Y}} \in \mathbb{R}^{I_1\times\cdots\times I_D}$, the standard CANDECOMP/PARAFAC (CP) decomposition~\cite{kolda2009tensor} approximates it as $\mathcalbold{{Y}} \approx \sum_{r=1}^R \lambda_r\, \boldsymbol{u}_r^{(1)} \circ \cdots \circ \boldsymbol{u}_r^{(D)}$, where $R$ is the CP rank, $\{\lambda_r\}_{r=1}^R$ are scalar weights, and $\bigl\{\boldsymbol{u}_r^{(d)} \in\mathbb{R}^{I_d}\bigr\}_{d=1}^D$ are mode-$d$ factor vectors. The TV/QV-constrained CP decomposition~\cite{yokota2016smooth} is formulated as:
\begin{equation} \label{Eq:SPCobj}
\begin{aligned}
\min_{\left\{\lambda_r\right\}_{r=1}^R,\left\{\boldsymbol{U}^{(d)}\right\}_{d=1}^D}\, &\frac{1}{2}\left\| \mathcalbold{{Y}}_{\Omega} - \mathcalbold{{M}}_{\Omega}\right\| _F^2 \\
&+ \sum_{r=1}^R \frac{\lambda_r^2}{2}\sum_{d=1}^D \rho_d \left\| \boldsymbol{L}^{(d)} \boldsymbol{u}_r^{(d)}\right\| _p^p,
\end{aligned}
\end{equation}
where $\mathcalbold{{M}} = \sum_{r=1}^R \lambda_r\, \boldsymbol{u}_r^{(1)} \circ \cdots \circ \boldsymbol{u}_r^{(D)}$, $\boldsymbol{U}^{(d)}=\left[\boldsymbol{u}_1^{(d)},\ldots,\boldsymbol{u}_{R}^{(d)}\right]\in\mathbb{R}^{I_d\times R}$ denotes the mode-$d$ factor matrix, $\boldsymbol{L}^{(d)} \in \mathbb{R}^{(I_d-1)\times I_d}$ is the first-order difference operator defined as $\boldsymbol{L}^{(d)}=\begin{bmatrix}
1 & -1 &  &  \\
      & 1  & -1 &  \\
      &  & \ddots & \ddots \\
      &  &  & 1 & -1
\end{bmatrix}$, $\rho_d$ controls the smoothness along mode $d$, and $p \in \{1,2\}$ selects TV ($p=1$) or QV ($p=2$) regularization.

While these smoothness-constrained methods effectively model low-frequency, long-range dependencies, they typically assume that the residual from the low-rank approximation is uncorrelated white noise. In practice, however, residuals often exhibit structured local correlations, such as short-term/abrupt temporal dependence in traffic fluctuations and high-frequency edge information in images. Ignoring these correlations can compromise reconstruction accuracy, especially when short-scale, high-frequency variations are significant.

\section{Methodology} \label{sec:Methodology}
In this section, we introduce GLSKF for multidimensional data completion. Consider an incomplete $D$-dimensional data tensor $\mathcalbold{{Y}} \in \mathbb{R}^{I_1 \times \cdots \times I_D}$, where $\Omega$ denotes the set of observed indices and $\Omega^{c}$ its complement. GLSKF models the data as the sum of two complementary components:
\begin{equation*}
\mathcalbold{{Y}} \approx \mathcalbold{{M}} + \mathcalbold{{R}},
\end{equation*}
where $\mathcalbold{{M}}$ captures global structure via a covariance-regularized CP decomposition:
\begin{equation} \label{Eq:CPdecomM}
\mathcalbold{{M}} = \sum_{r=1}^{R} \boldsymbol{u}_r^{(1)} \circ \boldsymbol{u}_r^{(2)} \circ \cdots \circ \boldsymbol{u}_r^{(D)},
\end{equation}
with $R$ being a pre-specified rank and $\boldsymbol{u}_r^{(d)} \in \mathbb{R}^{I_d}$ denoting the $r$th column of the latent factor matrix for mode $d$; $\mathcalbold{{R}}$ models locally correlated variations through a sparse compactly-supported covariance-regularized tensor. Both $\mathcalbold{{M}}$ and $\mathcalbold{{R}}$ are constrained using \textit{covariance norm} regularization (Definition~\ref{def:covnorm}), which encodes smoothness via covariance matrices. The covariance norm regularizer can be interpreted as Mahalanobis-distance-based penalty and provides a unified view of quadratic smoothness constraints (see Remark~\ref{remark:covnorm}; proof is given in Supplementary~\ref{appendix-proofRemark1}). In particular, if $\boldsymbol{K}^{-1}=\boldsymbol{L}^{\top}\boldsymbol{L}$ for a linear operator $\boldsymbol{L}$, then the covariance norm on $\boldsymbol{x}$ can be written as $\left\| \boldsymbol{x}\right\| _{\boldsymbol{K}}^2=\boldsymbol{x}^{\top}\boldsymbol{K}^{-1}\boldsymbol{x}=\boldsymbol{x}^{\top}\boldsymbol{L}^{\top}\boldsymbol{L}\boldsymbol{x}=\left\| \boldsymbol{L}\boldsymbol{x}\right\| _2^2$, showing that the QV penalty in Eq.~\eqref{Eq:SPCobj} can be represented as a covariance norm regularization. When $\boldsymbol{K}=\boldsymbol{I}$, the covariance norm reduces to the $\ell_2$ norm for vectors and Frobenius norm for matrices.

\begin{definition}[Covariance Norm] \label{def:covnorm}
Let $\boldsymbol{K}\succ {0}$ denote a symmetric positive definite covariance matrix. {(i) Vector.} For a vector $\boldsymbol{x} \in \mathbb{R}^{M}$ with covariance $\boldsymbol{K}_{\boldsymbol{x}} \in \mathbb{R}^{M\times M}$, the covariance norm is defined as $\left\| \boldsymbol{x}\right\| _{\boldsymbol{K}_{\boldsymbol{x}}}\triangleq\sqrt{\boldsymbol{x}^{\top}\boldsymbol{K}_{\boldsymbol{x}}^{-1}\boldsymbol{x}}$. (ii) Matrix. For a matrix $\boldsymbol{X}\in\mathbb{R}^{M\times T}$ with row and column covariance matrices $\boldsymbol{K}_{1}\in\mathbb{R}^{M\times M}$ and $\boldsymbol{K}_{2}\in\mathbb{R}^{T\times T}$, the covariance norm is defined as $\left\| \boldsymbol{X}\right\| _{\boldsymbol{K}_{1},\boldsymbol{K}_{2}}\triangleq\sqrt{\operatorname{tr}\left(\boldsymbol{K}_{2}^{-1}\boldsymbol{X}^{\top}\boldsymbol{K}_{1}^{-1}\boldsymbol{X}\right)}$. Equivalently, $\left\| \boldsymbol{X}\right\| _{\boldsymbol{K}_{1},\boldsymbol{K}_{2}}^2=\operatorname{vec}\left(\boldsymbol{X}\right)^{\top}\left(\boldsymbol{K}_2^{-1}\otimes\boldsymbol{K}_1^{-1}\right)\operatorname{vec}\left(\boldsymbol{X}\right)$, i.e.,~$\left\| \boldsymbol{X}\right\| _{\boldsymbol{K}_{1},\boldsymbol{K}_{2}}=\left\| \operatorname{vec}\left(\boldsymbol{X}\right)\right\| _{\boldsymbol{K}_2\otimes\boldsymbol{K}_1}$. (iii) Tensor. For a third-order tensor $\mathcalbold{{X}} \in \mathbb{R}^{M\times T\times P}$ with separable covariance $\boldsymbol{K}_{\mathcalbold{{X}}}=\boldsymbol{K}_3\otimes\boldsymbol{K}_2\otimes\boldsymbol{K}_1$, the covariance norm is defined as $\left\| \mathcalbold{{X}}\right\| _{\boldsymbol{K}_1,\boldsymbol{K}_2,\boldsymbol{K}_3}=\sqrt{\operatorname{vec}(\mathcalbold{{X}})^{\top}\Bigl(\boldsymbol{K}_3^{-1}\otimes\boldsymbol{K}_2^{-1}\otimes\boldsymbol{K}_1^{-1}\Bigr)\operatorname{vec}(\mathcalbold{{X}})}$. Equivalently, with mode-1 unfolding $\boldsymbol{X}_{(1)}\in\mathbb{R}^{M\times (TP)}$, $\left\| \mathcalbold{{X}}\right\| _{\boldsymbol{K}_1,\boldsymbol{K}_2,\boldsymbol{K}_3} = \sqrt{\operatorname{tr}\left(\left(\boldsymbol{K}_3^{-1}\otimes\boldsymbol{K}_2^{-1}\right)\boldsymbol{X}_{(1)}^{\top}\boldsymbol{K}_1^{-1}\boldsymbol{X}_{(1)}\right)}$. The definition naturally extends to higher-order tensors. For convenience, we write $\left\| \boldsymbol{X}\right\| _{\boldsymbol{K}_1,\boldsymbol{K}_2}$ as $\left\| \boldsymbol{X}\right\| _{\boldsymbol{K}_{\boldsymbol{X}}}$ with $\boldsymbol{K}_{\boldsymbol{X}}=\boldsymbol{K}_2\otimes\boldsymbol{K}_1$ for matrices, and $\left\| \mathcalbold{{X}}\right\| _{\boldsymbol{K}_1,\boldsymbol{K}_2,\boldsymbol{K}_3}$ as $\left\| \mathcalbold{{X}}\right\| _{\boldsymbol{K}_{\mathcalbold{{X}}}}$ with $\boldsymbol{K}_{\mathcalbold{{X}}}=\boldsymbol{K}_3\otimes\boldsymbol{K}_2\otimes\boldsymbol{K}_1$ for tensors.
\end{definition}

\begin{remark}[Covariance Norm as {a} Unified Quadratic Regularizer] \label{remark:covnorm}
Any quadratic smoothness penalty $\boldsymbol{x}^{\top}\boldsymbol{Q}\boldsymbol{x}$ with $\boldsymbol{Q}\succ 0$ can be written in the form of a covariance norm by choosing $\boldsymbol{K}=\boldsymbol{Q}^{-1}$, since then $\boldsymbol{K}^{-1}=\boldsymbol{Q}$.
\end{remark}

When the covariance structure of the residual is specified, the global component $\mathcalbold{{M}}$ can be estimated in closed form by marginalizing the correlated errors. However, the marginalization requires inverting an $|\Omega| \times |\Omega|$ matrix, which is computationally prohibitive for large-scale problems~\cite{sang2012full}. GLSKF instead learns the local/residual component $\mathcalbold{{R}}$ explicitly, enabling efficient and flexible modeling of both global structure and local dependencies.

\subsection{Optimization Problem} \label{subsec:ObjFun}
The GLSKF optimization problem for recovering an incomplete $D$-dimensional tensor $\mathcalbold{{Y}}\in\mathbb{R}^{I_1\times\cdots\times I_D}$ can be formulated as:
\begin{equation} \label{Eq:objAll}
\begin{aligned}
\min_{\left\{\boldsymbol{U}^{(d)}\right\}_{d=1}^D,\mathcalbold{{R}}} \mathcal F&\bigl(\{\boldsymbol{U}^{(d)}\}_{d=1}^D,\mathcalbold{ R}\bigr) = \frac{1}{2}\| \mathcalbold{{Y}}_{\Omega} - \mathcalbold{{M}}_{\Omega} - \mathcalbold{{R}}_{\Omega}\| _F^2 \\
& \quad + \rho\sum_{d=1}^D \frac{1}{2}{\bigl\| \boldsymbol{U}^{(d)}\bigr\|} _{\boldsymbol{K}_{\boldsymbol{U}}^{(d)}}^2 + \frac{\gamma}{2} \left\| \mathcalbold{{R}}\right\| _{\boldsymbol{K}_{\mathcalbold{{R}}}}^2.
\end{aligned}
\end{equation}
Here, $\boldsymbol{K}_{\boldsymbol{U}}^{(d)}=\boldsymbol{I}_{R}\otimes\boldsymbol{K}_{\boldsymbol{u}}^{(d)}\in\mathbb{R}^{RI_d\times RI_d}$ is a block-diagonal covariance matrix imposed on the vectorized mode-$d$ latent factor matrix $\boldsymbol{U}^{(d)}\in\mathbb{R}^{I_d\times R}$, where $\boldsymbol{K}_{\boldsymbol{u}}^{(d)}\in\mathbb{R}^{I_d\times I_d}$ denotes the covariance matrix for each column of $\boldsymbol{U}^{(d)}$, i.e.,~$\bigl\{\boldsymbol{u}_r^{(d)}\in\mathbb{R}^{I_d}\bigr\}_{r=1}^R$. In addition, $\boldsymbol{K}_{\mathcalbold{{R}}}=\bigotimes_{d=D}^1\boldsymbol{K}_{\mathcalbold{{R}}}^{(d)}\in\mathbb{R}^{N\times N}$ is the covariance matrix placed on $\operatorname{vec}(\mathcalbold{{R}})$, where $\boldsymbol{K}_{\mathcalbold{{R}}}^{(d)}\in\mathbb{R}^{I_d\times I_d}$ is the mode-$d$ covariance matrix and $N=\prod_{d=1}^D I_d$ is the total number of tensor entries. The weight parameters $\rho$ and $\gamma$ control the regularization strengths on the global latent factors and the local residual component, respectively.

The first term in Eq.~\eqref{Eq:objAll} penalizes the reconstruction error on the observed entries, whereas the second and third terms impose covariance norm regularization on the global factors and the local component, respectively. These regularization terms can be interpreted as Gaussian prior penalties: each column $\boldsymbol{u}_r^{(d)}$ is regularized according to $\mathcal{N}(\mathbf{0},\rho^{-1}\boldsymbol{K}_{\boldsymbol{u}}^{(d)})$, while $\operatorname{vec}(\mathcalbold{{R}})$ is regularized according to $\mathcal{N}(\mathbf{0},\gamma^{-1}\boldsymbol{K}_{\mathcalbold{{R}}})$.

The covariance matrices $\bigl\{\boldsymbol{K}_{\boldsymbol{u}}^{(d)}, \boldsymbol{K}_{\mathcalbold{{R}}}^{(d)}\bigr\}_{d=1}^D$ are built from valid kernel functions~\cite{williams2006gaussian}. A widely used example is the squared exponential (SE) kernel $k_{\text{SE}}\left(x_i, x_j; \ell, \sigma\right) = \sigma^2 \exp\left(-\frac{\left\| x_i - x_j\right\| _2^2}{2\ell^2}\right)$, with length-scale $\ell$ and variance $\sigma^2$. The kernel function and hyperparameter settings govern the correlation patterns encoded by the model and the algebraic structure (e.g.,~dimension and sparsity) of the linear systems solved in CG updates, thereby directly affecting computational complexity. Interpreting the kernel matrices as prior covariance structures and $\rho$ and $\gamma$ as precision parameters (cf. {Remark~\ref{remark:GLSKF}}; proof in Supplementary~\ref{appendix-proofRemark2}) turns kernel design and regularization parameter selection into principled modeling choices rather than heuristics. Moreover, under the additive framework, the residual component only needs to capture local dependencies. Accordingly, we construct $\boldsymbol{K}_{\mathcalbold{{R}}}^{(d)}$ as sparse, banded, compactly supported covariance matrices using covariance tapering~\cite{furrer2006covariance,kaufman2008covariance}, and further exploit sparsity and Toeplitz structure for efficient computation while maintaining effectiveness.

The local residual covariance in GLSKF is specified as a separable product of mode-wise compactly supported kernels. This design is computationally attractive because it preserves Kronecker structure and sparsity, enabling efficient MVMs in the CG updates. However, separability is also a modeling assumption. For data with strong cross-mode interactions, such as spatiotemporal coupling in videos or three-dimensional anatomical dependence in MRI, a fully nonseparable covariance may provide a richer representation. The main drawback is computational: a general nonseparable covariance over $N=\prod_{d=1}^D I_d$ entries requires storing and operating on an $N\times N$ matrix, which is prohibitive for large tensors and would eliminate the Kronecker acceleration used in GLSKF. One possible compromise is to use structured nonseparable approximations, such as sums of separable covariances or compactly supported nonseparable kernels, which can improve modeling flexibility while retaining matrix-free computational advantages. In this work, we focus on separable tapered covariance structures as a scalable modeling strategy and leave the development of scalable nonseparable residual covariance models for future work.

\begin{remark}[MAP Equivalence] \label{remark:GLSKF}
Consider the observation model on the observed entries $\Omega$: $\operatorname{vec}\left((\mathcalbold{{Y}}-\mathcalbold{{M}}-\mathcalbold{{R}})_{\Omega}\right)\sim\mathcal{N}\left(\mathbf{0},\boldsymbol{I}_{|\Omega|}\right)$, and Gaussian priors: $\operatorname{vec}\bigl(\boldsymbol{U}^{(d)}\bigr)\sim\mathcal{N}\bigl(\mathbf{0},\frac{1}{\rho}\boldsymbol{K}_{\boldsymbol{U}}^{(d)}\bigr)$, $\operatorname{vec}(\mathcalbold{{R}})\sim\mathcal{N}\bigl(\mathbf{0},\frac{1}{\gamma}\boldsymbol{K}_{\mathcalbold{{R}}}\bigr)$, where $\rho$ and $\gamma$ are precision (inverse-variance) parameters, and the covariance matrices $\{\boldsymbol{K}_{\boldsymbol{U}}^{(d)},\boldsymbol{K}_{\mathcalbold{{R}}}\}$ encode correlation structures. We assume $\{\boldsymbol{K}_{\boldsymbol{U}}^{(d)},\boldsymbol{K}_{\mathcalbold{{R}}}\}$ have unit marginal variances; any variance scale can be absorbed into $\rho$ and $\gamma$. Then the negative log-posterior of $\{\boldsymbol{U}^{(d)}\}_{d=1}^D$ and $\mathcalbold{{R}}$ given $\mathcalbold{{Y}}_{\Omega}$ (up to an additive constant) is exactly the GLSKF objective in Eq.~\eqref{Eq:objAll}. Thus, GLSKF can be interpreted as a maximum a posteriori (MAP) estimator for a complementary CP model regularized by Gaussian process (GP) priors (cf.~\cite{lei2022bckl}).
\end{remark}

To facilitate a compact representation of observed and missing entries, we introduce the following indexing notation. Let $\mathcalbold{{O}} \in \mathbb{R}^{I_1\times\cdots\times I_D}$ be a binary indicator tensor with entries $o_{i_1\dots i_D}=1$ if $(i_1,\dots,i_D)\in\Omega$ and 0 otherwise. Let $\boldsymbol{O}_{(d)} \in \mathbb{R}^{I_d\times \frac{N}{I_d}}$ be the mode-$d$ unfolding of $\mathcalbold{{O}}$. We define binary selection matrices $\boldsymbol{O}_d,\boldsymbol{O}_d' \in \mathbb{R}^{|\Omega| \times N}$ by extracting from $\boldsymbol{I}_N$ the rows indexed by ones in $\operatorname{vec}\left(\boldsymbol{O}_{(d)}\right)$ and $\operatorname{vec}\bigl(\boldsymbol{O}_{(d)}^{\top}\bigr)$, respectively. Then $\boldsymbol{O}_d$ and $\boldsymbol{O}_d^{\top}$ act as slicing and zero-padding operators mapping between the full vector space and the observed entries. Specifically, $\boldsymbol{O}_d\boldsymbol{y}=\boldsymbol{y}_{\Omega}:\mathbb{R}^N\rightarrow \mathbb{R}^{|\Omega|}$ extracts the observed entries, whereas $\boldsymbol{O}_d^\top \boldsymbol{y}_{\Omega}=\tilde{\boldsymbol{y}}: \mathbb{R}^{|\Omega|}\rightarrow\mathbb{R}^N$ pads the observed vector with zeros at missing locations. The matrix $\boldsymbol{O}_d'$ is used analogously when working with vectorized transposed unfoldings.

The projection/selection operators play a key role in preserving computational structure under missing data. Instead of forming dense covariance submatrices on the observed set, GLSKF uses $\boldsymbol{O}_d$, $\boldsymbol{O}_d'$, and the transposes to extract observed entries and zero-pad vectors back into the full tensor space. The covariance matrix--vector products are then performed in the full space, where Kronecker, sparsity, and Toeplitz structures remain exploitable. This design avoids the loss of Kronecker and sparse structure that would result from directly restricting the covariance matrices to irregularly observed entries and enables efficient CG-based updates.

\subsection{Model Solution}
The optimization problem in Eq.~\eqref{Eq:objAll} is solved via an alternating least squares (ALS) procedure. We alternately update the global component $\boldsymbol{\mathcal{M}}$ and the local residual component $\boldsymbol{\mathcal{R}}$, conditioning on the other. To update $\boldsymbol{\mathcal{M}}$, we fix $\boldsymbol{\mathcal{R}}$ and optimize the latent factor matrices $\bigl\{\boldsymbol{U}^{(d)}\bigr\}_{d=1}^D$ one mode at a time. The local objective function for updating the mode-$d$ factor matrix $\boldsymbol{U}^{(d)}$ is:
\begin{equation} \label{Eq:objLocalM}
\begin{aligned}
\mathcal{F}_{\boldsymbol{U}^{(d)}}&\Bigl(\bigl\{\boldsymbol{U}^{(d)}\bigr\}_{d=1}^D,\boldsymbol{\mathcal{R}}\Bigr) \\
&=\quad \frac{1}{2}\bigl\|\boldsymbol{\mathcal{G}}_{\Omega} - \boldsymbol{\mathcal{M}}_{\Omega}\bigr\|_F^2 
+\frac{\rho}{2}\sum_{d=1}^D \bigl\|\boldsymbol{U}^{(d)}\bigr\|_{\boldsymbol{K}_{\boldsymbol{U}}^{(d)}}^2,
\end{aligned}
\end{equation}
where $\boldsymbol{\mathcal{G}}_{\Omega}=\boldsymbol{\mathcal{Y}}_{\Omega}-\boldsymbol{\mathcal{R}}_{\Omega}$.

To update the residual component $\boldsymbol{\mathcal{R}}$, we fix $\boldsymbol{\mathcal{M}}$ and minimize the following local objective:
\begin{equation} \label{Eq:objLocalR}
\mathcal{F}_{\boldsymbol{\mathcal{R}}}\left(\boldsymbol{\mathcal{R}},\boldsymbol{\mathcal{M}}\right)
=\frac{1}{2}\bigl\|\boldsymbol{\mathcal{L}}_{\Omega} - \boldsymbol{\mathcal{R}}_{\Omega}\bigr\|_F^2 
+\frac{\gamma}{2}\left\|\boldsymbol{\mathcal{R}}\right\|_{\boldsymbol{K}_{\boldsymbol{\mathcal{R}}}}^2,
\end{equation}
with $\boldsymbol{\mathcal{L}}_{\Omega}=\boldsymbol{\mathcal{Y}}_{\Omega}-\boldsymbol{\mathcal{M}}_{\Omega}$.

At iteration $k+1$ in ALS, the model parameters are updated sequentially as follows:
\begin{equation} \label{eq:update}
\begin{cases}
\boldsymbol{U}^{(d),(k+1)} &= \argmin_{\boldsymbol{U}^{(d)}}\,\mathcal{F}_{\boldsymbol{U}^{(d)}}\Bigl(\boldsymbol{U}^{(1),(k+1)},\dots, \boldsymbol{U}^{(d-1),(k+1)},\\ &\boldsymbol{U}^{(d)},\boldsymbol{U}^{(d+1),(k)},\dots,\boldsymbol{U}^{(D),(k)},\boldsymbol{\mathcal{R}}^{(k)}\Bigr),\\
\boldsymbol{\mathcal{M}}^{(k+1)} &= \sum_{r=1}^R \boldsymbol{u}_r^{(1),(k+1)} \circ \cdots \circ \boldsymbol{u}_r^{(D),(k+1)},\\
\boldsymbol{\mathcal{R}}^{(k+1)} &= \argmin_{\boldsymbol{\mathcal{R}}}\,\mathcal{F}_{\boldsymbol{\mathcal{R}}}\bigl(\boldsymbol{\mathcal{R}},\boldsymbol{\mathcal{M}}^{(k+1)}\bigr).
\end{cases}
\end{equation}
The complete tensor is estimated as $\boldsymbol{\mathcal{M}} + \boldsymbol{\mathcal{R}}$. Efficient solutions for the subproblems are given below.

\subsubsection{Update Global Latent Factors \texorpdfstring{\(\bigl\{\boldsymbol{U}^{(d)}\bigr\}_{d=1}^D\)}{TEXT}}
For a given mode $d$, we update the factor matrix $\boldsymbol{U}^{(d)}$ by solving
\begin{equation}
\min_{\boldsymbol{U}^{(d)}} \; \frac{1}{2}\bigl\|\boldsymbol{\mathcal{G}}_{\Omega} - \boldsymbol{\mathcal{M}}_{\Omega}\bigr\|_F^2 + \frac{\rho}{2}\bigl\|\boldsymbol{U}^{(d)}\bigr\|_{\boldsymbol{K}_{\boldsymbol{U}}^{(d)}}^2.
\end{equation}

\paragraph{Solving for the transposed mode-$d$ factor matrix $\bigl(\boldsymbol{U}^{(d)}\bigr)^{\top}$}
Using the CP decomposition in Eq.~\eqref{Eq:CPdecomM}, the transposed mode-\(d\) unfolding of $\boldsymbol{\mathcal{M}}$ satisfies
\begin{equation} \label{Eq:Md}
\boldsymbol{M}_{(d)}^{\top}=\Bigl(\boldsymbol{U}^{(D)}\odot\cdots\odot\boldsymbol{U}^{(d+1)}\odot\boldsymbol{U}^{(d-1)}\odot\cdots\odot\boldsymbol{U}^{(1)}\Bigr)\bigl(\boldsymbol{U}^{(d)}\bigr)\!^{\top}.
\end{equation}
Let $\boldsymbol{H}_{\boldsymbol{u}}^{(d)}=\boldsymbol{U}^{(D)}\odot\cdots\odot\boldsymbol{U}^{(d+1)}\odot\boldsymbol{U}^{(d-1)}\odot\cdots\odot\boldsymbol{U}^{(1)}\in\mathbb{R}^{\frac{N}{I_d}\times R}$. Vectorizing and applying the identity $
\operatorname{vec}\left(\boldsymbol{A}\boldsymbol{X}\boldsymbol{B}\right)=\bigl(\boldsymbol{B}^{\top}\otimes\boldsymbol{A}\bigr)\operatorname{vec}\left(\boldsymbol{X}\right)$,
we can express Eq.~\eqref{Eq:Md} as:
\begin{equation}
\begin{aligned}
\operatorname{vec}\bigl(\boldsymbol{M}_{(d)}^{\top}\bigr)
=\bigr(\boldsymbol{I}_{I_d}\otimes\boldsymbol{H}_{\boldsymbol{u}}^{(d)}\bigl)\operatorname{vec}\Bigl(\bigl(\boldsymbol{U}^{(d)}\bigr)\!^{\top}\Bigr). 
\end{aligned}
\end{equation}
Let $\boldsymbol{H}_d=\boldsymbol{I}_{I_d}\otimes\boldsymbol{H}_{\boldsymbol{u}}^{(d)}\in\mathbb{R}^{N\times RI_d}$ and $\boldsymbol{u}^{(d)}=\operatorname{vec}\Bigl(\bigl(\boldsymbol{U}^{(d)}\bigr)\!^{\top}\Bigr)\in\mathbb{R}^{RI_d}$. The subproblem for $\boldsymbol{U}^{(d)}$ becomes:
\begin{equation} \label{Eq:ObjUdtransposeSolve}
\begin{aligned}
\min_{\boldsymbol{u}^{(d)}} \frac{1}{2}\bigl\|\boldsymbol{O}_d'\operatorname{vec}\bigl(\boldsymbol{G}_{(d)}^{\top}\bigr)-\boldsymbol{O}_d'\boldsymbol{H}_{d}\boldsymbol{u}^{(d)}\bigr\|_2^2 \\
+ \frac{\rho}{2}\big(\boldsymbol{u}^{(d)}\big)\!^{\top}\bigl(\boldsymbol{K}_{\boldsymbol{u}}^{(d)}\otimes\boldsymbol{I}_R\bigr)^{-1}\boldsymbol{u}^{(d)},
\end{aligned}
\end{equation}
where $\boldsymbol{G}_{(d)}$ is the mode-$d$ unfolding of $\boldsymbol{\mathcal{G}}$.
Taking the gradient of this objective with respect to $\boldsymbol{u}^{(d)}$ and setting it to zero yields the linear system:
\begin{equation} \label{Eq:Udtranspose}
\begin{aligned}
\boldsymbol{u}^{(d),(k+1)}=\Big(\boldsymbol{H}_{d}^{\top}\boldsymbol{O}_d'^{\top}\boldsymbol{O}_d'\boldsymbol{H}_{d}+\rho\bigl(\boldsymbol{K}_{\boldsymbol{u}}^{(d)}\bigr)^{-1}\otimes\boldsymbol{I}_R\Big)^{-1} \\
\times\boldsymbol{H}_{d}^{\top}\boldsymbol{O}_{d}'^{\top}\boldsymbol{O}_d'\operatorname{vec}\bigl(\boldsymbol{G}_{(d)}^{\top}\bigr),
\end{aligned}
\end{equation}
Define $\boldsymbol{A}_1=\boldsymbol{H}_{d}^{\top}\boldsymbol{O}_d'^{\top}\boldsymbol{O}_d'\boldsymbol{H}_{d}\in\mathbb{R}^{RI_d\times RI_d}$, $\boldsymbol{A}_2=\rho\bigl(\boldsymbol{K}_{\boldsymbol{u}}^{(d)}\bigr)^{-1}\otimes\boldsymbol{I}_R\in\mathbb{R}^{RI_d\times RI_d}$. Then $\boldsymbol{A}_1$ is block-diagonal as $\boldsymbol{O}_d'^{\top}\boldsymbol{O}_d'$ is a sparse diagonal matrix equal to $\operatorname{diag}\bigl(\operatorname{vec}(\boldsymbol{O}_{(d)}^{\top})\bigr)$, and $\boldsymbol{A}_2$ has Kronecker form. Let $\boldsymbol{A}=\boldsymbol{A}_1+\boldsymbol{A}_2$, and $\boldsymbol{b}=\boldsymbol{H}_{d}^{\top}\boldsymbol{O}_{d}'^{\top}\boldsymbol{O}_d'\operatorname{vec}\bigl(\boldsymbol{G}_{(d)}^{\top}\bigr)\in\mathbb{R}^{RI_d}$. Then Eq.~\eqref{Eq:Udtranspose} becomes $\boldsymbol{u}^{(d),(k+1)}=\boldsymbol{A}^{-1}\boldsymbol{b}$.

\paragraph{Computational strategy}
We solve Eq.~\eqref{Eq:Udtranspose} using the CG method, which requires repeated evaluations of the MVMs (matrix-vector multiplications) $\boldsymbol{A}_1 \boldsymbol{x}$ and $\boldsymbol{A}_2 \boldsymbol{x}$ with $\boldsymbol{x}$ being a vector of length $RI_d$. The matrix $\boldsymbol{A}_1$ is block-diagonal and computing $\boldsymbol{A}_1 \boldsymbol{x}$ requires only $\mathcal{O}\bigl(R|\Omega|\bigr)$ operations. For $\boldsymbol{A}_2 = \rho\,\bigl(\boldsymbol{K}_{\boldsymbol{u}}^{(d)}\bigr)^{-1}\otimes\boldsymbol{I}_R$, we leverage Kronecker MVMs to compute $\boldsymbol{A}_2 \boldsymbol{x}$ efficiently with cost $\mathcal{O}\bigl(R\,I_d^2\bigr)$. Thus, the per-iteration cost for the CG solution is $\mathcal{O}\bigl(R|\Omega| + R\,I_d^2\bigr)$. 
In addition, we initialize the CG iterations with the previous iterate $\boldsymbol{u}^{(d),(k)}$ to accelerate convergence. Algorithm~\ref{Alg:CG-Ud} (Supplementary~\ref{appendix-subsec:CG}) details the CG procedure for updating $\boldsymbol{U}^{(d)}$. Overall, the computational cost for updating the latent matrices using CG is 
$\mathcal{O}\left(J_{\text{CG}}^{glo}\bigl(R|\Omega| + R\,I_d^2\bigr)\right)$,
where $J_{\text{CG}}^{glo}\ll R\,I_d$ is the number of CG iterations required for convergence. This cost is much lower than the $\mathcal{O}((R\,I_d)^3)$ cost of directly computing the closed-form solution in Eq.~\eqref{Eq:Udtranspose}.

Moreover, if $\bigl(\boldsymbol{K}_{\boldsymbol{u}}^{(d)}\bigr)^{-1}$ is constructed as a sparse precision matrix, e.g., via a Gaussian Markov random field (GMRF) representation \cite{rue2005gaussian}, the cost of $\boldsymbol{A}_2\boldsymbol{x}$ can be further reduced to $\mathcal{O}\Bigl(R\,\mathrm{nnz}\left(\bigl(\boldsymbol{K}_{\boldsymbol{u}}^{(d)}\bigr)^{-1}\right)\Bigr)$, where $\mathrm{nnz}(\cdot)$ denotes the number of nonzero elements and scales linearly with $I_d$ for Markovian banded precision matrices.

\subsubsection{Update Local Component \texorpdfstring{$\boldsymbol{\mathcal{R}}$}{TEXT}}
We update the local component tensor $\boldsymbol{\mathcal{R}}$ by minimizing $\mathcal{F}_{\boldsymbol{\mathcal{R}}}$ in Eq.~\eqref{Eq:objLocalR}:
\begin{equation}
\min_{\boldsymbol{\mathcal{R}}} \, \frac{1}{2}\left\|\boldsymbol{\mathcal{L}}_{\Omega}-\boldsymbol{\mathcal{R}}_{\Omega}\right\|_F^2 + \frac{\gamma}{2}\left\|\boldsymbol{\mathcal{R}}\right\|_{\boldsymbol{K}_{\boldsymbol{\mathcal{R}}}}^2.
\end{equation}
\paragraph{Primal form}
Define $\boldsymbol{l}_{\Omega}=\boldsymbol{O}_1\operatorname{vec}\left(\boldsymbol{\mathcal{L}}_{\Omega}\right)\in\mathbb{R}^{|\Omega|}$ and $\boldsymbol{r}=\operatorname{vec}\left(\boldsymbol{\mathcal{R}}\right)\in\mathbb{R}^{N}$. The objective can be transformed into a Tikhonov-regularized least squares problem:
\begin{equation} \label{Eq:Rprimal}
\begin{aligned}
\min_{\boldsymbol{r}} \, \frac{1}{2}\left\|\boldsymbol{l}_{\Omega}-\boldsymbol{O}_1\boldsymbol{r}\right\|_2^2 + \frac{\gamma}{2}\left\|\boldsymbol{r}\right\|_{\boldsymbol{K}_{\boldsymbol{r}}}^2,
\end{aligned}
\end{equation}
where $\boldsymbol{K}_{\boldsymbol{r}}=\bigotimes_{d=D}^1\boldsymbol{K}_{\boldsymbol{\mathcal{R}}}^{(d)}=\boldsymbol{K}_{\boldsymbol{\mathcal{R}}}$. Setting the gradient with respect to $\boldsymbol{r}$, $\left(\boldsymbol{O}_1^{\top}\boldsymbol{O}_1+\gamma\boldsymbol{K}_{\boldsymbol{r}}^{-1}\right)\boldsymbol{r}-\boldsymbol{O}_1^{\top}\boldsymbol{l}_{\Omega}$, to $\boldsymbol{0}$, we have the closed-form solution of $\boldsymbol{r}$:
\begin{equation} \label{Eq:rSolutionInv}
\boldsymbol{r}=\left(\boldsymbol{O}_1^{\top}\boldsymbol{O}_1+\gamma\boldsymbol{K}_{\boldsymbol{r}}^{-1}\right)^{-1}\boldsymbol{O}_1^{\top}\boldsymbol{l}_{\Omega}.
\end{equation}
Computing $\boldsymbol{r}$ using Eq.~\eqref{Eq:rSolutionInv} requires solving an $N \times N$ linear system, which can be computationally expensive for large $N$.

\paragraph{Dual form}
To reduce computational cost, we leverage the dual form of Tikhonov regularization by introducing a dual variable $\boldsymbol{z} \in \mathbb{R}^{|\Omega|}$ and setting $\boldsymbol{r} = \boldsymbol{K}_{\boldsymbol{r}}\,\boldsymbol{O}_1^{\top}\,\boldsymbol{z}$. Replacing the decision variable in the primal form (Eq.~\eqref{Eq:Rprimal}) with $\boldsymbol{z}$ leads to the dual problem:
\begin{equation} \label{Eq:zdual}
\min_{\boldsymbol{z}} \, \frac{1}{2}\bigl\|\boldsymbol{l}_{\Omega} - \boldsymbol{O}_1\boldsymbol{K}_{\boldsymbol{r}}\,\boldsymbol{O}_1^{\top}\,\boldsymbol{z}\bigr\|_2^2 + \frac{\gamma}{2} \boldsymbol{z}^{\top} \boldsymbol{O}_1\boldsymbol{K}_{\boldsymbol{r}}\boldsymbol{O}_1^{\top}\boldsymbol{z}.
\end{equation}
The derivative with respect to $\boldsymbol{z}$ is $\boldsymbol{O}_1\boldsymbol{K}_{\boldsymbol{r}}\boldsymbol{O}_1^{\top}\big(\big(\boldsymbol{O}_1\boldsymbol{K}_{\boldsymbol{r}}\boldsymbol{O}_1^{\top}+\gamma \boldsymbol{I}_{|\Omega|}\big)\boldsymbol{z}-\boldsymbol{l}_{\Omega}\big)$. Since $\boldsymbol{K}_{\boldsymbol{r}}\succ 0$ and $\boldsymbol{O}_1$ has full row rank, $\boldsymbol{O}_1\boldsymbol{K}_{\boldsymbol{r}}\boldsymbol{O}_1^{\top}\succ 0$, thus the stationarity condition of Eq.~\eqref{Eq:zdual} is equivalent to $\big(\boldsymbol{O}_1\boldsymbol{K}_{\boldsymbol{r}}\boldsymbol{O}_1^{\top}+\gamma \boldsymbol{I}_{|\Omega|}\big)\boldsymbol{z}=\boldsymbol{l}_{\Omega}$. Setting the derivative to zero, the optimal $\boldsymbol{z}$ is given by
\begin{equation}
\boldsymbol{z} = \Bigl(\boldsymbol{O}_1\boldsymbol{K}_{\boldsymbol{r}}\,\boldsymbol{O}_1^{\top} + \gamma\,\boldsymbol{I}_{|\Omega|}\Bigr)^{-1}\boldsymbol{l}_{\Omega}\in\mathbb{R}^{|\Omega|}.
\end{equation}
The solution for $\boldsymbol{r}$ at iteration $k+1$ then becomes:
\begin{equation} \label{Eq:rSolutionK}
\begin{aligned}
\boldsymbol{r}^{(k+1)}&=\boldsymbol{K}_{\boldsymbol{r}}\boldsymbol{O}_1^{\top}\boldsymbol{z}^{(k+1)} \\
&=\boldsymbol{K}_{\boldsymbol{r}}\boldsymbol{O}_1^{\top}\Bigl(\boldsymbol{O}_1\boldsymbol{K}_{\boldsymbol{r}}\boldsymbol{O}_1^{\top}+\gamma\boldsymbol{I}_{|\Omega|}\Bigr)^{-1}\boldsymbol{l}_{\Omega}^{(k+1)}.
\end{aligned}
\end{equation}

\paragraph{Computational strategy}
Compared with the primal formulation in Eq.~\eqref{Eq:rSolutionInv} which solves an $N\times N$ system, the dual formulation in Eq.~\eqref{Eq:rSolutionK} solves an $|\Omega|\times|\Omega|$ linear system. Since $|\Omega|\leq N$, the dual system is smaller in dimension and is particularly advantageous under high missingness, where $|\Omega|\ll N$. When the sampling rate is high, however, $|\Omega|$ can be close to $N$, and the advantage of the dual formulation becomes less pronounced. In such cases, a matrix-free primal CG solve may also be competitive, especially if a sparse precision representation of $\boldsymbol{K}_{\boldsymbol{\mathcal{R}}}^{-1}$ is available. In our implementation, we use the \textit{dual formulation} as the experiments mainly involve moderate-to-high missingness and the required matrix-vector products with $\boldsymbol{K}_{\boldsymbol{\mathcal{R}}}$ can be efficiently computed by exploiting Kronecker, sparse, and Toeplitz structures.

Let $\boldsymbol{A}_3=\boldsymbol{O}_1\boldsymbol{K}_{\boldsymbol{r}}\boldsymbol{O}_1^{\top}$. We solve for $\boldsymbol{z}$ using the CG method, where the key matrix-vector product $\boldsymbol{A}_3\boldsymbol{x} = \boldsymbol{O}_1\,\boldsymbol{K}_{\boldsymbol{r}}\,\boldsymbol{O}_1^{\top}\boldsymbol{x}$ (with vector $\boldsymbol{x}\in\mathbb{R}^{|\Omega|}$) is computed efficiently by performing a series of MVMs: (i) $\boldsymbol{O}_1^{\top}$ zero-pads $\boldsymbol{x}$ to a length-$N$ vector $\boldsymbol{x}'$; (ii) the multiplication by $\boldsymbol{K}_{\boldsymbol{r}}$ is executed via Kronecker MVM; and (iii) $\boldsymbol{O}_1$ slices the result back to $\mathbb{R}^{|\Omega|}$. Detailed CG algorithm see Algorithm~\ref{Alg:Rupdate} in Supplementary~\ref{appendix-subsec:CG}. For modes corresponding to spatial or temporal dimensions, we set $\boldsymbol{K}_{\boldsymbol{\mathcal{R}}}^{(d)}$ as sparse banded covariances from compactly supported kernel functions and result in a cost $\mathcal{O}\left(\sum_d\mathrm{nnz}\bigl(\boldsymbol{K}_{\boldsymbol{\mathcal{R}}}^{(d)}\bigr) \frac{N}{I_d}\right)$ for Kronecker MVM $\boldsymbol{K}_{\boldsymbol{r}}\boldsymbol{x}'$. The overall cost of solving Eq.~\eqref{Eq:rSolutionK} is $\mathcal{O}\left(J_{\text{CG}}^{loc}\sum_{d=1}^D\!\left(\mathrm{nnz}\bigl(\boldsymbol{K}_{\boldsymbol{\mathcal{R}}}^{(d)}\bigr)\frac{N}{I_d}\right)\right)$.

In addition, when $\boldsymbol{K}_{\boldsymbol{\mathcal{R}}}^{(d)}$ is generated by a stationary kernel on a 1D regular grid, which is often the case for spatial/temporal modes, we further accelerate MVMs by exploiting Toeplitz structure of the covariance matrices via fast Fourier transforms (FFTs) \cite{wilson2015thoughts}, yielding per-mode cost $\mathcal{O}\!\left((2I_d-2)\log(2I_d-2)\right)$. Details are provided in Supplementary~\ref{appendix-subsec:Toeplitz}.

\subsection{Model Implementation}
There are several considerations that make GLSKF stable and fast in practice. First, due to its additive structure, jointly estimating the global component $\boldsymbol{\mathcal{M}}$ and the local residual $\boldsymbol{\mathcal{R}}$ from a random start can be challenging. We initialize $\boldsymbol{\mathcal{R}}$ as an all-zero tensor and update only $\boldsymbol{\mathcal{M}}$, i.e., the latent factors $\bigl\{\boldsymbol{U}^{(d)}\bigr\}_{d=1}^D$, for the first $K_0$ ALS iterations (with $K_0 < K$). After this warm-start, we activate $\boldsymbol{\mathcal{R}}$ and continue alternating updates. This staged strategy ensures the global structure settles before the fine-scale details are fitted, which improves convergence and reduces runtime.
Second, we typically initialize the last-mode residual covariance as the identity matrix, i.e., $\boldsymbol{K}_{\boldsymbol{\mathcal{R}}}^{(D)} = \boldsymbol{I}_{I_D}$, which may not reflect the true residual correlations. After $\boldsymbol{\mathcal{R}}$ has been updated, we compute an empirical $\boldsymbol{K}_{\boldsymbol{\mathcal{R}}}^{(D)}$ from the current estimate of $\boldsymbol{\mathcal{R}}$ (e.g., as the row covariance of the mode-$D$ unfolding), and plug the updated $\boldsymbol{K}_{\boldsymbol{\mathcal{R}}}^{(D)}$ back into subsequent $\boldsymbol{\mathcal{R}}$ updates.

With the estimates $\boldsymbol{\mathcal{M}}^{(k+1)}$ and $\boldsymbol{\mathcal{R}}^{(k+1)}$, the completed estimation tensor $\hat{\boldsymbol{\mathcal{Y}}}$ is updated as:
\begin{equation} \label{Eq:YestUpate}
\hat{\boldsymbol{\mathcal{Y}}}_{\Omega} = \boldsymbol{\mathcal{Y}}_{\Omega}, \quad
\hat{\boldsymbol{\mathcal{Y}}}_{\Omega^c} = \boldsymbol{\mathcal{M}}_{\Omega^c}^{(k+1)} + \boldsymbol{\mathcal{R}}_{\Omega^c}^{(k+1)}.
\end{equation}
We monitor the change in the reconstruction over the observed entries and stop when
\[
\Big\| \big(\boldsymbol{\mathcal{M}}^{(k+1)}+\boldsymbol{\mathcal{R}}^{(k+1)}\big)_{\Omega} - \big(\boldsymbol{\mathcal{M}}^{(k)}+\boldsymbol{\mathcal{R}}^{(k)}\big)_{\Omega} \Big\|_F^2 < \epsilon,
\]
where $\epsilon$ is a predefined threshold. The convergence guarantee is stated in {Proposition~\ref{prop:convergence}} (proof is given in Supplementary~\ref{appendix-proofProposition1}). In practice, we often consider a relative criterion $\frac{\big\| \big(\boldsymbol{\mathcal{M}}^{(k+1)}+\boldsymbol{\mathcal{R}}^{(k+1)}\big)_{\Omega} - \big(\boldsymbol{\mathcal{M}}^{(k)}+\boldsymbol{\mathcal{R}}^{(k)}\big)_{\Omega}\big\|_F}{\big\|\big(\boldsymbol{\mathcal{M}}^{(k)}+\boldsymbol{\mathcal{R}}^{(k)}\big)_{\Omega}\big\|_F}<\epsilon'$ (e.g., $\epsilon'=10^{-6}$), together with a cap on the maximum number of outer ALS iterations. For the inner CG solvers, a modest fixed tolerance (e.g., $10^{-6}$ in Algorithms~\ref{Alg:CG-Ud} and~\ref{Alg:Rupdate}) typically preserves monotone descent while keeping runtime low. The overall GLSKF implementation for multidimensional data recovery is summarized in Algorithm~\ref{Alg:GLSKF}. For the convergence analysis in Proposition~\ref{prop:convergence}, we consider the fixed-covariance GLSKF objective, where $\{\boldsymbol{K}_{\boldsymbol{u}}^{(d)}\}_{d=1}^D$ and $\{\boldsymbol{K}_{\boldsymbol{\mathcal{R}}}^{(d)}\}_{d=1}^{D}$ are fixed during ALS iterations. The empirical update of $\boldsymbol{K}_{\boldsymbol{\mathcal{R}}}^{(D)}$ from the current residual estimate is treated as an optional practical heuristic and is not covered by the formal convergence result. In practice, this update can be performed as an outer-loop refinement, after which the fixed-covariance ALS iterations are restarted.

\begin{proposition}[Monotone Descent of Fixed-covariance GLSKF] \label{prop:convergence}
Assume $\rho, \gamma>0$ and all covariance matrices $\big\{\boldsymbol{K}_{\boldsymbol{u}}^{(d)}\big\}_{d=1}^D$ and $\big\{\boldsymbol{K}_{\boldsymbol{\mathcal{R}}}^{(d)}\big\}_{d=1}^D$ are fixed and symmetric positive definite. At each ALS iteration, every block subproblem (for $\boldsymbol{U}^{(d)}$ or $\boldsymbol{\mathcal{R}}$) is solved exactly or by CG with a residual tolerance ensuring the block objective does not increase. Then the GLSKF objective $\mathcal{F}\big(\big\{\boldsymbol{U}^{(d)}\big\}_{d=1}^D,\boldsymbol{\mathcal{R}}\big)$ is bounded below and non-increasing along the iterates; consequently $\mathcal{F}^{(k)}$ converges.
\end{proposition}

\begin{algorithm}[t!]
\caption{GLSKF for data completion.}
\label{Alg:GLSKF}
\begin{algorithmic}[1]
\renewcommand{\algorithmicrequire}{\textbf{Input:}}
\renewcommand{\algorithmicensure}{\textbf{Output:}}
\REQUIRE $\boldsymbol{\mathcal{Y}}_{\Omega},~\rho,~\gamma,~R$, $K$, $K_0$, $\epsilon$.
\ENSURE Completed tensor $\hat{\boldsymbol{\mathcal{Y}}}$.
\STATE Initialize $\big\{\boldsymbol{U}^{(d)}\big\}_{d=1}^D$ as small random values; initialize $\boldsymbol{\mathcal{R}}$ as a zero tensor.
\FOR{$k=0$ {\bfseries to} $K-1$}
\FOR{$d=1$ {\bfseries to} $D$}
\STATE Update $\big(\boldsymbol{U}^{(d)}\big)^{(k+1)}$ by Eq.~\eqref{Eq:Udtranspose}, Algorithm~\ref{Alg:CG-Ud}; 
\ENDFOR
\STATE Update $\boldsymbol{\mathcal{M}}^{(k+1)}$ by Eq.~\eqref{eq:update};
\IF{$k \geq K_0$}
\STATE Update $\boldsymbol{\mathcal{R}}^{(k+1)}$ by Eq.~\eqref{Eq:rSolutionK}, Algorithm~\ref{Alg:Rupdate};
\STATE Compute $\boldsymbol{K}_{\boldsymbol{\mathcal{R}}}^{(D)}=\operatorname{cov}\left(\boldsymbol{R}_{(D)}^{(k+1)}\right)$, update $\boldsymbol{K}_{\boldsymbol{\mathcal{R}}}$;
\ENDIF
\IF{$\| \big(\boldsymbol{\mathcal{M}}^{(k+1)}+\boldsymbol{\mathcal{R}}^{(k+1)}\big)_{\Omega} - \big(\boldsymbol{\mathcal{M}}^{(k)}+\boldsymbol{\mathcal{R}}^{(k)}\big)_{\Omega} \|_F^2 < \epsilon$}
\STATE {\bfseries{break}};
\ENDIF
\ENDFOR
\STATE {\bfseries{return}} $\hat{\boldsymbol{\mathcal{Y}}}$ by Eq.~\eqref{Eq:YestUpate}.
\end{algorithmic}
\end{algorithm}

\subsection{Computational Complexity} \label{sec:computational}
We compare the theoretical computational complexity of representative low-rank solutions for multidimensional data completion in Table~\ref{tab:computational}. The proposed GLSKF framework fully exploits Kronecker MVMs, sparse matrix computations, and the inherent structure of the data to achieve linear scaling with the number of entries. In addition, by explicitly capturing local correlations through the local component, GLSKF typically requires a significantly smaller CP rank $R$ than single-component low-rank methods, further enhancing computational efficiency. Furthermore, the GPU implementation provides additional practical acceleration; detailed wall-clock runtime comparisons are given in Section~\ref{subsec:time}.

\begin{table*}[!t]
\centering
\caption{Comparison of model complexity.}
\begin{tabular}{lll}
\toprule
Solution & Global cost  & Local cost   \\
\midrule
{GD (e.g., SPC \cite{yokota2016smooth})} & $\mathcal{O}\left(J_{\text{GD}}R\sum_dI_d^2\right)$ & $\mathcal{O}\left(J_{\text{GD}}\left|\Omega\right|^2\right)$  \\
{CG (e.g., GRALS \cite{rao2015collaborative})} & $\mathcal{O}\Big(DJ_{\text{CG}}\Big(R|\Omega|+R\mathrm{nnz}\left(\boldsymbol{L}_U\right)\Big)\Big)$ & $\mathcal{O}\left(J_{\text{CG}}\left|\Omega\right|^2\right)$ \\
{Closed form} & $\mathcal{O}\left(\sum_d(RI_d)^3\right)$ [Eq.~\eqref{Eq:Udtranspose}] & $\mathcal{O}\left(|\Omega|^3\right)$ [Eq.~\eqref{Eq:rSolutionK}] \\
{GLSKF} & $\mathcal{O}\Big(\sum_dJ_{\text{CG}}\big(R|\Omega|+RI_d^2\big)\Big)$ & $\mathcal{O}\big(J_{\text{CG}}\sum_d (2I_d-2)\log(2I_d-2)\big)$ \\
\bottomrule
\multicolumn{3}{l}{$J_{\text{GD}}\gg J_{\text{CG}}$; $\boldsymbol{L}_U$ refers to the Laplacian regularization term in objective~\eqref{Eq:GRALS}.}
\end{tabular}
\label{tab:computational}
\end{table*}

\section{Experiments} \label{sec:Experiments}
We evaluate GLSKF for multivariate/spatiotemporal data completion on four real-world datasets: (i) highway traffic speed imputation, (ii) color image restoration, (iii) fourth-order color video completion, and (iv) MRI image reconstruction. The sampling rate (SR) is defined as $\text{SR}=\frac{|\Omega|}{N}~(N=\prod_{d=1}^DI_d)$. The experimental code is available at \url{https://github.com/MengyingLei/GLSKF}.

For traffic state completion (i), we compare model performance using mean absolute error (MAE) and root mean square error (RMSE) on the test (i.e., missing) entries:
\[
\begin{aligned}
\text{MAE} = \frac{1}{n}\sum_{i=1}^{n}\left|y_i - \hat{y}_i\right|,\qquad \text{RMSE} = \sqrt{\frac{1}{n}\sum_{i=1}^{n}\left(y_i - \hat{y}_i\right)^2},
\end{aligned}
\]
where $n$ is the number of test samples, and $y_i$ and $\hat{y}_i$ are the true and estimated values, respectively.

For {color image recovery} (ii) and {color video completion} (iii), we use peak signal-to-noise ratio (PSNR) and structural similarity (SSIM) as quantitative metrics; higher values indicate better reconstructions. For {MRI completion} (iv), we report PSNR, MAE, and RMSE. For video data, PSNR and SSIM are computed per frame and then averaged across frames; for MRI data, PSNR is calculated per frontal slice and averaged across slices.

\subsection{Traffic Speed Imputation}
\subsubsection{Datasets}
We evaluate GLSKF on two publicly available traffic speed datasets under random missing (RM) and nonrandom fiber missing (NM) scenarios at various SRs (sampling rates). The datasets are as follows:
\begin{itemize}
\item {Seattle Freeway Traffic Speed (S)}\footnote{\url{https://github.com/zhiyongc/Seattle-Loop-Data}}: Traffic speed measurements collected from 323 loop detectors on Seattle highways in 2015, sampled every 5 minutes (288 time points per day). We use data from 30 consecutive days (January 1st to 30th), forming a $\textit{location}\times\textit{time-of-day}\times\textit{day}$ tensor of size $323 \times 288 \times 30$.

\item {PeMS-Bay Traffic Speed (P)}\footnote{\url{https://github.com/liyaguang/DCRNN}}: Traffic speeds from 325 loop detectors across the Bay Area, CA. We use a one-month subset (February 1st to 28th, 2017) from 319 detectors, aggregated to 10-minute intervals (144 time points per day), yielding a $\textit{location}\times\textit{time-of-day}\times\textit{day}$ tensor of size $319 \times 144 \times 28$.
\end{itemize}
For both datasets, we generate RM scenarios by uniformly retaining an $\text{SR}\in\{0.7,\,0.3,\,0.1\}$ fraction of entries, corresponding to 30\%, 70\%, and 90\% random missingness, respectively. For NM scenarios, we consider nonrandom fiber missing along the \textit{time-of-day} mode with $\text{SR}\in\{0.7,\,0.5\}$, corresponding to 30\% and 50\% structured missingness, respectively. In this setting, entire time-of-day fibers are removed to mimic systematic recording or sensor failures.

\subsubsection{Configuration}
\paragraph{Baselines} We compare GLSKF with representative low-rank completion methods that are widely used in the literature: \textbf{HaLRTC} (High accuracy Low-Rank Tensor Completion) \cite{liu2012tensor}: a basic convex tensor completion approach based on nuclear-norm minimization. \textbf{SPC} (Smooth PARAFAC tensor Completion) \cite{yokota2016smooth}: CP factorization with total variation (TV) or quadratic variation (QV) smoothness constraints. \textbf{GRALS} (Graph Regularized Alternating Least Squares) \cite{rao2015collaborative}: Low-rank factorization with graph Laplacian regularization. \textbf{TRMF} (Temporal Regularized Matrix Factorization) \cite{yu2016temporal}: MF with temporal autoregressive (AR) regularization.

To assess the contribution of each component in GLSKF, we additionally compare against the following ablations: \textbf{LRTF} (Low-Rank Tensor Factorization): plain CP tensor factorization without smoothness constraints. The optimization problem is: $\min_{\{\boldsymbol{U}^{(d)}\}_{d=1}^D}\frac{1}{2}\left\|\boldsymbol{\mathcal{Y}}_{\Omega}-\boldsymbol{\mathcal{M}}_{\Omega}\right\|_F^2+\frac{\alpha}{2}\sum_{d=1}^D\big\|\boldsymbol{U}^{(d)}\big\|_F^2$, where $\boldsymbol{\mathcal{M}}=\sum_{r=1}^R\boldsymbol{u}_r^{(1)}\circ\cdots\circ\boldsymbol{u}_r^{(D)}$ and $\alpha$ is a regularization parameter. \textbf{LKTF} (Low-rank Kernelized Tensor Factorization): GLSKF with only the covariance-regularized global component $\boldsymbol{\mathcal{M}}$. The problem becomes: $\min_{\{\boldsymbol{U}^{(d)}\}_{d=1}^D}\frac{1}{2}\left\|\boldsymbol{\mathcal{Y}}_{\Omega}-\boldsymbol{\mathcal{M}}_{\Omega}\right\|_F^2+\frac{\rho}{2}\sum_{d=1}^D\big\|\boldsymbol{U}^{(d)}\big\|_{\boldsymbol{K}_{\boldsymbol{U}}^{(d)}}^2$. \textbf{GLSlocal}: GLSKF with only the local residual component \(\boldsymbol{\mathcal{R}}\) (covariance-constrained GLS; no global factorization). The problem is: $\min_{\boldsymbol{\mathcal{R}}}\frac{1}{2}\left\|\boldsymbol{\mathcal{Y}}_{\Omega}-\boldsymbol{\mathcal{R}}_{\Omega}\right\|_F^2+\frac{\gamma}{2}\|\boldsymbol{\mathcal{R}}\|_{\boldsymbol{K}_{\boldsymbol{\mathcal{R}}}}^2$.

\paragraph{Parameter settings} For both datasets, we set the CP rank to $R=20$ for the global component $\boldsymbol{\mathcal{M}}$ in all experiments, and use the same kernel settings. Let $\Delta_{ij}$ denote the distance between inputs $x_i$ and $x_j$ in the corresponding dimension.

For the {spatial} \textit{location} dimension ($d=1$), we first build a weighted adjacency matrix $\boldsymbol{\Phi}\in\mathbb{R}^{I_1\times I_1}$ from an exponential function: $\boldsymbol{\Phi}_{ij}=k_{\text{exp}}\left(x_i,x_j; \ell\right)=\begin{cases}
    \exp\left(-\frac{\Delta_{ij}}{\ell^2}\right),~i\neq j \\
    0,~i=j
\end{cases}$,
where $\Delta_{ij}$ denotes the road network distance between sensors $i\in\{1,\ldots,I_1\}$ and $j\in\{1,\ldots,I_1\}$, and $\ell$ is considered as a length-scale kernel hyperparameter. Based on $\boldsymbol{\Phi}$, we then form the $I_1\times I_1$ normalized Laplacian matrix $\boldsymbol{\Lambda}=\boldsymbol{I}_{I_1}-\boldsymbol{D}^{-\frac{1}{2}}\boldsymbol{\Phi}\boldsymbol{D}^{-\frac{1}{2}}$, where $\boldsymbol{D}\in\mathbb{R}^{I_1\times I_1}$ is a diagonal matrix with ${d}_{ii}=\sum_{j=1}^{I_1}\boldsymbol{\Phi}_{ij}$ for $i=1,\dots,I_1$, and use a graph regularized Laplacian (RL) kernel $k_{\text{RL}}\left({\Delta}_{ij}; \ell,\sigma\right)$ \cite{smola2003kernels} to construct $\boldsymbol{K}_{\boldsymbol{u}}^{(1)}$ for the spatial latent factors: $\boldsymbol{K}_{\boldsymbol{u}}^{(1)}=\left(\boldsymbol{I}_{I_1}+\sigma^2\boldsymbol{\Lambda}\right)^{-1}$, where $\sigma^2$ is the variance hyperparameter. For CG updates of the mode-1 factor $\boldsymbol{u}^{(1)}$ in Eq.~\eqref{Eq:Udtranspose}, the explicit inversion computation of $\boldsymbol{K}_{\boldsymbol{u}}^{(1)}$ can be avoided by applying $\big(\boldsymbol{K}_{\boldsymbol{u}}^{(1)}\big)^{-1}=\boldsymbol{I}_{I_1}+\sigma^2\boldsymbol{\Lambda}$ directly. For the local residual, we use a tapered RL kernel ($k_{\text{RL}}k_{\text{taper}}$) to compute a sparse banded/compactly-supported covariance $\boldsymbol{K}_{\boldsymbol{\mathcal{R}}}^{(1)}$, where $k_{\text{taper}}$ is the Bohman taper \cite{gneiting2002compactly}: 
\[
\begin{aligned}
k_{\text{taper}}&\left(x_i,x_j;\lambda\right) \\
&=
\begin{cases}
\left(1-\frac{\Delta_{ij}}{\lambda}\right)\cos\left(\pi\frac{\Delta_{ij}}{\lambda}\right)+\frac{1}{\pi}\sin\left(\pi\frac{\Delta_{ij}}{\lambda}\right), & \Delta_{ij}<\lambda \\
0, & \Delta_{ij}\geq\lambda
\end{cases},
\end{aligned}
\]
with $\lambda$ being the taper range.

For the second {temporal} \textit{time-of-day} dimension ($d=2$), we use a Mat\'{e}rn 3/2 kernel for $\boldsymbol{K}_{\boldsymbol{u}}^{(2)}$: $k_{\text{Mat\'ern~3/2}}\left(x_i,x_j;\ell,\sigma\right)=\sigma^2\left(1+\frac{\sqrt{3}\Delta_{ij}}{\ell}\right)\exp\left(-\frac{\sqrt{3}\Delta_{ij}}{\ell}\right)$, and use a tapered Mat\'{e}rn 3/2 kernel ($k_{\text{Mat\'ern~3/2}}k_{\text{taper}}$) with also a Bohman taper to construct $\boldsymbol{K}_{\boldsymbol{\mathcal{R}}}^{(2)}$. For the third {\textit{day}} dimension ($d=3$), we set both $\boldsymbol{K}_{\boldsymbol{u}}^{(3)}$ and $\boldsymbol{K}_{\boldsymbol{\mathcal{R}}}^{(3)}$ as the identity matrix $\boldsymbol{I}_{I_3}$.

For kernel hyperparameters, we fix the variance to $\sigma^2=1$ for all covariances and balance the weights of the global and local regularizers through $\rho$ and $\gamma$. Let $\ell_{\boldsymbol{u}}^{(d)}$ and $\ell_{\boldsymbol{r}}^{(d)}$ denote the length-scales for $\boldsymbol{U}^{(d)}$ and the $d$th mode of $\boldsymbol{\mathcal{R}}$, respectively, and let $\lambda_d$ denote the tapering range for the $d$th mode. In practice, these kernel hyperparameters should reflect the spatial/temporal correlation characteristics of the studied dataset \cite{sang2012full}. We therefore specify the kernel hyperparameters according to prior knowledge of the application and the expected correlation scales along each dimension. We provide an empirical validation by fitting associated empirical correlation profiles computed from the observed entries to support the specification process~\cite{cressie1985fitting,cressie2015statistics}. The empirical results show that the pre-specified values are consistent with the estimated correlation patterns. Details of the kernel hyperparameter selection procedure and the empirical validation are given in Supplementary~\ref{supp:kernelHypeSelection}. We further conduct a sensitivity analysis to evaluate the effect of varying the kernel hyperparameters around the corresponding pre-specified values; see Section~\ref{subsec:sensitivity}. For example, for dataset (S) with a 5-minute temporal resolution, $\ell_{\boldsymbol{u}}^{(2)}=40$ and $\ell_{\boldsymbol{r}}^{(2)}=5$ time points in the temporal global and local covariances $\boldsymbol{K}_{\boldsymbol{u}}^{(2)}$ and $\boldsymbol{K}_{\boldsymbol{\mathcal{R}}}^{(2)}$ correspond to characteristic temporal scales of 200 minutes and 25 minutes, respectively, indicating a long-range temporal dependence over roughly 3.3 hours for the global component and a short-range temporal variation over about 25 minutes for the local residual.

For regularization parameters, we optimally tune $\rho$ and $\gamma$ from $\{1, 5, 10, 15, 20\}$ and $\{1/10, 1/5, 1, 5, 10\}$ respectively based on cross-validation; detailed results see Supplementary~\ref{appendix:regularization}, Figure~\ref{fig:traffic_parameter}. Summary of the selected regularization parameters and the kernel configurations in each scenario for the two traffic datasets are given in Supplementary~\ref{appendix-sec:configuration}, Tables~\ref{tab:seattle-kernel}--\ref{tab:traffic-regularization}.

For baselines, we follow the specific settings as below: \textbf{LRTF}, \textbf{LKTF}, \textbf{GLSlocal}: we use the same rank ($R=20$), covariance settings, and $\{\rho,\gamma\}$ as GLSKF, and additionally tune $\alpha$ from $\{0.1,\,0.2,\,1,\,5,\,10\}$ for LRTF. \textbf{SPC}: we test with both TV and QV constraints and report the better result. \textbf{GRALS}: we employ a kernelized CP implementation, where the spatial and time-of-day dimensions are regularized by graph Laplacian covariances (the spatial Laplacian matches GLSKF, the time-of-day mode uses a first-order adjacency Laplacian, and the day mode is set to identity). \textbf{TRMF}: we reshape the data into a $\textit{location}\times\textit{time points}$ matrix, and set the temporal lags to $\{1,\,2,\,288\}$ for Seattle and $\{1,\,2,\,144\}$ for PeMS-Bay.

\subsubsection{Results}
\begin{figure*}[!t]
\centering
\includegraphics[width=1\textwidth]{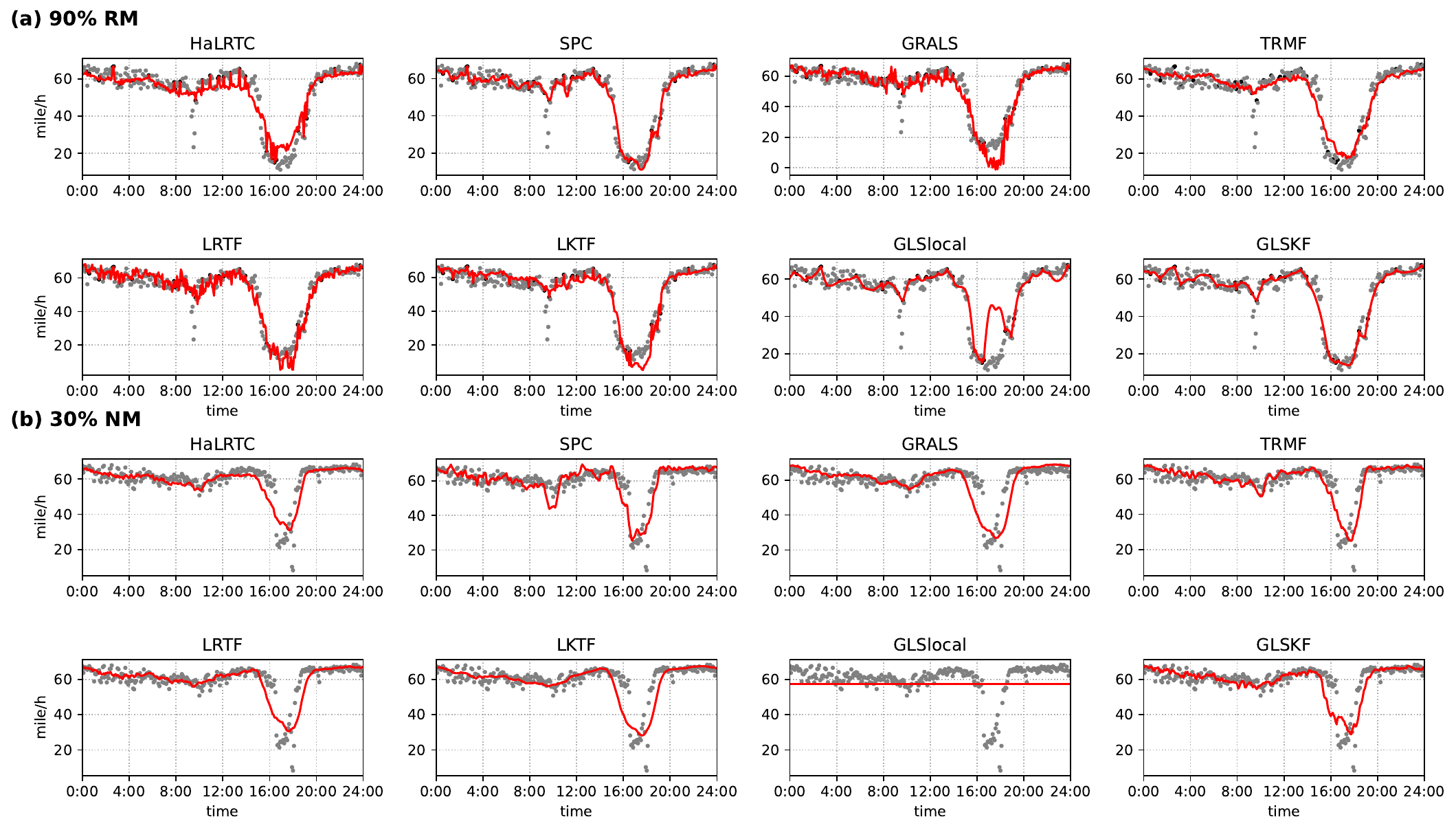}
\caption{Illustration of traffic speed imputation for dataset (S) under (a) 90\% RM and (b) 30\% NM, showing the imputed results for day 15 under RM and day 6 under NM at location 200. Black and gray dots represent the observed and missing entries, respectively.}
\label{fig:traffic-completion}
\end{figure*}

Table~\ref{tab:Traffic-comTable} presents MAE and RMSE for the two datasets under different missing scenarios and sampling rates. Figure~\ref{fig:traffic-completion} shows examples of daily reconstructions on the Seattle dataset under 90\% RM and 30\% NM; results under 70\% RM and 50\% NM are given in Supplementary~\ref{appendix_traffic_results}, Figure~\ref{fig:traffic-completion_appendix}. GLSKF consistently achieves the lowest estimation errors across most scenarios, effectively recovering both long-range daily patterns and short-scale local variations, even under severe missingness, such as 90\% RM and 50\% NM. By contrast, HaLRTC produces basic low-rank reconstructions, while SPC, GRALS, and TRMF capture the global trends but often fail to recover localized fluctuations. The ablation study further supports the benefit of combining the global and local components: LKTF improves over LRTF by enforcing covariance-based constraints on the global factors, while GLSlocal performs competitively to SPC at moderate random missingness but degrades as the missing rate increases or when the missing pattern becomes nonrandom. In particular, for a completely unobserved day, GLSlocal shrinks toward the prior mean, which corresponds to the mean of the observed data after mean-centering/rescaling. This limitation is especially pronounced when the true data exhibit clear low-rank patterns and relatively weak local correlations. Overall, the results demonstrate that jointly modeling global low-rank structure and locally correlated residuals leads to consistent gains in accuracy for traffic speed imputation.

\begin{table*}[!t]
  \centering
  \caption{Comparison of traffic speed imputation (MAE/RMSE).}
    \begin{tabular}{lll|rrrrrrrr}
    \toprule
    & MS & {SR} & {HaLRTC} & {SPC} & {GRALS} & {TRMF} & LRTF & LKTF & GLSlocal & {GLSKF}  \\
    \midrule
    \multirow{5}{*}{(S)} & RM & 0.7 & 2.42/3.70 & 2.52/3.70 & 3.14/5.11 & 2.87/4.52 & 3.17/5.13 & 3.15/5.10 & \underline{2.28}/\underline{3.40} & \textbf{2.21}/\textbf{3.30}   \\
    \multirow{2}{*} & & 0.3 & 3.13/4.81 & \underline{2.66}/\underline{3.98} & 3.17/5.13 & 2.95/4.64 & 3.20/5.18 & 3.17/5.15 & 2.78/4.26 & \textbf{2.47}/\textbf{3.80} \\
    & & 0.1 & 4.29/6.29 & \underline{2.96}/\underline{4.54} & 3.31/5.33 & 3.44/5.34 & 3.32/5.36 & 3.24/5.25 & 4.16/6.65 & \textbf{2.81}/\textbf{4.46} \\
    \cmidrule{2-11}
    & NM & 0.7 & 3.10/5.02 & 2.91/\textbf{4.42} & 3.38/5.52 & \underline{2.89}/4.54 & 3.38/5.51 & 3.35/5.48 & 5.78/8.93 & \textbf{2.87}/\underline{4.69} \\
    & & 0.5 & 3.56/5.78 & \underline{3.11}/\textbf{4.75} & 3.55/5.98 & 3.20/5.11 & 3.57/5.86 & 3.57/5.85 & 6.24/9.77 & \textbf{3.10}/\underline{5.11} \\
    \midrule
    \multirow{5}{*}{(P)} & RM & 0.7 & 1.24/2.29 & 2.07/3.68 & 2.02/3.76 & 2.13/3.75 & 2.04/3.78 & 2.02/3.75 & \underline{0.92}/\underline{1.77} & \textbf{0.86}/\textbf{1.64} \\
    \multirow{2}{*} & & 0.3 & 1.98/3.57 & 2.18/3.81 & 2.06/3.81 & 2.27/4.04 & 2.10/3.83 & 2.06/3.79 & \underline{1.46}/\underline{2.88} & \textbf{1.19}/\textbf{2.36}  \\
    & & 0.1 & 3.05/4.95 & 2.53/4.34 & \underline{2.13}/4.31 & 3.11/5.21 & 2.41/4.50 & 2.20/\underline{4.07} & 2.84/5.32 & \textbf{1.73}/\textbf{3.43} \\
    \cmidrule{2-11}
    & NM & 0.7 & \underline{2.12}/\underline{3.93} & 2.19/3.96 & 2.46/4.41 & 2.22/3.97 & 2.44/4.58 & 2.46/4.42 & 4.40/7.26 & \textbf{2.05}/\textbf{3.80} \\
    & & 0.5 & 2.48/4.47 & 2.36/4.26 & 2.75/4.97 & \underline{2.26}/\textbf{4.05} & 2.83/4.92 & 2.75/4.98 & 4.66/7.68 & \textbf{2.26}/\underline{4.13} \\
    \bottomrule
    \multicolumn{10}{l}{Best results are highlighted in bold, and second-best results are underlined.}
    \end{tabular}
  \label{tab:Traffic-comTable}
\end{table*}

\subsection{Color Image Inpainting}
\subsubsection{Datasets}
We evaluate GLSKF on color image inpainting with random pixel masking at sampling rates $\text{SR}\in\{0.2, 0.1, 0.05\}$ (i.e., 80\%, 90\%, and 95\% missingness) and structural missing (SM), including text- and stripe/fiber- masked inpainting and image demosaicing. We use seven benchmark color images as test data. Each image is represented as a third-order tensor with dimensions $\textit{height} \times \textit{width} \times \textit{channel}$; four images have size $256\times256\times3$ and three are $512\times512\times3$.

\subsubsection{Configuration}
\paragraph{Baselines} We compare GLSKF with HaLRTC \cite{liu2012tensor}, SPC \cite{yokota2016smooth}, and three related variants: LRTF, LKTF, GLSlocal. In addition, we include two learning-based baselines, the deep net prior (DNP) model and GLON \cite{zhao2021tensor}, a related Global-LOcal-Nonlocal framework built upon deep neural networks.

\paragraph{Parameter settings} We use the same kernel and rank settings for all the test images. For the first and second \textit{pixel} dimensions, we use a Mat\'ern 3/2 kernel to construct $\boldsymbol{K}_{\boldsymbol{u}}^{(1)}$ and $\boldsymbol{K}_{\boldsymbol{u}}^{(2)}$, and a tapered Mat\'ern 3/2 kernel with a Bohman taper for $\boldsymbol{K}_{\boldsymbol{\mathcal{R}}}^{(1)}$ and $\boldsymbol{K}_{\boldsymbol{\mathcal{R}}}^{(2)}$. For the third \textit{channel} dimension, we set $\boldsymbol{K}_{\boldsymbol{u}}^{(3)} = \boldsymbol{I}_3$ and learn $\boldsymbol{K}_{\boldsymbol{\mathcal{R}}}^{(3)}$ as the covariance across rows of the mode-3 unfolding of $\boldsymbol{\mathcal{R}}$.

We fix the variance to 1 for all kernels. For kernel hyperparameters, we assume the length-scales are shared across the row and column pixel dimensions. Following the kernel hyperparameter specification protocol described in Supplementary~\ref{supp:kernelHypeSelection}, we empirically set the values as $\ell_{\boldsymbol{u}}^{(1)}=\ell_{\boldsymbol{u}}^{(2)}=30$ for the latent factor covariances, and $\ell_{\boldsymbol{r}}^{(1)}=\ell_{\boldsymbol{r}}^{(2)}=5$ with tapering ranges $\lambda_1=\lambda_2=30$ for the local residual. We fix the CP rank at $R=10$ and select the regularization parameters $\rho$ and $\gamma$ from \{1,\,5,\,10,\,15,\,20\} and \{0.1,\,0.2,\,1,\,5,\,10\}, respectively; see Figure~\ref{fig:image_parameter} in Supplementary~\ref{appendix:regularization} for details. The parameter settings are summarized in Tables~\ref{tab:image-kernel}-\ref{tab:image-config-regularization-SM} in Supplementary~\ref{appendix-sec:configuration}.

For the baseline methods, we conduct SPC with TV and QV and report the better result; use the same rank ($R=10$) and analogous covariance configurations as GLSKF for \{LRTF, LKTF, and GLSlocal\}; additionally, tune the $\alpha$ parameter in LRTF from \{0.1,\,0.2,\,1,\,5,\,10\}.

\subsubsection{Results}
\begin{figure*}[!t]
\centering
\includegraphics[width=0.9\textwidth]{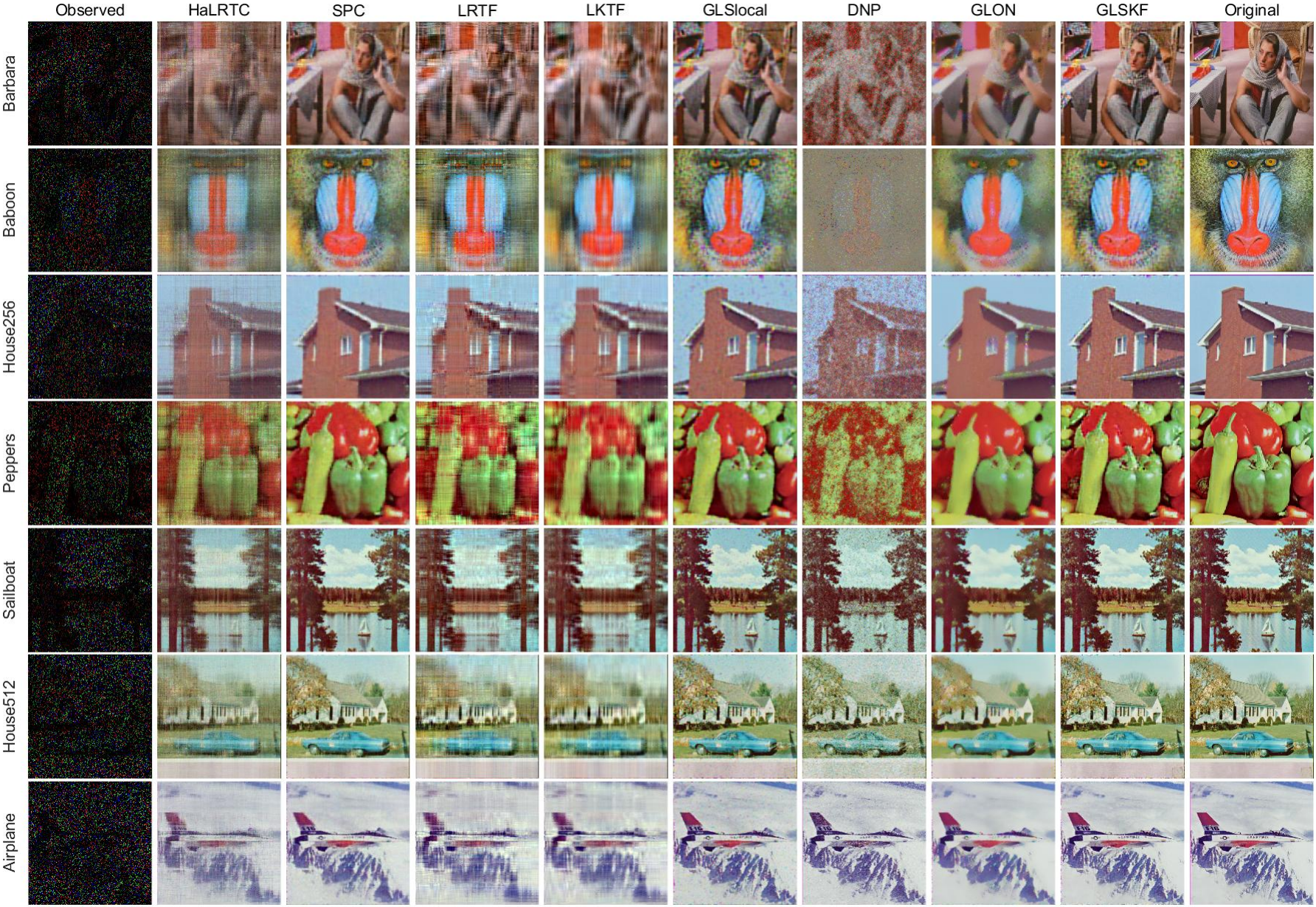}
\caption{Results of image inpainting under 90\% RM, i.e., SR $=0.1$.}
\label{fig:image-inpainting-RM}
\end{figure*}

\begin{figure*}[!t]
\centering
\includegraphics[width=0.9\textwidth]{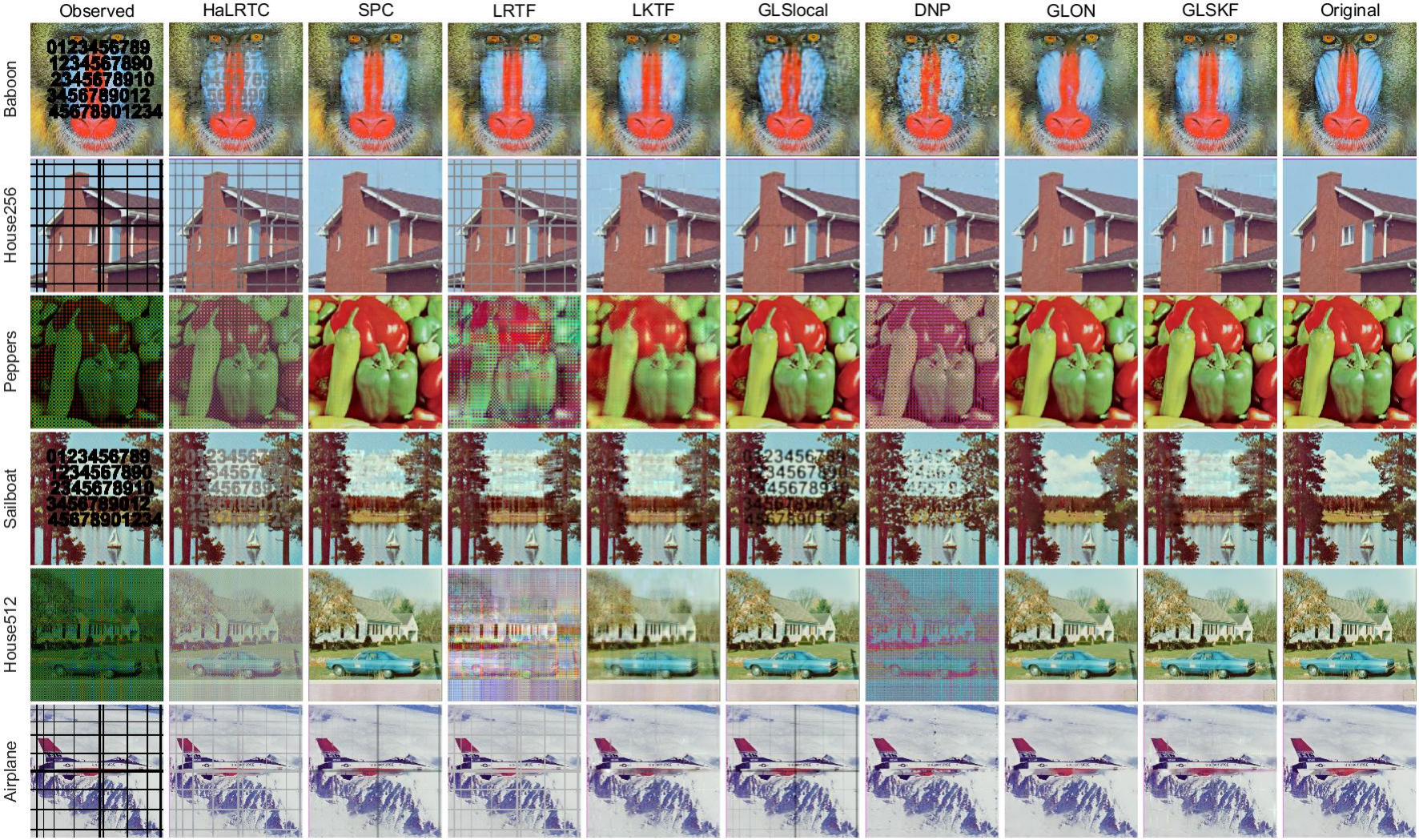}
\caption{Illustration of color image completion under SM: \textit{Baboon}, \textit{Sailboat} with text missingness, \textit{House256}, \textit{Airplane} with stripe missingness, and \textit{Peppers}, \textit{House512} for demosaicing.}
\label{fig:image-inpainting-SM}
\end{figure*}

Tables~\ref{tab:Image-comTable-RM} and~\ref{tab:Image-comTable-SM} in Supplementary~\ref{appendix_image_results} report the inpainting performance in terms of PSNR and SSIM for the test images under various missing scenarios. Figures~\ref{fig:image-inpainting-RM} and~\ref{fig:image-inpainting-SM} compare the completed images under 90\% RM and SM, respectively. The full completion results under the three SM scenarios are given in Supplementary~\ref{appendix_image_results}, Figures~\ref{fig:image_text}-\ref{fig:image_demosaic}. As illustrated, GLSKF consistently delivers the best inpainting results, both quantitatively (with higher PSNR/SSIM) and visually, across nearly all test images. Particularly, LKTF improves over the purely low-rank LRTF at extreme missingness (e.g., 95\% RM) benefiting from the covariance-based smoothness constraints on the global factors; GLSlocal is comparable to SPC at lower missing rates but underperforms as missingness increases or under structured missingness due to the lack of an explicit global low-rank structure. The deep neural network baselines may fail under severe random missingness, while performing better under SM scenarios where the missing rates are relatively lower. Nevertheless, GLSKF still provides comparable or better results using purely the observed entries and with much faster training. By combining a global low-rank component with a locally correlated residual module, GLSKF achieves superior reconstruction quality in both RM and SM scenarios even with a small rank ($R=10$), validating the effectiveness of the proposed complementary global-local model structure.

\begin{table*}[!t]
  \centering
  \caption{Comparison of color image inpainting under RM (PSNR/SSIM).}
    \begin{tabular}{ll|rrrrrrrr}
    \toprule
     & {SR} & {HaLRTC} & {SPC} & LRTF & LKTF & GLSlocal & DNP & GLON & GLSKF  \\
    \midrule
    \multirow{3}{*}{\rotatebox{90}{Barbara}} & 0.2 & 22.36/0.599 & \underline{27.01}/\underline{0.812} & 21.14/0.531 & 21.55/0.580 & 26.26/0.808 & 23.22/0.598 & 25.29/0.764 & \textbf{27.85}/\textbf{0.857}   \\
    & 0.1 & 18.58/0.393 & \underline{24.83}/\underline{0.730} & 19.69/0.412 & 20.70/0.523 & 23.77/0.717 & 16.64/0.208 & 22.76/0.648 & \textbf{25.60}/\textbf{0.792} \\
    & 0.05 & 15.32/0.249 & \underline{22.68}/\underline{0.636} & 12.77/0.085 & 19.64/0.452 & 20.24/0.578 & 14.51/0.221 & 20.41/0.535 & \textbf{23.36}/\textbf{0.716}  \\
    \cmidrule(r){2-10}
    \multirow{3}{*}{\rotatebox{90}{Baboon}} & 0.2 & 20.17/0.441 & \textbf{22.79}/\underline{0.621} & 20.38/0.436 & 20.79/0.469 & 22.15/0.597 & 18.16/0.303 & 22.38/0.581 & \underline{22.65}/\textbf{0.638}  \\
    & 0.1 & 17.81/0.289 & \textbf{21.38}/\underline{0.486} & {18.22}/0.265 & 19.93/0.364 & 20.54/0.469 & 14.33/0.198 & 21.06/0.448 & \underline{21.32}/\textbf{0.510} \\
    & 0.05 & 15.00/0.191 & \underline{20.31}/\underline{0.384} & {13.12}/0.083 & 18.94/0.282 & 18.26/0.345 & 14.05/0.165 & 19.97/0.355 & \textbf{22.07}/\textbf{0.629} \\
    \cmidrule(r){2-10}
    \multirow{3}{*}{\rotatebox{90}{House256}} & 0.2 & 23.98/0.659 & 27.61/0.803 & 22.43/0.590 & 23.21/0.653 & \underline{27.97}/\underline{0.817} & 25.07/0.543 & 26.58/0.811 & \textbf{29.43}/\textbf{0.856}  \\
    & 0.1 & 20.27/0.471 & \underline{25.73}/\underline{0.748} & 20.97/0.463 & 22.17/0.588 & 24.62/0.714 & 17.48/0.179 & 24.01/0.738 & \textbf{27.17}/\textbf{0.788} \\
    & 0.05 & 16.64/0.312 & \underline{24.03}/\underline{0.695} & 15.23/0.145 & 21.70/0.549 & 19.49/0.539 & 14.88/0.130 & 21.79/0.673 & \textbf{25.08}/\textbf{0.739} \\
    \cmidrule(r){2-10}
    \multirow{3}{*}{\rotatebox{90}{Peppers}} & 0.2 & 20.58/0.537 & 25.85/0.810 & 19.41/0.463 & 19.59/0.523 & \underline{26.65}/\underline{0.862} & 23.16/0.574 & 25.21/0.821 & \textbf{27.38}/\textbf{0.864}  \\
    & 0.1 & 16.62/0.334 & 23.60/0.738 & 17.65/0.316 & 18.74/0.457 & \underline{23.61}/\underline{0.773} & 15.50/0.216 & 22.34/0.723 & \textbf{24.92}/\textbf{0.798} \\
    & 0.05 & 13.38/0.210 & \underline{21.55}/\underline{0.659} & 11.35/0.071 & 18.04/0.393 & 19.70/0.627 & 13.15/0.221 & 19.84/0.628 & \textbf{22.91}/\textbf{0.735} \\
    \midrule
    \multirow{3}{*}{\rotatebox{90}{Sailboat}} & 0.2 & 22.24/0.761 & \underline{25.93}/0.884 & 20.28/0.620 & 20.44/0.639 & 25.62/\underline{0.889} & 23.48/0.780 & 24.71/0.866 & \textbf{26.77}/\textbf{0.908}  \\
    & 0.1 & 19.27/0.593 & \underline{23.79}/\underline{0.806} & 19.20/0.526 & 19.82/0.568 & 23.07/0.776 & 19.01/0.537 & 22.43/0.765 & \textbf{24.56}/\textbf{0.847} \\
    & 0.05 & 16.74/0.442 & \underline{21.91}/\underline{0.707} & {16.62}/0.364 & 19.35/0.516 & 19.47/0.578 & 13.40/0.222 & 20.60/0.660 & \textbf{22.99}/\textbf{0.786}  \\
    \cmidrule(r){2-10}
    \multirow{3}{*}{\rotatebox{90}{House512}} & 0.2 & 23.56/0.802 & \underline{26.39}/0.885 & 21.30/0.682 & 21.30/0.688 & 26.15/\underline{0.904} & 25.12/0.789 & 24.78/0.870 & \textbf{27.52}/\textbf{0.922}  \\
    & 0.1 & 20.41/0.642 & \underline{24.30}/\underline{0.814} & 20.31/0.594 & 20.60/0.617 & 23.25/0.777 & 20.17/0.554 & 22.52/0.768 & \textbf{24.86}/\textbf{0.867} \\
    & 0.05 & 17.91/0.500 & \underline{22.45}/\underline{0.724} & {18.03}/0.461 & 20.05/0.561 & 19.18/0.540 & 17.17/0.350 & 20.84/0.665 & \textbf{23.30}/\textbf{0.804}  \\
    \cmidrule(r){2-10}
    \multirow{3}{*}{\rotatebox{90}{Airplane}} & 0.2 & 24.57/0.843 & 26.54/0.892 & 21.61/0.709 & 21.67/0.719 & \underline{27.76}/\underline{0.932} & 25.47/0.802 & 26.19/0.912 & \textbf{30.13}/\textbf{0.959}  \\
    & 0.1 & 20.92/0.693 & \underline{24.92}/\underline{0.843} & 20.70/0.621 & 21.24/0.674 & 23.96/0.790 & 23.79/0.780 & 23.68/0.833 & \textbf{27.43}/\textbf{0.933} \\
    & 0.05 & 18.21/0.552 & \underline{23.35}/\underline{0.784} & 18.18/0.484 & 20.48/0.603 & 18.46/0.501 & 21.70/0.610 & 21.82/0.753 & \textbf{25.03}/\textbf{0.882}  \\
    \bottomrule
    \multicolumn{8}{l}{Best results are highlighted in bold, and second-best results are underlined.}
    \end{tabular}
  \label{tab:Image-comTable-RM}
\end{table*}

\subsection{Color Video Completion}
\subsubsection{Datasets}We next evaluate GLSKF on fourth-order color video completion using three benchmark datasets\footnote{\url{http://trace.eas.asu.edu/yuv/}}: \textit{Hall} (H), \textit{Foreman} (F), and \textit{Carphone} (C). Each video is represented as a $\textit{height}\times\textit{width}\times\textit{channel}\times\textit{frame}$ tensor of size $144 \times 176 \times 3 \times 300$. For each video, we randomly mask pixels according to sampling rates $\text{SR}\in\{0.2,\, 0.1,\, 0.05\}$, corresponding to 80\%, 90\%, and 95\% missingness, respectively.

\subsubsection{Configuration}
\paragraph{Baselines} We compare GLSKF with HaLRTC \cite{liu2012tensor}, SPC \cite{yokota2016smooth}, and three baselines that share similar model structures: LRTF, LKTF, and GLSlocal. We also include the neural-network-based deep net prior (DNP) model as an additional baseline.

\paragraph{Parameter settings} We use the same kernel and rank settings for the three video datasets. For the latent-factor covariances, we use a Mat\'ern 3/2 kernel to compute $\{\boldsymbol{K}_{\boldsymbol{u}}^{(1)},\,\boldsymbol{K}_{\boldsymbol{u}}^{(2)},\, \boldsymbol{K}_{\boldsymbol{u}}^{(4)}\}$ for the first (pixel-row), second (pixel-column), and fourth (frame) dimensions. For the third (channel) dimension, we assume independence and set $\boldsymbol{K}_{\boldsymbol{u}}^{(3)} = \boldsymbol{I}_3$. For the local component, we use a tapered Mat\'ern 3/2 kernel with a Bohman taper to construct $\{\boldsymbol{K}_{\boldsymbol{\mathcal{R}}}^{(1)},\, \boldsymbol{K}_{\boldsymbol{\mathcal{R}}}^{(2)},\, \boldsymbol{K}_{\boldsymbol{\mathcal{R}}}^{(4)}\}$. The channel dimension covariance $\boldsymbol{K}_{\boldsymbol{\mathcal{R}}}^{(3)}$ is computed as the row covariance of the mode-3 unfolding of $\boldsymbol{\mathcal{R}}$.

We empirically select the kernel hyperparameters $\{\ell_{\boldsymbol{u}}^{(d)},\ell_{\boldsymbol{r}}^{(d)},\lambda_d\}$ based on prior knowledge of the data. For the first two pixel dimensions, we set the length-scales to $\ell_{\boldsymbol{u}}^{(1)}=\ell_{\boldsymbol{u}}^{(2)}=30$ for the global factor covariances and $\ell_{\boldsymbol{r}}^{(1)}=\ell_{\boldsymbol{r}}^{(2)}=5$ with tapering ranges $\lambda_1=\lambda_2=10$ for the residual component kernel. For the fourth (frame) dimension, we set $\ell_{\boldsymbol{u}}^{(4)}=\ell_{\boldsymbol{r}}^{(4)}=5$ for both the factor and local components, with a tapering range $\lambda_4=10$. We fix the CP rank at $R=10$ for global factorization estimation and select the regularization parameters $\{\rho,\gamma\}$ from $\{0.1,\, 0.2,\, 1,\, 5,\, 10\}$ based on observed-validation. Examples of the regularization parameter selection procedure are provided in Supplementary~\ref{appendix:regularization}, Figure~\ref{fig:video_parameter}, and the applied parameters together with kernel settings are in Supplementary~\ref{appendix-sec:configuration}, Tables~\ref{tab:video-kernel}-\ref{tab:video-regularization}.

For baselines, we implement SPC using QV-based regularization, use the same rank $R=10$ for LRTF and LKTF, and select $\alpha$ from $\{0.1,\, 0.2,\, 1,\, 5,\, 10\}$ for LRTF.

\subsubsection{Results}
\begin{figure*}[!t]
\centering
\includegraphics[width=0.9\textwidth]{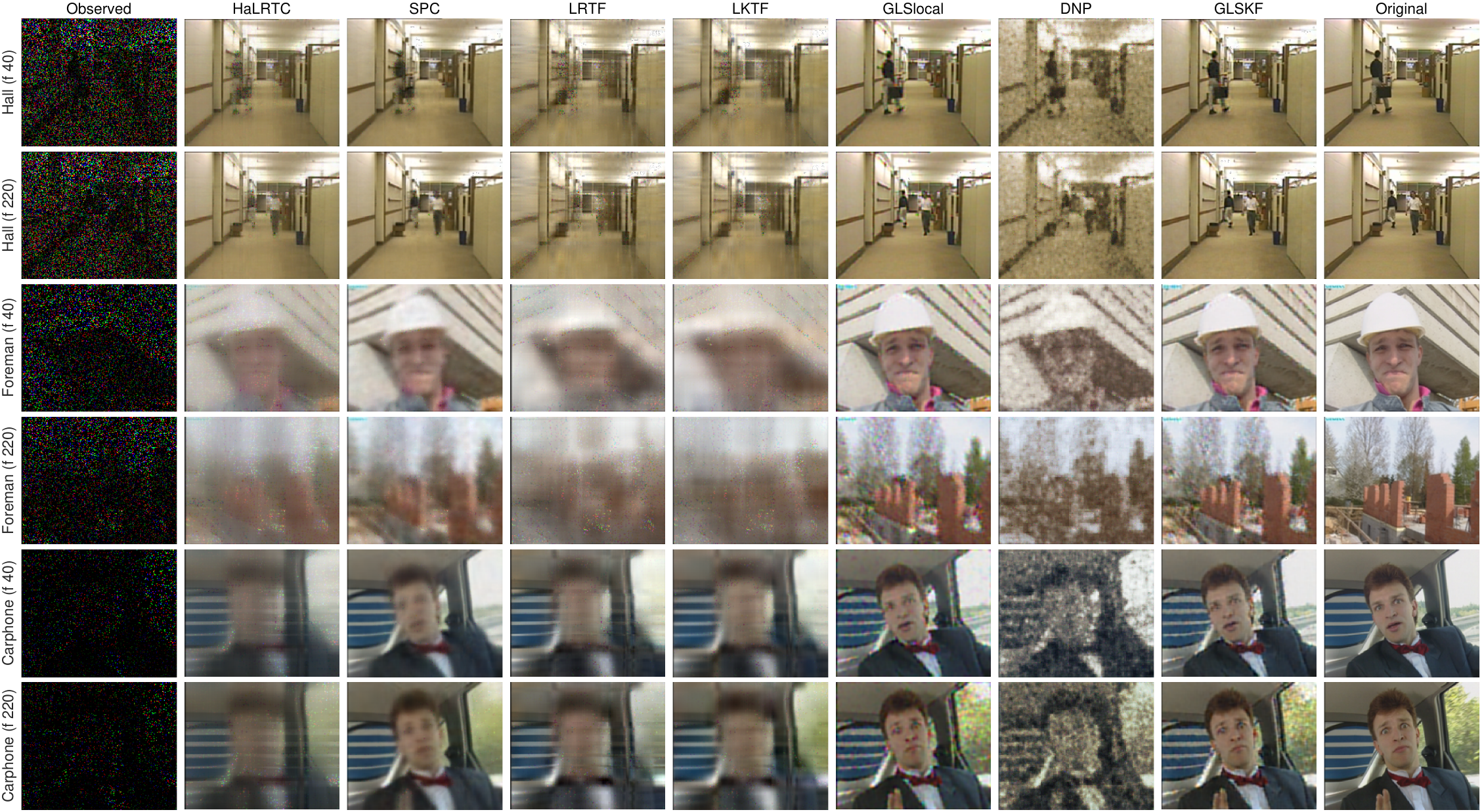}
\caption{Illustration of video completion. Rows 1-2, 3-4, and 5-6 correspond to \textit{Hall} 80\% RM, \textit{Foreman} 90\% RM, and \textit{Carphone} 95\% RM, respectively.
}
\label{fig:video-completion}
\end{figure*}

Table~\ref{tab:video-comTable} summarizes PSNR/SSIM on the three video datasets under different sampling rates. Figure~\ref{fig:video-completion} presents visual reconstructions for the 40th and 220th frames of the three datasets at 80\%, 90\%, and 95\% missingness, respectively. Among the competing methods, SPC outperforms most baselines, while GLSlocal remains comparable particularly on \textit{Foreman} and \textit{Carphone}, mainly by exploiting the temporal/frame continuity through the local covariance structure. The neural network baseline underperforms in this setting due to the limited amount of observed data available for training. Across all datasets and sampling rates, GLSKF achieves superior performance with a small CP rank ($R=10$), indicating that jointly modeling a low-rank global component with a locally correlated residual improves both overall structure recovery and fine-detail reconstruction.

\begin{table*}[!t]
  \centering
  \caption{Comparison of color video completion (PSNR/SSIM).}
    \begin{tabular}{ll|rrrrrrr}
    \toprule
    & {SR} & {HaLRTC} & {SPC} & LRTF & LKTF & GLSlocal & DNP & {GLSKF}  \\
    \midrule
    \multirow{3}{*}{\rotatebox{90}{Hall}} & 0.2 & 27.41/0.886 & 30.70/0.924 & 24.80/0.757 & 24.60/0.745 & \underline{30.94}/\underline{0.945} & 21.92/0.530 & \textbf{34.06}/\textbf{0.961}  \\
    & 0.1 & 23.59/0.786 & \underline{29.81}/\underline{0.914} & 24.25/0.741 & 23.89/0.730 & 27.16/0.891 & 20.79/0.442 & \textbf{31.74}/\textbf{0.935} \\
    & 0.05 & 20.57/0.665 & \underline{27.97}/\underline{0.885} & 23.82/0.733 & 23.63/0.706 & 23.93/0.802 & 18.16/0.264 & \textbf{29.39}/\textbf{0.900}  \\
    \midrule
    \multirow{3}{*}{\rotatebox{90}{Foreman}} & 0.2 & 23.00/0.642 & 26.31/0.761 & 20.05/0.488 & 20.06/0.491 & \underline{29.26}/\underline{0.874} & 23.39/0.601 & \textbf{32.25}/\textbf{0.920} \\
    & 0.1 & 19.69/0.476 & 24.91/0.705 & 19.62/0.467 & 19.43/0.463 & \underline{26.64}/\underline{0.801} & 22.08/0.518 & \textbf{29.48}/\textbf{0.865} \\
    & 0.05 & 17.25/0.390 & 23.50/0.645 & 19.30/0.452 & 18.97/0.449 & \underline{24.34}/\underline{0.715} & 19.86/0.378 & \textbf{27.34}/\textbf{0.802} \\
    \midrule
    \multirow{3}{*}{\rotatebox{90}{Carphone}} & 0.2 & 25.98/0.807 & 29.12/0.864 & 21.87/0.626 & 21.91/0.625 & \underline{30.76}/\underline{0.919} & 22.94/0.571 & \textbf{33.01}/\textbf{0.943}  \\
    & 0.1 & 22.21/0.676 & 27.63/0.828 & 21.46/0.609 & 21.42/0.605 & \underline{28.11}/\underline{0.869} & 21.95/0.512 & \textbf{30.60}/\textbf{0.903} \\
    & 0.05 & 19.21/0.562 & \underline{27.19}/\underline{0.818} & 21.31/0.601 & 21.23/0.597 & 25.77/0.805 & 19.97/0.376 & \textbf{28.16}/\textbf{0.866} \\
    \bottomrule
    \multicolumn{8}{l}{{Best results are highlighted in bold, and second-best results are underlined.}}
    \end{tabular}
  \label{tab:video-comTable}
\end{table*}

\subsection{MRI Image Completion}
\subsubsection{Datasets}
We lastly evaluate GLSKF on MRI image completion using a simulated MRI volume from BrainWeb\footnote{\url{https://zenodo.org/records/8067595}}. The dataset is represented as a third-order tensor of size $220 \times 220 \times 184$ ($\textit{height}\times\textit{width}\times\textit{depth}$). Following the settings for color image and video completion, we randomly mask 80\%, 90\%, 95\%, and 99\% of the entries as missing/test data, corresponding to sampling rates $\text{SR}\in\{0.2,\, 0.1,\, 0.05,\, 0.01\}$. In addition, we also test the model under structured patch missing (PM) with patch size $16\times 16 \times 2$ ($\textit{height}\times\textit{width}\times\textit{depth}$) and sampling rates $\text{SR}\in\{0.2, 0.1\}$ to better illustrate the component role and model advantage.

\subsubsection{Configuration}
\paragraph{Baselines} We compare GLSKF with HaLRTC \cite{liu2012tensor}, SPC \cite{yokota2016smooth}, three related baselines, namely LRTF, LKTF, and GLSlocal, and the neural-network DNP model.

\paragraph{Parameter settings} We employ a Mat\'ern 3/2 kernel to compute $\{\boldsymbol{K}_{\boldsymbol{u}}^{(1)},\boldsymbol{K}_{\boldsymbol{u}}^{(2)},\boldsymbol{K}_{\boldsymbol{u}}^{(3)}\}$ for the latent factors and use a tapered Mat\'ern 3/2 kernel with a Bohman taper to construct $\{\boldsymbol{K}_{\boldsymbol{\mathcal{R}}}^{(1)},\boldsymbol{K}_{\boldsymbol{\mathcal{R}}}^{(2)},\boldsymbol{K}_{\boldsymbol{\mathcal{R}}}^{(3)}\}$. The kernel hyperparameters are selected similarly to the settings for color image and video completion. For the first two dimensions ($d=1,2$), we set the length-scales to $\ell_{\boldsymbol{u}}^{(d)}=30$ and $\ell_{\boldsymbol{r}}^{(d)}=5$ for $\boldsymbol{K}_{\boldsymbol{u}}^{(d)}$ and $\boldsymbol{K}_{\boldsymbol{\mathcal{R}}}^{(d)}$, respectively, and use tapering ranges $\lambda_1=\lambda_2=10$, as in the video experiments. For the third (depth) dimension, we use a shorter length-scale $\ell_{\boldsymbol{u}}^{(3)}=\ell_{\boldsymbol{r}}^{(3)}=5$ for both global and local covariances with tapering range $\lambda_3=10$. The rank is fixed at $R=10$, and the weight parameters $\rho$ and $\gamma$ are selected from $\{0.1,\,0.2,\,1,\,5,\,10\}$. Related parameter tuning process see Figure~\ref{fig:MRI_parameter} in Supplementary~\ref{appendix-sec:selection}, and the applied parameter settings are summarized in Supplementary~\ref{appendix-sec:configuration}, Table~\ref{tab:MRI-config}. For the baseline methods, we configure SPC with QV regularization; for LRTF and LKTF, we use the same rank as GLSKF ($R=10$) and adopt parameter settings analogous to those for color image/video completion.

\subsubsection{Results}
\begin{figure*}[!t]
\centering
\includegraphics[width=0.9\textwidth]{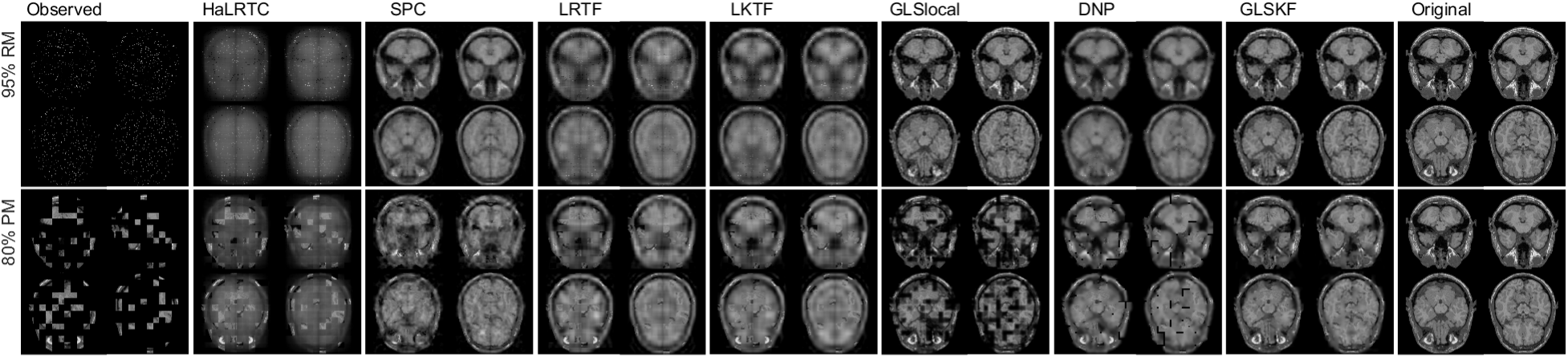}
\caption{MRI completion for four frontal slices (30th, 40th, 50th, 60th) under 95\% random missingness (RM) and 80\% structured patch missingness (PM).}
\label{fig:MRIcompletion}
\end{figure*}

Table~\ref{tab:MRI-comTable} reports PSNR and MAE results for MRI completion across sampling rates under random missingness (RM) and patch missingness (PM). The corresponding RMSE results are provided in Supplementary~\ref{appendix_MRI}, Table~\ref{tab:MRI-comTable-RMSE}. Figure~\ref{fig:MRIcompletion} shows reconstruction examples for four frontal slices under 95\% RM and 80\% PM, with additional results given in Supplementary~\ref{appendix_MRI}, Figure~\ref{fig:MRIcompletion_appendix}. GLSlocal outperforms SPC and other low-rank baselines for $\text{SR}\in\{0.2,0.1,0.05\}$ under RM, highlighting the importance of modeling local spatial structure/correlation for MRI data under random missingness. This observation suggests that MRI volumes contain strong local anatomical smoothness; thus, the local covariance component alone can recover much of the missing structure when sufficient neighboring observations are available. However, under extreme missingness, e.g., 99\% RM, or under nonrandom PM, the performance of GLSlocal decreases, indicating that incorporating a global component becomes necessary for accurate recovery in such scenarios.

Across all sampling rates and both missing scenarios, GLSKF achieves the highest PSNR and the lowest estimation errors. The visual results further indicate faithful recovery of both fine structural details and the global anatomical structure, even under severe missing conditions. The role of the global component becomes more important when the sampling rate is lower or when missing entries form larger structured gaps, such as under PM, where local neighborhoods provide insufficient information. Therefore, the relative contribution of the global and local components is data- and missingness-dependent. GLSKF is designed to provide a complementary representation: the global low-rank component captures large-scale anatomical structure, while the local residual component captures short-range spatial variation.

\begin{table*}[!t]
  \centering
  \caption{Comparison of MRI image completion (PSNR/MAE).}
    \begin{tabular}{ll|rrrrrrr}
    \toprule
    MS & {SR} & {HaLRTC} & {SPC} & LRTF & LKTF & GLSlocal & DNP & {GLSKF}  \\
    \midrule
    RM & 0.2 & 23.35/0.0396 & 26.55/0.0263 & 21.48/0.0515 & 21.24/0.0514 & \underline{29.04}/\underline{0.0179} & 25.79/0.0346 & \textbf{29.05}/\textbf{0.0178}  \\
    & 0.1 & 20.18/0.0559 & 25.13/0.0289 & 20.74/0.0528 & 20.82/0.0509 & \underline{27.31}/\underline{0.0207} & 25.47/0.0347 & \textbf{27.34}/\textbf{0.0205} \\
    & 0.05 & 18.07/0.0744 & 23.76/0.0327 & 20.49/0.0529 & 20.36/0.0523 & \underline{25.65}/\underline{0.0246} & 24.95/0.0351 & \textbf{25.77}/\textbf{0.0238} \\
    & 0.01 & 14.33/0.1248 & 20.29/\underline{0.0497} & 18.19/0.0727 & 19.98/0.0537 & 19.80/0.0526 & \underline{20.66}/0.0722 & \textbf{22.90}/\textbf{0.0353}  \\
    \midrule
    PM & 0.2 & 18.95/0.0757 & \underline{22.44}/\underline{0.0452} & 20.95/0.0562 & 20.73/0.0556 & 19.57/0.0756 & 22.42/0.0473 & \textbf{23.42}/\textbf{0.0376} \\
    & 0.1 & 15.84/0.1023 & \underline{20.05}/\underline{0.0563} & 17.05/0.0879 & 18.75/0.0689 & 17.10/0.1073 & 19.78/0.0646 & \textbf{21.05}/\textbf{0.0496} \\
    \bottomrule
    \multicolumn{9}{l}{{Best results are highlighted in bold, and second-best results are underlined.}}
    \end{tabular}
  \label{tab:MRI-comTable}
\end{table*}

\subsection{Sensitivity Analysis} \label{subsec:sensitivity}
We conduct sensitivity analyses on the rank, kernel hyperparameters, and kernel function settings using Seattle traffic speed (S) imputation and \textit{House256} color image completion.

\begin{figure*}[!t]
\centering
\includegraphics[width=0.9\textwidth]{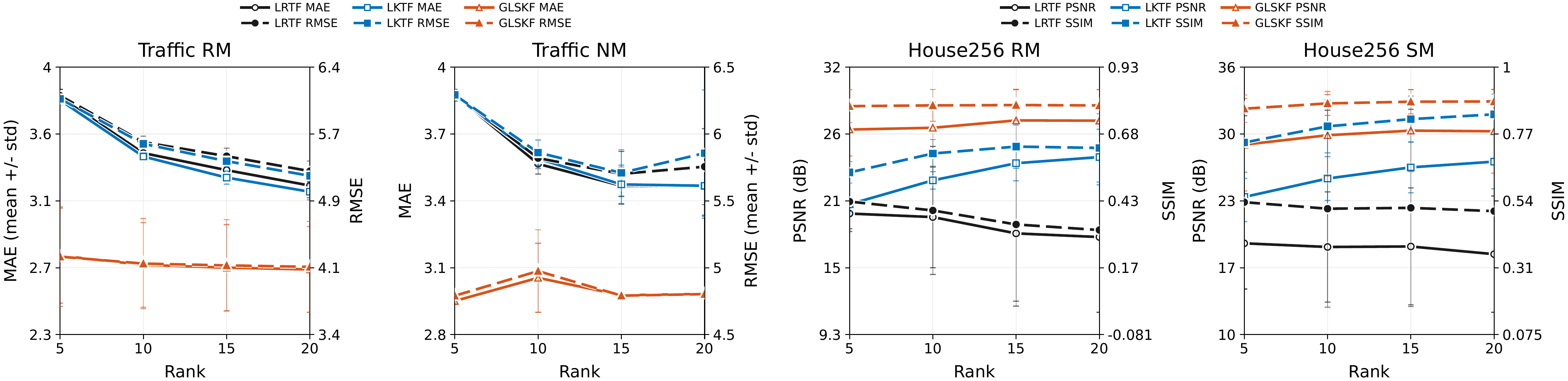}
\caption{Effect of rank on completion performance for traffic speed dataset (S) under RM and NM, and for color image \textit{House256} under RM and SM.}
\label{fig:rank}
\end{figure*}

\paragraph{Effect of rank}
To assess the sensitivity of GLSKF to the rank of the global low-rank component, we vary the rank over $\{5,\,10,\,15,\,20\}$ on the Seattle traffic speed dataset (S) and the \textit{House256} color image. Figure~\ref{fig:rank} compares the reconstruction errors of GLSKF and related low-rank methods \{LRTF, LKTF\} under different rank settings. The results show that GLSKF achieves stable and high-quality completion performance even with a small rank, e.g., $R=5$, benefiting from the complementary local residual component. This indicates that the proposed global-local structure can reduce the reliance on a large global rank and thereby lower the associated computational cost, since the locally correlated residual component captures short-range variations that would otherwise require a larger rank to be represented by the low-rank global component.

\begin{figure}[!t]
\centering
\includegraphics[width=0.5\textwidth]{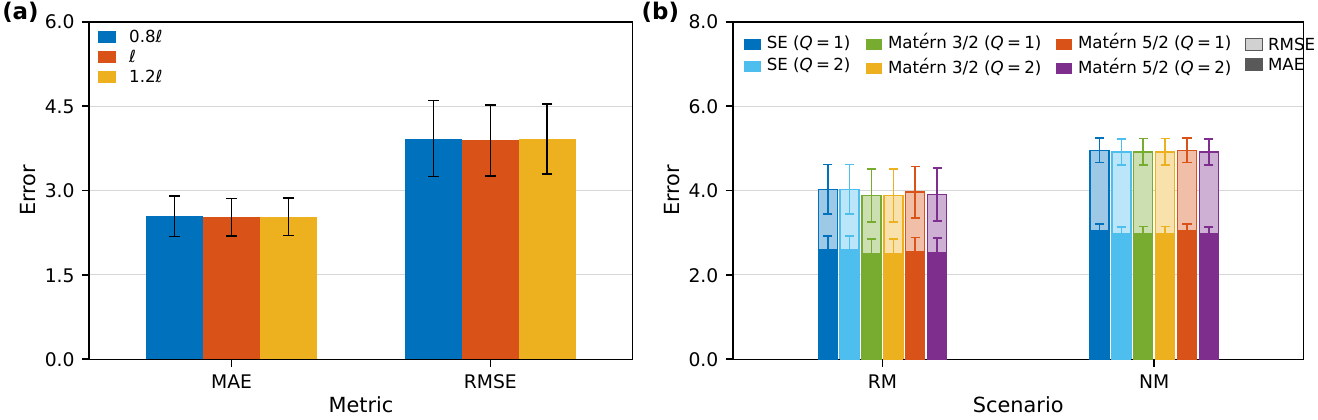}
\caption{Sensitivity analysis of GLSKF with respect to (a) kernel hyperparameters and (b) kernel functions.}
\label{fig:kernelHyper}
\end{figure}

\paragraph{Sensitivity analysis of kernel hyperparameters} 
To evaluate the sensitivity of GLSKF to the kernel hyperparameters, we perturb the pre-specified global length-scales around the corresponding original values. Specifically, we keep the local kernel hyperparameters and other model parameters fixed, and set $\ell_{\boldsymbol{u}}^{(d)}\in\{0.8\ell,\ell,1.2\ell\}$, where $\ell$ denotes the corresponding pre-specified value used in the associated experiments. We vary the temporal length-scale $\ell_{\boldsymbol{u}}^{(2)}$ for the traffic speed dataset (S). The results are shown in Figure~\ref{fig:kernelHyper}(a). As can be seen, GLSKF remains stable under moderate perturbations of the kernel length-scales. This indicates that GLSKF is not overly sensitive to small changes in the kernel hyperparameters, supporting the robustness of the pre-specified kernel setting protocol.

\paragraph{Sensitivity analysis of kernel functions}
We further evaluate the sensitivity of GLSKF to the choice of kernel functions. For traffic speed imputation on dataset (S), we keep the spatial kernel fixed as the regularized Laplacian kernel to leverage efficient inverse computation, and vary the temporal kernel among \{SE, Mat\'ern 5/2, Mat\'ern 3/2\} kernels. For the local residual component, we use the same kernel function as the global factor covariance along each corresponding dimension, and further consider nonseparable covariance structures formed by a sum of $Q$ product kernels, with $Q\in\{1,2\}$. The CP rank is fixed at $R=10$ for all experiments in this analysis. The results are shown in Figure~\ref{fig:kernelHyper}(b). As can be seen, GLSKF achieves stable performance across different kernel function choices. Using a sum of product kernels for the local component may slightly improve the performance, but it also increases the computational cost for parameter selection. Overall, the results indicate that GLSKF is not highly sensitive to a specific kernel family. The proposed covariance-regularized complementary global-local framework remains robust with respect to kernel specification and allows the underlying data structure to be effectively captured using different valid covariance functions, as long as they provide reasonable smoothness and local correlation structures.

\section{Discussion} \label{sec:Discussion}
\subsection{The Roles of Global and Local Components}
In GLSKF, the low-rank global component $\boldsymbol{\mathcal{M}}$ and the locally correlated residual tensor $\boldsymbol{\mathcal{R}}$ are additively integrated to complement each other. The complementary model structure offers two main advantages: (i) accurate reconstructions can be achieved with a small CP rank, which improves both effectiveness and efficiency for completion; (ii) the two components explain various data characteristics and yield an interpretable decomposition, facilitating downstream tasks such as pattern discovery, feature extraction, robust recovery/unusual event detection \cite{du2025joint}. Figures~\ref{fig:Component-RM} and~\ref{fig:Component-NM} in Supplementary~\ref{appendix_subsec:components} illustrate the distinct roles of $\boldsymbol{\mathcal{M}}$ and $\boldsymbol{\mathcal{R}}$ on traffic speed imputation (dataset (S) at 70\% RM and 30\% NM), image inpainting (\textit{House256}, \textit{Peppers}, and \textit{Airplane} at 90\% RM; \textit{House256} under text SM), video completion (\textit{Hall} at 80\% RM and \textit{Carphone} at 95\% RM), and MRI completion (90\% RM and 90\% PM).

As shown, under both random and structured missingness, $\boldsymbol{\mathcal{M}}$ captures the background and low-frequency structures, providing a smooth overall approximation, whereas $\boldsymbol{\mathcal{R}}$ recovers high-frequency details that preserve salient local features, e.g., short-term fluctuations in traffic speed and edges/textures in images and videos. The two components together enable GLSKF to deliver high-quality and interpretable reconstructions across diverse data modalities and various missing scenarios.

\subsection{Cross-dataset Applicability and Limitation}
GLSKF is designed as a general tensor completion framework rather than a modality-specific model. Traffic speed data, color images, videos, and MRI volumes can all be represented as incomplete multidimensional tensors, where each mode carries structural information such as sensor location, time, pixel coordinate, frame index, or voxel coordinate. GLSKF exploits this common representation by combining a covariance-regularized low-rank global component with a locally correlated residual component. The global component captures large-scale and long-range structure, while the local component captures short-range corrections. The covariance kernels are specified according to the geometry of each mode, such as graph- or distance-based kernels for traffic networks, spatial kernels for image and MRI coordinates, and spatiotemporal kernels for video data.

The generality of GLSKF therefore comes from its covariance-regularized tensor formulation, not from assuming that a single kernel family is optimal for all datasets. Kernel choice should reflect the dependence structure of the corresponding mode, and observed-validation is used to select kernel hyperparameters. Higher tensor order, larger mode sizes, and lower sampling rates generally make completion more challenging, since fewer observations are available relative to the number of unknown entries. GLSKF addresses this challenge computationally through matrix-free Kronecker/sparse/Toeplitz CG updates and statistically through complementary global--local regularization. However, we do not claim a universal recovery guarantee or generalization bound for arbitrary high-dimensional sparse tensors. Formal guarantees would require stronger assumptions on sampling, rank, incoherence, noise, covariance specification, and separation between global and local components, which are left for future work.

\subsection{Model Scalability} \label{subsec:time}
\begin{figure*}[!t]
\centering
\includegraphics[width=0.85\textwidth]{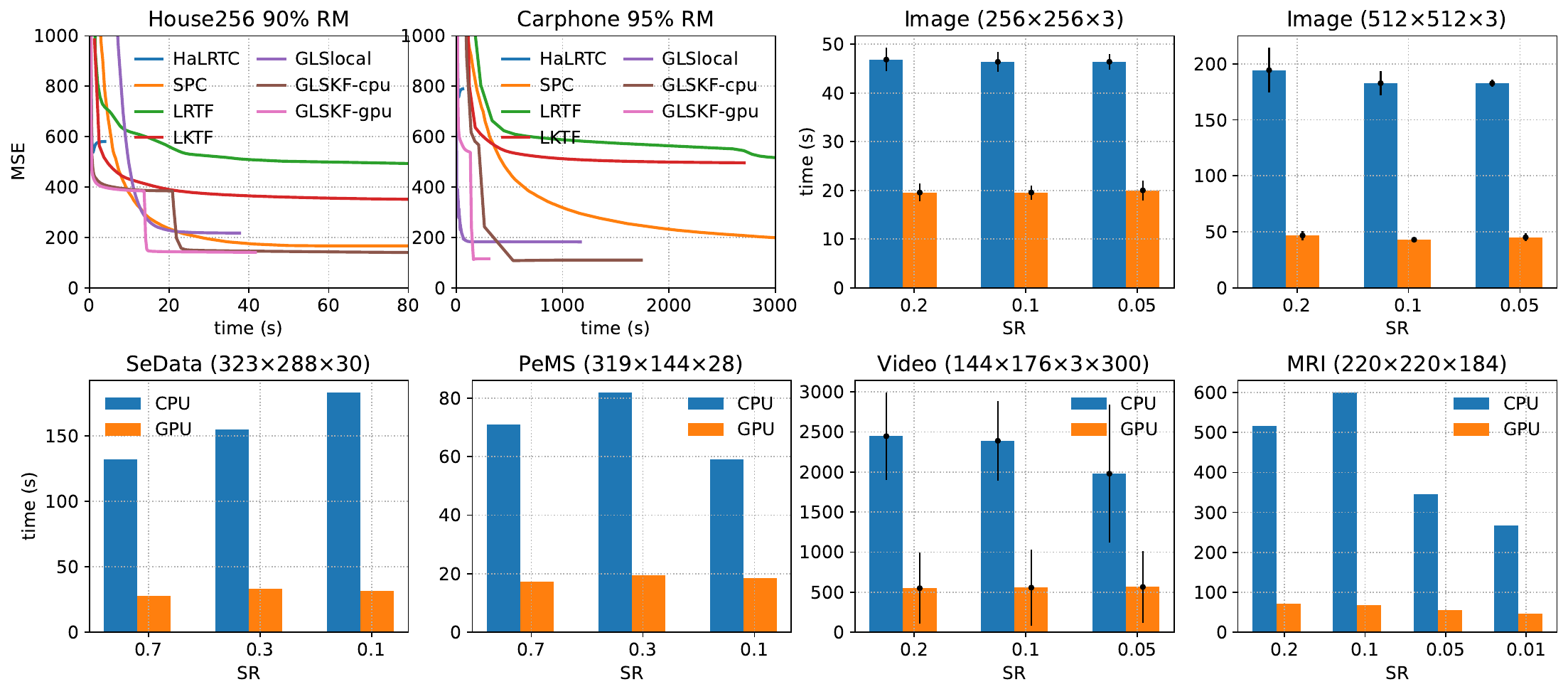}
\caption{Illustration of model scalability. We compare wall-clock time versus MSE for \textit{House256} 90\% RM and \textit{Carphone} 95\% RM and show CPU and GPU runtime of GLSKF across experimental scenarios. For datasets with multiple samples (color images and videos), we show the mean runtime along with std.}
\label{fig:image-time}
\end{figure*}

A main advantage of the proposed GLSKF framework is the scalability to large multidimensional datasets. The overall computational efficiency arises from (i) the global component exploits latent-factor dimensionality reduction, and (ii) the local component uses mask-aware optimization together with sparse and structured (Kronecker and Toeplitz) MVMs. In addition, practical GPU acceleration further improves efficiency. In Figure~\ref{fig:image-time}, we compare wall-clock time versus mean squared error (MSE) for GLSKF and competing methods on two representative tasks: 90\% RM color image inpainting for \textit{House256} ($256\times256\times3$) and 95\% RM video completion for \textit{Carphone} ($144\times176\times3\times300$). All methods use the same stopping criterion. As can be seen, GLSKF (CPU and GPU) first updates only the global component; after $K_0$ iterations, the local component is incorporated, which triggers a sharp MSE drop: e.g., around 15 seconds for \textit{House256} and 130 seconds for \textit{Carphone} with GLSKF-gpu. As the problem size increases from image to video, the advantages of GLSKF become clearer: it attains the lowest error within comparable runtime and can be substantially more cost-efficient than baselines such as SPC. We provide the CPU and GPU GLSKF runtime across the applied experiments in Figure~\ref{fig:image-time} as well. Overall, the GPU implementation offers approximately 4-5$\times$ speedup over CPU, which is particularly beneficial for large-scale settings such as color video completion. Notably, GLSKF recovers multidimensional data in seconds under high missing ratios such as 90\% or 95\% missingness.

\section{Conclusion} \label{sec:Conclusion}
In this paper, we propose GLSKF, a complementary covariance-regularized tensor factorization model with a GLS loss for multidimensional data completion. By combining a smoothed low-rank global decomposition with a locally correlated residual and enforcing structure through covariance-norm regularization, GLSKF captures both long-range, low-frequency patterns and short-scale, high-frequency variations. We exploit sparse Toeplitz covariance matrices for the local module and projection/selection operators that preserve Kronecker structure under missing data, enabling fast Kronecker MVMs and efficient CG-based updates. Experiments on four real-world datasets demonstrate that GLSKF consistently outperforms state-of-the-art low-rank baselines in both effectiveness (reconstruction accuracy) and efficiency (runtime). 
There are several directions for future research. For the global low-rank component, the current CP decomposition can be extended to more expressive tensor models, such as Tucker, tensor-ring, or tensor-train decompositions. For the local component, the separable covariance structure can be generalized to nonseparable covariances, for example through covariance constructions based on spectral representations or sums of separable kernels. In addition, the GLSKF framework can be adapted to related multidimensional data tasks, including denoising and robust recovery under corruptions, further broadening the applicability.

\section*{Acknowledgments}
This work was supported by the Natural Sciences and Engineering Research Council (NSERC) of Canada Discovery Grant (RGPIN-2025-04479).

\bibliographystyle{IEEEtran}
\bibliography{ref}

\vfill

\clearpage
\pagenumbering{arabic}
\setcounter{page}{1}

\setcounter{section}{0}
\setcounter{subsection}{0}
\setcounter{figure}{0}
\setcounter{table}{0}
\setcounter{equation}{0}
\renewcommand{\thesection}{\Roman{section}}
\renewcommand{\thesubsection}{\thesection-\Alph{subsection}}
\renewcommand{\thefigure}{S\arabic{figure}}
\renewcommand{\thetable}{S\arabic{table}}
\renewcommand{\theequation}{S\arabic{equation}}
\setcounter{algorithm}{0}
\renewcommand{\thealgorithm}{S\arabic{algorithm}}

\makeatletter
\renewcommand{\theHsection}{supp.\arabic{section}}
\renewcommand{\theHsubsection}
  {supp.\arabic{section}.\arabic{subsection}}
\renewcommand{\theHsubsubsection}
  {supp.\arabic{section}.\arabic{subsection}.\arabic{subsubsection}}
\renewcommand{\theHfigure}{supp.\arabic{figure}}
\renewcommand{\theHtable}{supp.\arabic{table}}
\renewcommand{\theHequation}{supp.\arabic{equation}}
\makeatother

\twocolumn[
\begin{center}
{\LARGE Supplementary Material for: Generalized Least Squares Kernelized Tensor Factorization\par}
\vspace{6pt}
\end{center}
]

\startcontents[supp]
\section*{Table of Contents}
\printcontents[supp]{}{1}[3]{}

\section{Supplementary Comparison Table}
A detailed comparison of global--local frameworks is provided in Table~\ref{tab:ref-sup}.

\begin{table*}[t!]
\centering
\caption{Detailed comparison of representative global-local modeling frameworks.}
\begin{tabular}{p{3cm}p{3cm}p{3cm}p{3cm}p{4cm}}
\toprule
Method & Global component & Local component & Estimation & Relation to GLSKF \\
\midrule
Spatial full-scale approximations \cite{sang2012full, katzfuss2013bayesian}
& Reduced-rank spatial covariance
& Tapered covariance for fine-scale residuals
& Bayesian or likelihood-based inference
& Captures long- and short-range spatial dependence, but is mainly designed for spatial processes rather than incomplete tensors. \\
\cmidrule{2-5}
Functional matrix completion \cite{descary2019functional, masak2022random}
& Low-rank functional or matrix structure
& Banded covariance for local dependence
& Likelihood-based or estimating-equation approaches
& Models local dependence in functional data, but is less directly applicable to high-order tensors with general missing patterns. \\
\cmidrule{2-5}
Generalized low-rank modeling \cite{allen2014generalized}
& Covariance-regularized low-rank matrix factors
& Correlated error structure integrated out
& Optimization-based estimation
& Uses GLS-type covariance modeling, but does not explicitly recover a separate local residual component. \\
\cmidrule{2-5}
Global-local deep prior \cite{zhao2021tensor}
& Low-rank tensor representation
& Plug-and-play neural local prior
& Alternating or PnP optimization
& Provides flexible local modeling, but the neural prior is less transparent than covariance-based residual modeling. \\
\cmidrule{2-5}
BCKL \cite{lei2022bckl}
& Bayesian kernelized low-rank tensor factorization
& Tapered covariance-based local residual
& Full Bayesian MCMC
& Closely related global-local structure, but computationally more expensive due to posterior sampling. \\
\cmidrule{2-5}
GLSKF
& Kernelized low-rank tensor factorization
& Explicit locally correlated residual
& Deterministic GLS/MAP estimation
& Designed for scalable incomplete tensor recovery with covariance-based global-local modeling. \\
\bottomrule
\end{tabular}
\label{tab:ref-sup}
\end{table*}

\section{Methodology Details} \label{appendix-sec:method}
\subsection{Conjugate Gradient Algorithms} \label{appendix-subsec:CG}
The conjugate gradient (CG) algorithms used to update the global latent factor matrices and the local component are presented in Algorithms~\ref{Alg:CG-Ud} and~\ref{Alg:Rupdate}, respectively.

\begin{algorithm}[t!]
\caption{CG update for $\boldsymbol{U}^{(d),(k+1)}$ with Eq.~\eqref{Eq:Udtranspose}.}
\label{Alg:CG-Ud}
\begin{algorithmic}[1]
\renewcommand{\algorithmicrequire}{\textbf{Input:}}
\renewcommand{\algorithmicensure}{\textbf{Output:}}
\REQUIRE $\bigl\{\boldsymbol{U}^{(h),(k+1)}\bigr\}_{h=1}^{d-1}$, $\bigl\{\boldsymbol{U}^{(h),(k)}\bigr\}_{h=d}^D$, $\boldsymbol{\mathcal{G}}$, $\boldsymbol{\mathcal{O}}$, $\bigl(\boldsymbol{K}_{\boldsymbol{u}}^{(d)}\bigr)^{-1}$, maximum number of iterations $J_{\text{CG}}=1000$.
\ENSURE $\boldsymbol{U}^{(d),(k+1)}$.
\STATE Compute the vector right-hand-side the inverse, i.e., $\boldsymbol{b}=\boldsymbol{H}_{d}^{\top}\boldsymbol{O}_{d}'^{\top}\boldsymbol{O}_d'\operatorname{vec}\bigl(\boldsymbol{G}_{(d)}^{\top}\bigr)=\boldsymbol{H}_{d}^{\top}\operatorname{vec}\bigl(\left[\boldsymbol{\mathcal{G}}\circledast\boldsymbol{\mathcal{O}}\right]_{(d)}^{\top}\bigr)$;
\STATE Let $\boldsymbol{x}_0=\boldsymbol{u}^{(d),(k)}$, \(\boldsymbol{A}=\boldsymbol{A}_1+\boldsymbol{A}_2 \) as defined in Eq.~\eqref{Eq:Udtranspose};
\STATE Compute $\boldsymbol{A}\boldsymbol{x}_0$;
\STATE Compute the initial residual $\boldsymbol{e}_0=\boldsymbol{b}-\boldsymbol{A}\boldsymbol{x}_0$;
\STATE Let the direction $\boldsymbol{p}_1=\boldsymbol{e}_0$;
\FOR{$j=1$ {\bfseries to} $J_{\text{CG}}$}
\STATE Compute $\boldsymbol{A}\boldsymbol{p}_j=\boldsymbol{A}_1\boldsymbol{p}_j+\boldsymbol{A}_2\boldsymbol{p}_j$; 
\STATE Compute the step size $\alpha_j=\frac{\boldsymbol{e}_{j-1}^{\top}\boldsymbol{e}_{j-1}}{\boldsymbol{p}_j^{\top}\left(\boldsymbol{A}\boldsymbol{p}_j\right)}$;
\STATE Compute the updated solution $\boldsymbol{x}_j=\boldsymbol{x}_{j-1}+\alpha_j\boldsymbol{p}_j$;
\STATE Compute residual $\boldsymbol{e}_j=\boldsymbol{e}_{j-1}-\alpha_j\left(\boldsymbol{A}\boldsymbol{p}_j\right)$;
\IF{$\sqrt{\boldsymbol{e}_j^{\top}\boldsymbol{e}_j}<10^{-6}$}
\STATE {\bfseries{break}};
\ENDIF
\STATE Compute $\beta_j=\frac{\boldsymbol{e}_j^{\top}\boldsymbol{e}_j}{\boldsymbol{e}_{j-1}^{\top}\boldsymbol{e}_{j-1}}$;
\STATE Compute the direction $\boldsymbol{p}_{j+1}=\boldsymbol{e}_j+\beta_j\boldsymbol{p}_j$;
\ENDFOR
\STATE {\bfseries{return}} $\boldsymbol{U}^{(d),(k+1)}=\left(\operatorname{reshape}\left(\boldsymbol{x}_j, [R, I_d]\right)\right)^{\top}$.
\end{algorithmic}
\end{algorithm}

\begin{algorithm}[t!]
\caption{CG update for $\boldsymbol{\mathcal{R}}^{(k+1)}$ with Eq.~\eqref{Eq:rSolutionK}.}
\label{Alg:Rupdate}
\begin{algorithmic}[1]
\renewcommand{\algorithmicrequire}{\textbf{Input:}}
\renewcommand{\algorithmicensure}{\textbf{Output:}}
\REQUIRE $\boldsymbol{l}_{\Omega}^{(k+1)}$, $\bigl\{\boldsymbol{K}_{\boldsymbol{\mathcal{R}}}^{(d)}\bigr\}_{d=1}^D$, maximum number of iterations $J_{\text{CG}}=1000$.
\ENSURE $\boldsymbol{\mathcal{R}}^{(k+1)}$.

\STATE Let $\boldsymbol{b}=\boldsymbol{l}_{\Omega}^{(k+1)}\in\mathbb{R}^{|\Omega|}$;

\STATE Let $\boldsymbol{x}_0=\boldsymbol{z}^{(k)}\in\mathbb{R}^{|\Omega|}$,  $\boldsymbol{A}'=\gamma\boldsymbol{I}_{|\Omega|}+\boldsymbol{A}_3\in\mathbb{R}^{|\Omega|\times|\Omega|}$ with $\boldsymbol{A}_3=\boldsymbol{O}_1\boldsymbol{K}_{r}\boldsymbol{O}_1^{\top}$ as given in Eq.~\eqref{Eq:rSolutionK};

\STATE Compute $\boldsymbol{A}'\boldsymbol{x}_0=\gamma\boldsymbol{x}_0+\boldsymbol{A}_3\boldsymbol{x}_0$, where $\boldsymbol{A}_3\boldsymbol{x}_0$ is computed as $\boldsymbol{O}_1\boldsymbol{K}_{r}\boldsymbol{O}_1^{\top}\boldsymbol{x}_0$ from right to left via zero-padding: $\boldsymbol{O}_1^{\top}\boldsymbol{x}_0$, Kronecker MVM: $\boldsymbol{K}_{r}\times\left(\cdot\right)$, and slicing: $\boldsymbol{O}_1\times\left(\cdot\right)$;

\STATE Compute the initial residual $\boldsymbol{e}_0=\boldsymbol{b}-\boldsymbol{A}'\boldsymbol{x}_0\in\mathbb{R}^{|\Omega|}$;

\STATE Let the direction $\boldsymbol{p}_1=\boldsymbol{e}_0$;
\FOR{$j=1$ {\bfseries to} $J_{\text{CG}}$}

\STATE Compute $\boldsymbol{A}'\boldsymbol{p}_j=\gamma\boldsymbol{p}_j+\boldsymbol{A}_3\boldsymbol{p}_j$, where $\boldsymbol{A}_3\boldsymbol{p}_j$ is computed similarly as \(\boldsymbol{A}_3\boldsymbol{x}_0\);

\STATE Compute the step size $\alpha_j=\frac{\boldsymbol{e}_{j-1}^{\top}\boldsymbol{e}_{j-1}}{\boldsymbol{p}_j^{\top}\left(\boldsymbol{A}'\boldsymbol{p}_j\right)}$;

\STATE Compute the updated solution $\boldsymbol{x}_j=\boldsymbol{x}_{j-1}+\alpha_j\boldsymbol{p}_j$;
\STATE Compute residual $\boldsymbol{e}_j=\boldsymbol{e}_{j-1}-\alpha_j\left(\boldsymbol{A}'\boldsymbol{p}_j\right)$;
\IF{$\sqrt{\boldsymbol{e}_j^{\top}\boldsymbol{e}_j}<10^{-6}$}
\STATE {\bfseries{break}};
\ENDIF
\STATE Compute $\beta_j=\frac{\boldsymbol{e}_j^{\top}\boldsymbol{e}_j}{\boldsymbol{e}_{j-1}^{\top}\boldsymbol{e}_{j-1}}$;
\STATE Compute the direction $\boldsymbol{p}_{j+1}=\boldsymbol{e}_j+\beta_j\boldsymbol{p}_j$;
\ENDFOR
\STATE $\boldsymbol{z}^{(k+1)}=\boldsymbol{x}_j$;
\STATE Compute $\boldsymbol{r}^{(k+1)}=\boldsymbol{K}_{\boldsymbol{r}}\boldsymbol{O}_1^{\top}\boldsymbol{z}^{(k+1)}$ by Kronecker MVM;
\STATE {\bfseries{return}} $\boldsymbol{\mathcal{R}}^{(k+1)}=\operatorname{reshape}\left(\boldsymbol{r}^{(k+1)},\text{size}\left(\boldsymbol{\mathcal{R}}\right)\right)$.
\end{algorithmic}
\end{algorithm}

\subsection{Toeplitz Structured MVM with FFT} \label{appendix-subsec:Toeplitz}
Given that the last-mode covariance $\boldsymbol{K}_{\boldsymbol{\mathcal{R}}}^{(D)}$, e.g., the channel mode in color image or the day mode in traffic data, is typically either identity (independent) or a dense empirical covariance (non-Toeplitz), we assume only $\big\{\boldsymbol{K}_{\boldsymbol{\mathcal{R}}}^{(d)}\big\}_{d=1}^{D-1}$ are Toeplitz. For $\boldsymbol{x}'\in\mathbb{R}^N$, we compute $\boldsymbol{K}_{\boldsymbol{r}}\boldsymbol{x}'$ as
\[
\boldsymbol{K}_{\boldsymbol{r}}\boldsymbol{x}'=\operatorname{vec}\Big(\big(\bigotimes_{d=D-1}^1\boldsymbol{K}_{\boldsymbol{\mathcal{R}}}^{(d)}\big)\boldsymbol{X}\boldsymbol{K}_{\boldsymbol{\mathcal{R}}}^{(D)}\Big),
\]
where $\boldsymbol{X}=\operatorname{reshape}\big(\boldsymbol{x}',[\frac{N}{I_D}, I_D]\big)$. We accelerate $\big(\bigotimes_{d=D-1}^1\boldsymbol{K}_{\boldsymbol{\mathcal{R}}}^{(d)}\big)\boldsymbol{X}$ by circulant embedding. For a Toeplitz $\boldsymbol{K}_{\boldsymbol{\mathcal{R}}}^{(d)}\in\mathbb{R}^{I_d\times I_d}$ with first row $[k_0,k_1,\dots,k_{I_d-2},k_{I_d-1}]$, we embed it into the $(2I_d-2)\times(2I_d-2)$ circulant matrix $\boldsymbol{C}_{\boldsymbol{\mathcal{R}}}^{(d)}$ with the vector representation:
\begin{equation}
\boldsymbol{c}_{\boldsymbol{\mathcal{R}}}^{(d)}=[k_0,k_1,\\\ldots,k_{I_d-1},k_{I_d-2},k_{I_d-3},\ldots,k_1]^{\top}\in\mathbb{R}^{2I_d-2}.
\end{equation}
We represent $\boldsymbol{x}'$ as a $D$th-order tensor $\boldsymbol{\mathcal{X}}\in\mathbb{R}^{I_1\times\cdots\times I_D}$, and zero-pad to $\boldsymbol{\mathcal{X}}_c\in\mathbb{R}^{(2I_1-2)\times(2I_2-2)\times\cdots\times I_D}$ with $\boldsymbol{\mathcal{X}}_c\left(1:I_1,\dots,1:I_{D-1},:\right)=\boldsymbol{\mathcal{X}}$. We then perform sequential 1D FFT-based convolutions along each Toeplitz axis using $\big\{\boldsymbol{c}_{\boldsymbol{\mathcal{R}}}^{(d)}\big\}_{d=1}^{D-1}$, followed by cropping back to the original size. Lastly, we multiply by $\boldsymbol{K}_{\boldsymbol{\mathcal{R}}}^{(D)}$ on the right. In this way, the Kronecker MVM cost is reduced from $\mathcal{O}\left(\sum_d\mathrm{nnz}\bigl(\boldsymbol{K}_{\boldsymbol{\mathcal{R}}}^{(d)}\bigr) \frac{N}{I_d}\right)$ to $\mathcal{O}\big(\sum_{d=1}^{D-1}(2I_d-2)\log(2I_d-2) + \mathrm{nnz}\bigl(\boldsymbol{K}_{\boldsymbol{\mathcal{R}}}^{(D)}\bigr) \frac{N}{I_D}\big)$. If $\boldsymbol{K}_{\boldsymbol{\mathcal{R}}}^{(D)}=\boldsymbol{I}_{I_D}$, the last term $\mathrm{nnz}\bigl(\boldsymbol{K}_{\boldsymbol{\mathcal{R}}}^{(D)}\bigr) \frac{N}{I_D}$ vanishes.

\section{Proofs} \label{appendix-sec:proof}
\subsection{Proof of Remark~\ref{remark:covnorm}} \label{appendix-proofRemark1}
\begin{proof}
Let $\boldsymbol{Q}\succ\boldsymbol{0}$ be symmetric with eigen-decomposition $\boldsymbol{Q}=\boldsymbol{U}\operatorname{diag}\left(\lambda_1,\dots,\lambda_n\right)\boldsymbol{U}^{\top}$, where $\{\lambda_i\}_{i=1}^n\geq 0$. The inverse of $\boldsymbol{Q}$ is $\boldsymbol{Q}^{-1}=\boldsymbol{U}\operatorname{diag}\left(\lambda_1^{-1},\dots,\lambda_n^{-1}\right)\boldsymbol{U}^{\top}$, where $\lambda_i^{-1}=\frac{1}{\lambda_i}$ if $\lambda_i>0$ and 0 otherwise. Define $\boldsymbol{K}=\boldsymbol{Q}^{-1}$, then $\boldsymbol{K}\succ\boldsymbol{0}$. Thus, $\boldsymbol{K}$ is a valid covariance matrix. For symmetric positive definite matrices, $\left(\boldsymbol{Q}^{-1}\right)^{-1}=\boldsymbol{Q}$, hence $\boldsymbol{K}^{-1}=\boldsymbol{Q}$. Therefore, the covariance norm can be written as $\left\|\boldsymbol{x}\right\|_{\boldsymbol{K}}^2=\boldsymbol{x}^{\top}\boldsymbol{K}^{-1}\boldsymbol{x}=\boldsymbol{x}^{\top}\boldsymbol{Q}\boldsymbol{x}$.
\end{proof}

\subsection{Proof of Remark~\ref{remark:GLSKF}} \label{appendix-proofRemark2}

\begin{proof}
Let
$\boldsymbol{e}_{\Omega}=\operatorname{vec}\!\big((\boldsymbol{\mathcal{Y}}-\boldsymbol{\mathcal{M}}-\boldsymbol{\mathcal{R}})_{\Omega}\big)$.
Under the likelihood $\boldsymbol{e}_{\Omega}\sim\mathcal{N}(\boldsymbol{0},\boldsymbol{I}_{|\Omega|})$, one can write:
\[
-\log p(\boldsymbol{\mathcal{Y}}_{\Omega}\mid \boldsymbol{\mathcal{M}},\boldsymbol{\mathcal{R}})
\propto \frac{1}{2}\left\|\boldsymbol{e}_{\Omega}\right\|_2^2
= \frac{1}{2}\left\|(\boldsymbol{\mathcal{Y}}-\boldsymbol{\mathcal{M}}-\boldsymbol{\mathcal{R}})_{\Omega}\right\|_F^2.
\]
For each mode $d$, the prior
$\operatorname{vec}(\boldsymbol{U}^{(d)})\sim\mathcal{N}(\boldsymbol{0},\rho^{-1}\boldsymbol{K}_{\boldsymbol{U}}^{(d)})$
gives
\[
\begin{aligned}
&-\log p(\boldsymbol{U}^{(d)})
\propto \frac{\rho}{2}\operatorname{vec}(\boldsymbol{U}^{(d)})^{\top}(\boldsymbol{K}_{\boldsymbol{U}}^{(d)})^{-1}\operatorname{vec}(\boldsymbol{U}^{(d)}) \\
& \quad = \frac{\rho}{2}\|\boldsymbol{U}^{(d)}\|_{\boldsymbol{K}_{\boldsymbol{U}}^{(d)}}^{2},
\end{aligned}
\]
where we used Definition~\ref{def:covnorm}. Similarly, the residual prior
$\operatorname{vec}(\boldsymbol{\mathcal{R}})\sim\mathcal{N}(\boldsymbol{0},\gamma^{-1}\boldsymbol{K}_{\boldsymbol{\mathcal{R}}})$
implies
\[
-\log p(\boldsymbol{\mathcal{R}})
\propto \frac{\gamma}{2}\|\boldsymbol{\mathcal{R}}\|_{\boldsymbol{K}_{\boldsymbol{\mathcal{R}}}}^{2}.
\]
Summing the negative log-likelihood and negative log-priors yields the negative log-posterior:
\[
\begin{aligned}
&-\log p(\{\boldsymbol{U}^{(d)}\},\boldsymbol{\mathcal{R}}\mid \boldsymbol{\mathcal{Y}}_{\Omega})\\
&\quad\propto
\frac{1}{2}\|(\boldsymbol{\mathcal{Y}}-\boldsymbol{\mathcal{M}}-\boldsymbol{\mathcal{R}})_{\Omega}\|_F^2
+\frac{\rho}{2}\sum_{d=1}^{D}\|\boldsymbol{U}^{(d)}\|_{\boldsymbol{K}_{\boldsymbol{U}}^{(d)}}^{2}
+\frac{\gamma}{2}\|\boldsymbol{\mathcal{R}}\|_{\boldsymbol{K}_{\boldsymbol{\mathcal{R}}}}^{2},
\end{aligned}
\]
which matches the \textsc{GLSKF} objective~\eqref{Eq:objAll} up to an additive constant.

\end{proof}

\subsection{Proof of Proposition~\ref{prop:convergence}} \label{appendix-proofProposition1}
Here we establish a monotone-descent and stationarity result for the
\emph{fixed-covariance} GLSKF objective. That is, throughout the ALS iterations,
the covariance matrices
$\{\boldsymbol K_u^{(d)}\}_{d=1}^D$ and
$\{\boldsymbol K_{\mathcal R}^{(d)}\}_{d=1}^D$
are assumed fixed.

Consider the GLSKF objective
\[
\begin{aligned}
\mathcal F&\big(\{\boldsymbol U^{(d)}\}_{d=1}^D,\boldsymbol{\mathcal R}\big)\\
&=
\frac{1}{2}
\big\|
(\boldsymbol{\mathcal Y}-\boldsymbol{\mathcal M}-\boldsymbol{\mathcal R})_{\Omega}
\big\|_F^2
+
\frac{\rho}{2}
\sum_{d=1}^D
\big\|
\boldsymbol U^{(d)}
\big\|_{\boldsymbol K_U^{(d)}}^2
+
\frac{\gamma}{2}
\big\|
\boldsymbol{\mathcal R}
\big\|_{\boldsymbol K_{\mathcal R}}^2,
\end{aligned}
\]
where $\rho,\gamma>0$,
$\boldsymbol K_U^{(d)}=\boldsymbol I_R\otimes \boldsymbol K_u^{(d)}$,
and $\boldsymbol K_{\mathcal R}=\bigotimes_{d=D}^1 \boldsymbol K_{\mathcal R}^{(d)}$.
Assume that all covariance matrices
$\{\boldsymbol K_u^{(d)}\}_{d=1}^D$ and
$\{\boldsymbol K_{\mathcal R}^{(d)}\}_{d=1}^D$
are fixed and symmetric positive definite.

Let
\[
\boldsymbol\theta
:=
\Big(
\operatorname{vec}(\boldsymbol U^{(1)}),\dots,
\operatorname{vec}(\boldsymbol U^{(D)}),
\operatorname{vec}(\boldsymbol{\mathcal R})
\Big).
\]
At ALS sweep $k$, let $q_j^k(x)$ denote the quadratic subproblem associated with the
$j$th block. Suppose that each block is updated either exactly or by CG started from
the current block iterate $x_j^k$, producing $x_j^{k+1}$ with residual
\[
\boldsymbol r_j^{\,k+1}:=\nabla q_j^k(x_j^{k+1})
\]
satisfying
\[
\|\boldsymbol r_j^{\,k+1}\|_2
\le
\eta_k \|x_j^{k+1}-x_j^k\|_2,
\quad
0\le \eta_k \le \bar\eta < \mu/2,
\quad
\eta_k \to 0,
\]
where $\mu:=\min_j \mu_j>0$, and $\mu_j$ is a uniform lower bound on the smallest
eigenvalue of the Hessian of the $j$th block subproblem.

Then:

(i) $\mathcal F(\boldsymbol\theta^{k+1})\le \mathcal F(\boldsymbol\theta^k)$ for all $k$;

(ii) the sequence $\{\mathcal F(\boldsymbol\theta^k)\}_{k\ge 0}$ converges;

(iii) the iterate sequence $\{\boldsymbol\theta^k\}_{k\ge 0}$ is bounded and satisfies
\[
\|\boldsymbol\theta^{k+1}-\boldsymbol\theta^k\|_2 \to 0;
\]

(iv) every accumulation point of $\{\boldsymbol\theta^k\}_{k\ge 0}$ is a stationary point of
$\mathcal F$.

\begin{proof}
We write
\[
\begin{aligned}
\mathcal F&\big(\{\boldsymbol U^{(d)}\}_{d=1}^D,\boldsymbol{\mathcal R}\big) \\
&=
\frac{1}{2}
\big\|
(\boldsymbol{\mathcal Y}-\boldsymbol{\mathcal M}-\boldsymbol{\mathcal R})_{\Omega}
\big\|_F^2
+
\frac{\rho}{2}
\sum_{d=1}^D
\big\|
\boldsymbol U^{(d)}
\big\|_{\boldsymbol K_U^{(d)}}^2
+
\frac{\gamma}{2}
\big\|
\boldsymbol{\mathcal R}
\big\|_{\boldsymbol K_{\mathcal R}}^2.
\end{aligned}
\]
Since $\rho,\gamma>0$ and all covariance matrices are symmetric positive definite,
all covariance-norm terms are well defined and nonnegative. Hence
$\mathcal F \ge 0$, so the objective is bounded below.

We partition the variables into $D+1$ blocks:
\[
x_1=\operatorname{vec}(\boldsymbol U^{(1)}),\dots,
x_D=\operatorname{vec}(\boldsymbol U^{(D)}),
\quad
x_{D+1}=\operatorname{vec}(\boldsymbol{\mathcal R}).
\]
At ALS sweep $k$, when all blocks except the $j$th are fixed, the objective restricted
to block $x_j$ becomes a quadratic function
\[
q_j^k(x)=\frac{1}{2}x^\top \boldsymbol H_j^k x-(\boldsymbol b_j^k)^\top x+c_j^k,
\]
where $\boldsymbol H_j^k$ is the corresponding block Hessian.

For the factor block $x_j=\operatorname{vec}(\boldsymbol U^{(d)})$, Eq.~(13) shows that
the Hessian is
\[
\boldsymbol H_d^k
=
\boldsymbol H_d^\top \boldsymbol O_d^{\prime\top}\boldsymbol O_d' \boldsymbol H_d
+
\rho (\boldsymbol K_u^{(d)})^{-1}\otimes \boldsymbol I_R,
\]
where the first term is positive semidefinite. Therefore
\[
\boldsymbol H_d^k
\succeq
\rho\,\lambda_{\min}\big((\boldsymbol K_u^{(d)})^{-1}\big)\boldsymbol I.
\]

For the residual block in primal form, Eq.~(16) gives the Hessian
\[
\boldsymbol H_{\mathcal R}^k
=
\boldsymbol O_1^\top \boldsymbol O_1 + \gamma \boldsymbol K_{\mathcal R}^{-1},
\]
hence
\[
\boldsymbol H_{\mathcal R}^k
\succeq
\gamma\,\lambda_{\min}(\boldsymbol K_{\mathcal R}^{-1})\boldsymbol I.
\]
If the residual is solved in dual form as in Eq.~(18), then the Hessian is
\[
\widetilde{\boldsymbol H}_{\mathcal R}^k
=
\boldsymbol O_1 \boldsymbol K_{\mathcal R}\boldsymbol O_1^\top
+
\gamma \boldsymbol I_{|\Omega|}
\succeq
\gamma \boldsymbol I_{|\Omega|}.
\]
Thus, every block subproblem is strongly convex. Let
\[
\mu:=\min_j \mu_j>0
\]
be a common lower bound on the smallest eigenvalues of all block Hessians.

Now fix one block $j$ at sweep $k$, and define the step
\[
\boldsymbol s_j^k:=x_j^{k+1}-x_j^k,
\qquad
\boldsymbol r_j^{\,k+1}:=\nabla q_j^k(x_j^{k+1})
=
\boldsymbol H_j^k x_j^{k+1}-\boldsymbol b_j^k.
\]
Since $q_j^k$ is quadratic, we have the exact identity
\[
q_j^k(x_j^k)-q_j^k(x_j^{k+1})
=
\frac{1}{2}(\boldsymbol s_j^k)^\top \boldsymbol H_j^k \boldsymbol s_j^k
-
(\boldsymbol r_j^{\,k+1})^\top \boldsymbol s_j^k.
\]
Using $\boldsymbol H_j^k \succeq \mu_j \boldsymbol I$ and the residual condition,
\[
q_j^k(x_j^k)-q_j^k(x_j^{k+1})
\ge
\frac{\mu_j}{2}\|\boldsymbol s_j^k\|_2^2
-
\|\boldsymbol r_j^{\,k+1}\|_2\|\boldsymbol s_j^k\|_2
\ge
\left(\frac{\mu_j}{2}-\eta_k\right)\|\boldsymbol s_j^k\|_2^2.
\]
Since $\eta_k\le \bar\eta < \mu/2 \le \mu_j/2$, it follows that
\[
q_j^k(x_j^k)-q_j^k(x_j^{k+1}) \ge 0.
\]
Therefore each block update is non-increasing. Because one ALS sweep updates the
blocks sequentially, we obtain
\[
\mathcal F(\boldsymbol\theta^{k+1})\le \mathcal F(\boldsymbol\theta^k),
\qquad \forall k.
\]
This proves monotone descent.

Since $\mathcal F\ge 0$ and is monotone non-increasing, the monotone convergence
theorem implies that there exists $\mathcal F_\star\ge 0$ such that
\[
\mathcal F(\boldsymbol\theta^k)\to \mathcal F_\star.
\]

Next, because the covariance matrices are positive definite,
\[
\begin{aligned}
\|\boldsymbol U^{(d)}\|_{\boldsymbol K_U^{(d)}}^2
&=
\operatorname{vec}(\boldsymbol U^{(d)})^\top
\Big(
(\boldsymbol K_u^{(d)})^{-1}\otimes \boldsymbol I_R
\Big)
\operatorname{vec}(\boldsymbol U^{(d)})\\
&\ge
\lambda_{\min}\big((\boldsymbol K_u^{(d)})^{-1}\big)
\|\boldsymbol U^{(d)}\|_F^2,
\end{aligned}
\]
and similarly
\[
\|\boldsymbol{\mathcal R}\|_{\boldsymbol K_{\mathcal R}}^2
=
\operatorname{vec}(\boldsymbol{\mathcal R})^\top
\boldsymbol K_{\mathcal R}^{-1}
\operatorname{vec}(\boldsymbol{\mathcal R})
\ge
\lambda_{\min}(\boldsymbol K_{\mathcal R}^{-1})
\|\boldsymbol{\mathcal R}\|_F^2.
\]
Hence
\[
\mathcal F(\boldsymbol\theta)
\ge
\frac{\rho}{2}\sum_{d=1}^D
\lambda_{\min}\big((\boldsymbol K_u^{(d)})^{-1}\big)
\|\boldsymbol U^{(d)}\|_F^2
+
\frac{\gamma}{2}
\lambda_{\min}(\boldsymbol K_{\mathcal R}^{-1})
\|\boldsymbol{\mathcal R}\|_F^2.
\]
Therefore every sublevel set of $\mathcal F$ is bounded. Since
\[
\mathcal F(\boldsymbol\theta^k)\le \mathcal F(\boldsymbol\theta^0),
\]
all iterates lie in a bounded sublevel set, so $\{\boldsymbol\theta^k\}$ is bounded.

Summing the blockwise descent bound over all blocks and all iterations yields
\[
\sum_{k=0}^\infty \sum_{j=1}^{D+1}
\left(\frac{\mu}{2}-\bar\eta\right)
\|\boldsymbol s_j^k\|_2^2
\le
\mathcal F(\boldsymbol\theta^0)-\mathcal F_\star
<\infty.
\]
Hence
\[
\|\boldsymbol s_j^k\|_2 \to 0,
\qquad j=1,\dots,D+1.
\]
Consequently,
\[
\|\boldsymbol\theta^{k+1}-\boldsymbol\theta^k\|_2 \to 0.
\]

Because $\{\boldsymbol\theta^k\}$ is bounded, it has at least one accumulation point.
Let $\bar{\boldsymbol\theta}$ be any accumulation point, and let
$\{\boldsymbol\theta^{k_\ell}\}$ be a subsequence such that
\[
\boldsymbol\theta^{k_\ell}\to \bar{\boldsymbol\theta}.
\]
Since successive differences vanish, all intermediate block-updated states within sweep
$k_\ell$ converge to the same limit $\bar{\boldsymbol\theta}$.

For each block $j$, the gradient of the quadratic subproblem at the updated point is
exactly the CG residual:
\[
\nabla q_j^{k_\ell}(x_j^{k_\ell+1})
=
\boldsymbol r_j^{\,k_\ell+1}.
\]
By the residual condition and the vanishing-step property,
\[
\|\boldsymbol r_j^{\,k_\ell+1}\|_2
\le
\eta_{k_\ell}\|\boldsymbol s_j^{k_\ell}\|_2 \to 0.
\]
Since $\mathcal F$ is continuously differentiable and the block gradient of $q_j^k$
coincides with the corresponding partial gradient of $\mathcal F$, passing to the limit
gives
\[
\nabla_{x_j}\mathcal F(\bar{\boldsymbol\theta})=\boldsymbol 0,
\qquad
j=1,\dots,D+1.
\]
Therefore
\[
\nabla \mathcal F(\bar{\boldsymbol\theta})=\boldsymbol 0,
\]
so every accumulation point is a stationary point of $\mathcal F$.
This completes the proof.
\qed

\paragraph*{Remark}
Proposition~\ref{prop:convergence} is stated for the fixed-covariance GLSKF
objective, namely when
$\{\boldsymbol K_u^{(d)}\}_{d=1}^D$
and
$\{\boldsymbol K_{\mathcal R}^{(d)}\}_{d=1}^D$
are held fixed during the ALS iterations.
The empirical plug-in update of $\boldsymbol K_{\mathcal R}^{(D)}$ from the current
residual estimate in Section~4.3 is a practical heuristic and is not covered by the
formal convergence theorem.
A rigorous treatment of the adaptive-covariance variant would require either
(i) an outer-loop scheme that freezes the covariance within each inner ALS stage, or
(ii) an acceptance rule ensuring that the covariance update does not increase the
objective.
\end{proof}

\section{GLSKF Parameter Selection} \label{appendix-sec:selection}
Kernel hyperparameters and regularization parameters are selected in two separate stages. The kernel hyperparameters include the length-scales ${\ell_{\boldsymbol{u}}^{(d)}}$ for the global latent factors $\boldsymbol{U}^{(d)}$ and the length-scales and tapering ranges $\{\ell_{\boldsymbol{r}}^{(d)},\lambda_d\}$ for mode-$d$ of the local residual component $\boldsymbol{\mathcal{R}}$. The regularization parameters are the global and local penalty weights $\rho$ and $\gamma$.

In the second stage, after the covariance matrices are fixed, the regularization parameters $\rho$ and $\gamma$ are selected by validation on observed entries. These parameters control the relative strengths of the global and local regularization penalties in GLSKF. After selecting $\rho$ and $\gamma$, the final GLSKF model is refitted using all available observed entries. No ground-truth missing entries are used in either stage of parameter selection.

This two-stage specification and validation procedure is related to standard covariogram and variogram fitting ideas in spatial statistics \cite{cressie1985fitting,cressie2015statistics}, while the subsequent selection of regularization weights by observed-entry validation follows common model-selection practice for kernel and Gaussian-process models \cite{williams2006gaussian}.

\subsection{Selection of Kernel Hyperparameters} \label{supp:kernelHypeSelection}
In the first stage, the kernel hyperparameters are specified according to prior knowledge of the application and the expected correlation scales along each dimension. Specifically, the global length-scales $\ell_{\boldsymbol{u}}^{(d)}$ are chosen to represent medium-to-long-range smooth variation, whereas the local residual length-scales $\ell_{\boldsymbol{r}}^{(d)}$ and tapering ranges $\lambda_d$ are chosen to represent short-range local dependence. We fix the kernel variances to one, and control the relative strengths of the global and local regularization penalties by $\rho$ and $\gamma$.

To support the pre-specification protocol for kernel hyperparameters, we provide an empirical validation on the traffic speed datasets by fitting associated empirical correlation profiles computed from the observed entries. This empirical validation is used to examine whether the pre-specified kernel hyperparameters are consistent with the observed spatial or temporal correlation patterns. We do not use this empirical estimation as an additional optimization step for tuning GLSKF, and it does not require latent factor warm starts, pre-trained models, or ground-truth missing values.

Given an incomplete tensor $\boldsymbol{\mathcal Y}\in\mathbb{R}^{I_1\times\cdots\times I_D}$ with observed index set $\Omega$, we compute empirical directional correlations for each mode $d\in\{1,\dots,D\}$ using observed entry pairs separated by different lags along that mode. For a multi-index $\boldsymbol i=(i_1,\ldots,i_D)$, let $y_{\boldsymbol i}$ denote the corresponding tensor entry. For mode $d$ and lag $h$, $\mathcal P_d(h)$ contains all observed entry pairs that differ by $h$ only along mode $d$, while sharing the same indices in all other modes. In practice, $\mathcal P_d(h)$ is obtained by scanning the observed entries and retaining pairs $(\boldsymbol i,\boldsymbol j)$ such that $j_d=i_d+h$ and $j_{d'}=i_{d'}$ for all $d'\neq d$, provided that both $\boldsymbol i$ and $\boldsymbol j$ belong to the observed set $\Omega$.

Formally, for mode $d$, define
\[
\mathcal P_d(h)
=
\left\{
(\boldsymbol i,\boldsymbol j):
\boldsymbol i,\boldsymbol j\in\Omega,\ 
j_d=i_d+h,\ 
i_{d'}=j_{d'}\ \text{for } d'\neq d
\right\}.
\]
The empirical directional covariance is computed as
\[
\hat C_d(h)
=
\frac{1}{|\mathcal P_d(h)|}
\sum_{(\boldsymbol i,\boldsymbol j)\in\mathcal P_d(h)}
\big(y_{\boldsymbol i}-\hat\mu\big)
\big(y_{\boldsymbol j}-\hat\mu\big),
\]
where $\hat\mu$ is the mean of the observed entries. The corresponding empirical directional correlation is
\[
\hat\rho_d(h)
=
\frac{\hat C_d(h)}{\hat C_d(0)}.
\]
When multiple channels or repeated slices are available, empirical correlations are computed separately and then pooled using the numbers of observed pairs as weights.

Given a parametric kernel correlation function $\rho(h;\ell)$, an empirical reference length-scale is obtained by weighted least squares:
\[
\ell_d^{\mathrm{emp}}
=
\arg\min_{\ell>0}
\sum_{h\in\mathcal H_d}
\omega_h
\left[
\hat\rho_d(h)-\rho(h;\ell)
\right]^2,
\quad
\omega_h=|\mathcal P_d(h)|,
\]
where $\mathcal H_d$ denotes the set of lags used for fitting. For example, for the Mat\'ern $3/2$ kernel, the correlation function is
\[
\rho(h;\ell)
=
\left(1+\sqrt{3}\frac{h}{\ell}\right)
\exp\left(-\sqrt{3}\frac{h}{\ell}\right).
\]
For the global component, $\mathcal H_d$ is chosen to contain medium-to-long-range lags, so that the fitted reference length-scale reflects large-scale smoothness rather than only local variation. For the local residual component, $\mathcal H_d$ is chosen to contain short-range lags, and the short-range empirical correlation decay is used as a proxy for local residual dependence. The tapering range $\lambda_d$ is validated from the observed short-range correlation decay, for example, by examining the lag at which the empirical correlation becomes weak or by comparing with a small set of local-neighborhood candidate values. Since the global and local components are not separately observed, this empirical fitting step is used only as a diagnostic validation of the pre-specified kernel hyperparameters, rather than as a formal decomposition of the covariance structure.

\begin{table}[t!]
  \centering
  \caption{Empirical correlation-profile fitting results used to validate the pre-specified kernel hyperparameters for the Seattle traffic speed dataset (S) imputation.}
    \begin{tabular}{ll|rrrrrr}
    \toprule
    MS & SR & $\ell_{\boldsymbol{u}}^{(1)}$ & $\ell_{\boldsymbol{u}}^{(2)}$ & $\ell_{\boldsymbol{r}}^{(1)}$ & $\lambda_1$ & $\ell_{\boldsymbol{r}}^{(2)}$ & $\lambda_2$ \\
    \midrule
    RM & 0.7 & 1.45 & 26.30 & 1.00 & 20.00 & 2.50 & 25.00  \\
    & 0.3 & 5.74 & 26.44 & 2.01 & 20.00 & 7.50 & 37.50  \\
    & 0.1 & 3.13 & 39.41 & 1.99 & 20.00 & 2.50 & 25.50  \\
    \cmidrule{2-8}
    NM & 0.7 & 2.93 & 26.13 & 1.00 & 10.00 & 5.00 & 50.00  \\
    & 0.5 & 2.83 & 26.27 & 1.00 & 10.00 & 5.00 & 50.00  \\
    \bottomrule
    \end{tabular}
  \label{tab:sedata-hyper-empirical}
\end{table}

\begin{table}[t!]
  \centering
  \caption{Empirical correlation-profile fitting results used to validate the pre-specified kernel hyperparameters for the PeMS traffic speed dataset (P) imputation.}
    \begin{tabular}{ll|rrrrrr}
    \toprule
    MS & SR & $\ell_{\boldsymbol{u}}^{(1)}$ & $\ell_{\boldsymbol{u}}^{(2)}$ & $\ell_{\boldsymbol{r}}^{(1)}$ & $\lambda_1$ & $\ell_{\boldsymbol{r}}^{(2)}$ & $\lambda_2$ \\
    \midrule
    RM & 0.7 & 2.10 & 11.12 & 1.00 & 10.00 & 11.12 & 33.36 \\
    & 0.3 & 2.10 & 11.17 & 1.00 & 10.00 & 11.17 & 33.51 \\
    & 0.1 & 1.00 & 11.11 & 1.00 & 10.00 & 10.02 & 33.32 \\
    \cmidrule{2-8}
    NM & 0.7 & 2.10 & 11.12 & 1.77 & 10.00 & 11.12 & 33.36 \\
    & 0.5 & 2.12 & 11.11 & 1.77 & 10.00 & 11.11 & 33.34 \\
    \bottomrule
    \end{tabular}
  \label{tab:pems-hyper-empirical}
\end{table}

Tables~\ref{tab:sedata-hyper-empirical}--\ref{tab:pems-hyper-empirical} report the empirical correlation-profile fitting results for the two traffic speed datasets. The results are broadly consistent with the pre-specified kernel hyperparameters used in the GLSKF experiments. For the Seattle dataset (S), the fitted temporal length-scales for the global component are mostly around 26--39 time points, corresponding to approximately 130--200 minutes under the 5-minute sampling interval. This supports the use of a global temporal length-scale of approximately 3 hours. In contrast, the fitted temporal length-scales for the local residual component are much smaller, supporting the use of a short-range temporal scale of approximately 25--30 minutes. Similar patterns are observed for the PeMS dataset (P), where the fitted temporal scales also support a separation between the smoother global component and the short-range local residual component.

For the image, video, and MRI datasets, the kernel hyperparameters are specified using the same principle. The global component is assigned smoother covariance structures to capture low-rank spatial or spatiotemporal variation, whereas the local residual component is assigned shorter length-scales and compact tapering ranges to capture localized pixel-, frame-, or voxel-level dependence. The final selected kernel hyperparameters for all applied datasets are summarized in Supplementary~\ref{appendix-sec:configuration}. We further examine the robustness of GLSKF to these choices through the sensitivity analysis in Section~\ref{subsec:sensitivity}.

\subsection{Tuning of Regularization Parameters} \label{appendix:regularization}
In the second stage, after the covariance matrices are fixed, the regularization parameters $\rho$ and $\gamma$ are selected by validation on observed entries. These parameters control the relative strengths of the global and local regularization penalties in GLSKF. After selecting $\rho$ and $\gamma$, the final GLSKF model is refitted using all available observed entries. No ground-truth missing entries are used in either stage of parameter selection.

\begin{figure*}[t!]
\centering
\includegraphics[width=1\textwidth]{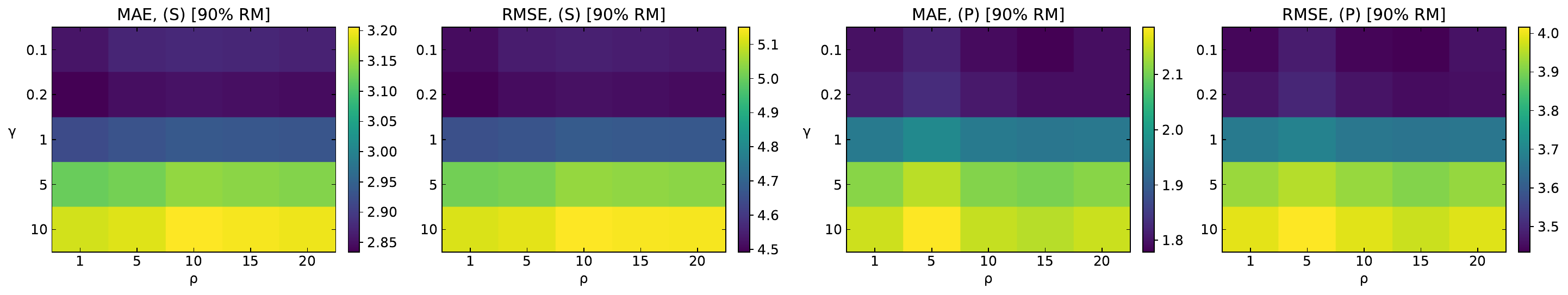}
\caption{Regularization parameter selection for traffic speed imputation at 90\% missingness. We show the MAE and RMSE across different combinations of $\rho$ and $\gamma$. The best result is obtained by $\rho=1$, $\gamma=0.2$ for (S); $\rho=15$, $\gamma=0.1$ for (P); see Table~\ref{tab:traffic-regularization}.}
\label{fig:traffic_parameter}
\end{figure*}

\begin{figure*}[t!]
\centering
\includegraphics[width=0.8\textwidth]{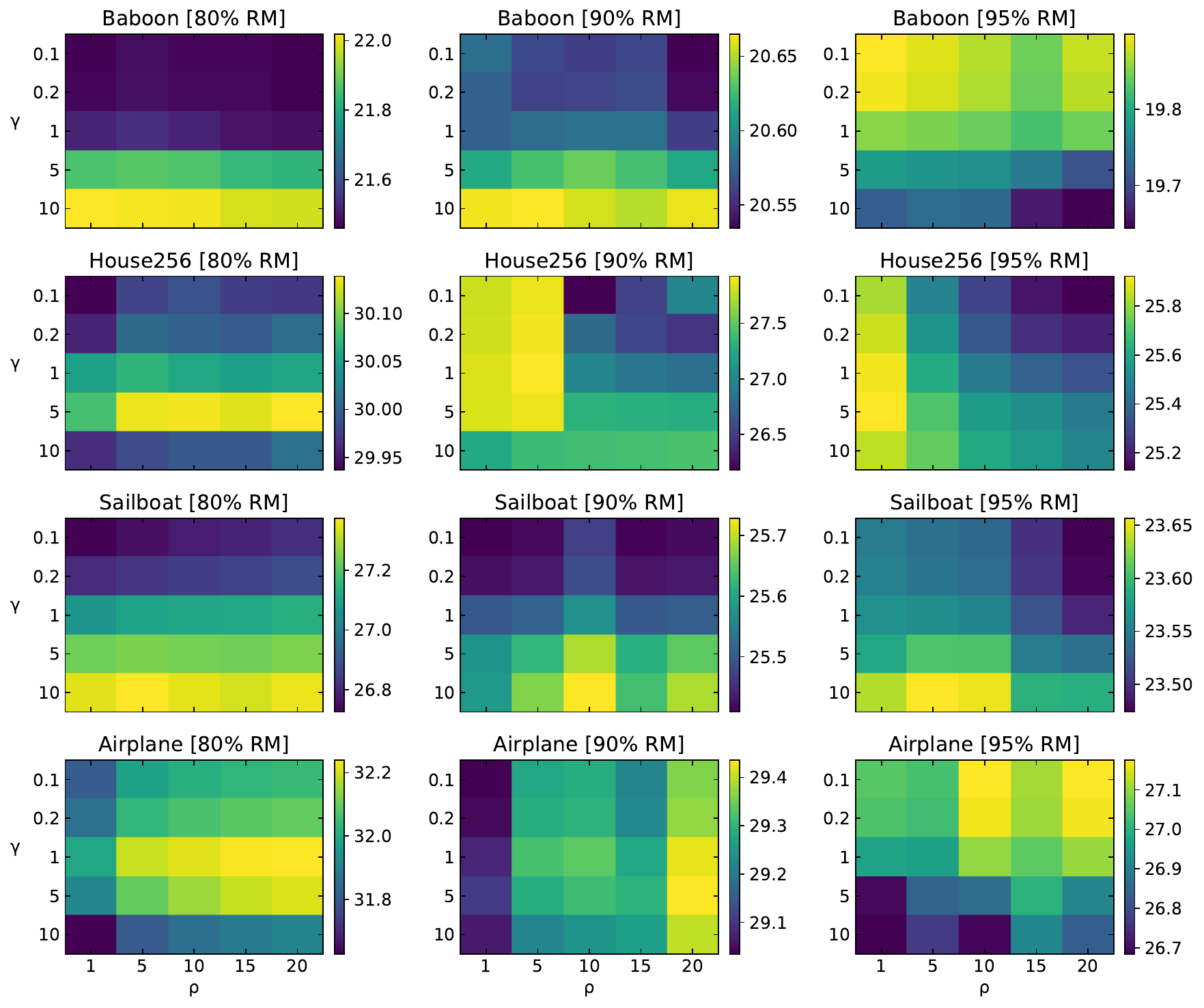}
\caption{Regularization parameter selection for image inpainting on \textit{Baboon}, \textit{House256}, \textit{Sailboat}, \textit{Airplane}. We show the PSNR across different combinations of $\rho$ and $\gamma$. The applied regularization parameters are summarized in Table~\ref{tab:image-config-regularization-RM}.}
\label{fig:image_parameter}
\end{figure*}

\begin{figure*}[t!]
\centering
\includegraphics[width=0.55\textwidth]{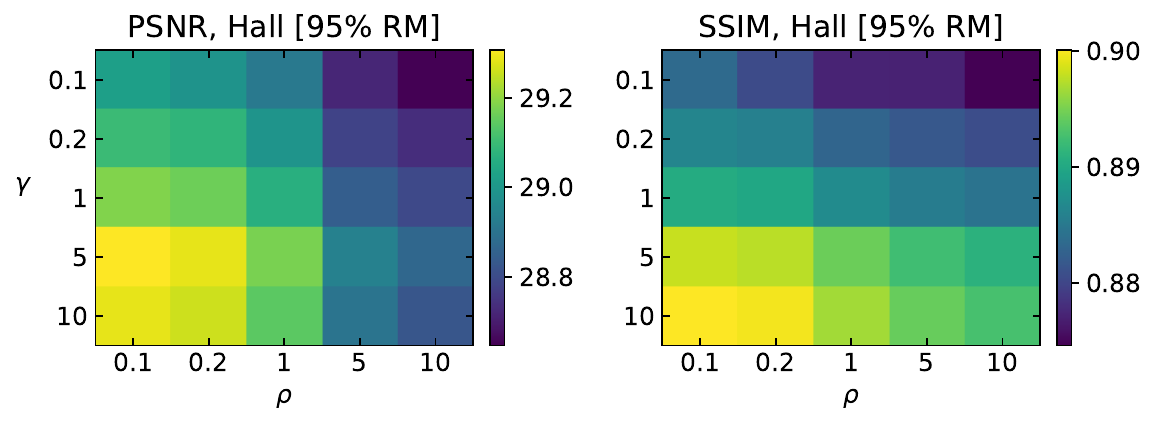}
\caption{Regularization parameter selection for video completion on \textit{Hall} at 95\% missingness. We show the PSNR and SSIM across different $\{\rho,\gamma\}$ settings. The applied regularization parameters are given in Table~\ref{tab:video-regularization}.}
\label{fig:video_parameter}
\end{figure*}

\begin{figure*}[t!]
\centering
\includegraphics[width=0.8\textwidth]{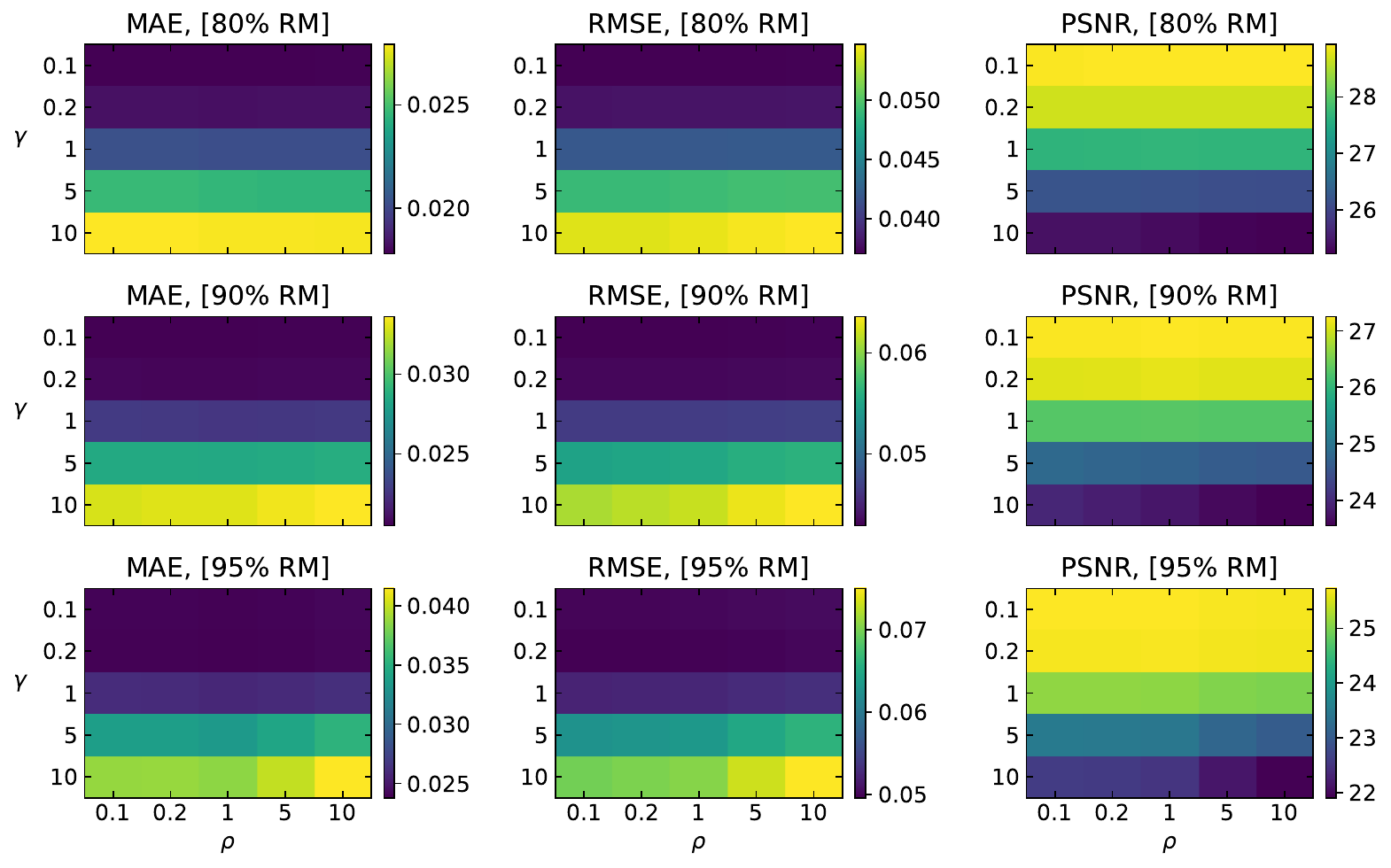}
\caption{Regularization parameter selection for MRI reconstruction at 80\%, 90\%, 95\% missingness. We show the MAE, RMSE, and PSNR under different $\{\rho,\gamma\}$ settings. The applied $\{\rho,\gamma\}$ are summarized in Table~\ref{tab:MRI-config}.}
\label{fig:MRI_parameter}
\end{figure*}

We use 5-fold cross validation to select the regularization parameters $\{\rho,\gamma\}$. Here we provide partial of the selection procedure. Specifically, we show tuning results for (i) traffic speed imputation at 90\% RM in Figure~\ref{fig:traffic_parameter}; (ii) color image inpainting on \{\textit{Baboon}, \textit{House256}, \textit{Sailboat}, \textit{Airplane}\} in Figure~\ref{fig:image_parameter}; (iii) video completion on \textit{Hall} at 95\% RM in Figure~\ref{fig:video_parameter}; and (iv) MRI reconstruction at 80\%, 90\%, 95\% missingness in Figure~\ref{fig:MRI_parameter}.

\section{GLSKF Parameter Configuration} \label{appendix-sec:configuration}
The kernel settings and regularization parameters used for each dataset in the experiments are summarized below. Throughout, we fix the variance hyperparameter in all kernel functions to $\sigma^2=1$.

\paragraph{Traffic speed imputation}
We use graph regularized Laplacian kernel $k_{\text{RL}}(\ell)$ for the spatial location dimension ($d=1$) and Mat\'ern 3/2 kernel $k_{\text{Mat\'ern 3/2}}(\ell)$ for the temporal time-of-day dimension ($d=2$), the kernel function configurations for the two traffic speed datasets are summarized in Tables~\ref{tab:seattle-kernel}--\ref{tab:pems-kernel}. The regularization parameters are summarized in Table~\ref{tab:traffic-regularization}.

\begin{table*}[t!]
  \centering
  \caption{Kernel settings (covariance configuration) for Seattle traffic speed (S) imputation.}
    \begin{tabular}{l|rr}
    \toprule
    Dimension (mode) & Global & Local  \\
    \midrule
    $d=1$: location & $\boldsymbol{K}_{\boldsymbol{u}}^{(1)}$: $k_{\text{RL}}\left(\ell_{\boldsymbol{u}}^{(1)}=1\right)$ & $\boldsymbol{K}_{\boldsymbol{\mathcal{R}}}^{(1)}$: $k_{\text{RL}}\left(\ell_{\boldsymbol{r}}^{(1)}=1\right)k_{\text{taper}}\left(\lambda_1=10\right)$ \\
    $d=2$: time-of-day & $\boldsymbol{K}_{\boldsymbol{u}}^{(2)}$: $k_{\text{Mat\'ern~3/2}}\left(\ell_{\boldsymbol{u}}^{(2)}=40\right)$ & $\boldsymbol{K}_{\boldsymbol{\mathcal{R}}}^{(2)}$: $k_{\text{Mat\'ern~3/2}}\left(\ell_{\boldsymbol{r}}^{(2)}=5\right)k_{\text{taper}}\left(\lambda_2=10\right)$ \\
    $d=3$: day & $\boldsymbol{K}_{\boldsymbol{u}}^{(3)}$: $\boldsymbol{I}$ & $\boldsymbol{K}_{\boldsymbol{\mathcal{R}}}^{(3)}$: $\text{cov}(\boldsymbol{R}_{(3)})$ \\
    \bottomrule
    \end{tabular}
  \label{tab:seattle-kernel}
\end{table*}

\begin{table*}[t!]
  \centering
  \caption{Kernel function configuration for PeMS traffic speed (P) imputation.}
    \begin{tabular}{l|rr}
    \toprule
    Dimension (mode) & Global & Local  \\
    \midrule
    $d=1$: location & $\boldsymbol{K}_{\boldsymbol{u}}^{(1)}$: $k_{\text{RL}}\left(\ell_{\boldsymbol{u}}^{(1)}=1\right)$ & $\boldsymbol{K}_{\boldsymbol{\mathcal{R}}}^{(1)}$: $k_{\text{RL}}\left(\ell_{\boldsymbol{r}}^{(1)}=1\right)k_{\text{taper}}\left(\lambda_1=10\right)$ \\
    $d=2$: time-of-day & $\boldsymbol{K}_{\boldsymbol{u}}^{(2)}$: $k_{\text{Mat\'ern~3/2}}\left(\ell_{\boldsymbol{u}}^{(2)}=20\right)$ & $\boldsymbol{K}_{\boldsymbol{\mathcal{R}}}^{(2)}$: $k_{\text{Mat\'ern~3/2}}\left(\ell_{\boldsymbol{r}}^{(2)}=5\right)k_{\text{taper}}\left(\lambda_2=10\right)$ \\
    $d=3$: day & $\boldsymbol{K}_{\boldsymbol{u}}^{(3)}$: $\boldsymbol{I}$ & $\boldsymbol{K}_{\boldsymbol{\mathcal{R}}}^{(3)}$: $\text{cov}(\boldsymbol{R}_{(3)})$ \\
    \bottomrule
    \end{tabular}
  \label{tab:pems-kernel}
\end{table*}

\begin{table*}[t!]
  \centering
  \caption{Regularization parameters for traffic speed imputation.}
    \begin{tabular}{ll|rr|rr}
    \toprule
    & & \multicolumn{2}{c}{(S)} & \multicolumn{2}{c}{(P)} \\
    \cmidrule{3-6}
    MS & SR & $\rho$ & $\gamma$ & $\rho$ & $\gamma$  \\
    \midrule
    RM & 0.7 & 1 & 0.2 & 15 & 0.1  \\
    & 0.3 & 1 & 0.2 & 15 & 0.1 \\
    & 0.1 & 1 & 0.2 & 15 & 0.1 \\
    \cmidrule{2-6}
    NM & 0.7 & 1 & 0.2 & 10 & 0.1 \\
    & 0.5 & 5 & 0.2 & 15 & 0.1\\
    \bottomrule
    \end{tabular}
  \label{tab:traffic-regularization}
\end{table*}

\paragraph{Color image inpainting}
We use Mat\'ern 3/2 kernel for both row and column pixel dimensions, and assume the length-scales and taper ranges are the same for the two dimensions. We use the same kernel hyperparameters for all the test color images. The kernel settings are summarized in Table~\ref{tab:image-kernel} and the regularization parameters for RM inpainting and SM completion are given in Tables~\ref{tab:image-config-regularization-RM}-\ref{tab:image-config-regularization-SM}.

\begin{table*}[t!]
  \centering
  \caption{Kernel configurations for Color image completion.}
    \begin{tabular}{l|rr}
    \toprule
    Dimension (mode) & Global & Local  \\
    \midrule
    $d=1$: row pixel & $\boldsymbol{K}_{\boldsymbol{u}}^{(1)}$: $k_{\text{Mat\'ern 3/2}}\left(\ell_{\boldsymbol{u}}^{(1)}=30\right)$ & $\boldsymbol{K}_{\boldsymbol{\mathcal{R}}}^{(1)}$: $k_{\text{Mat\'ern 3/2}}\left(\ell_{\boldsymbol{r}}^{(1)}=5\right)k_{\text{taper}}\left(\lambda_1=10\right)$ \\
    $d=2$: column pixel & $\boldsymbol{K}_{\boldsymbol{u}}^{(2)}$: $k_{\text{Mat\'ern~3/2}}\left(\ell_{\boldsymbol{u}}^{(2)}=30\right)$ & $\boldsymbol{K}_{\boldsymbol{\mathcal{R}}}^{(2)}$: $k_{\text{Mat\'ern~3/2}}\left(\ell_{\boldsymbol{r}}^{(2)}=5\right)k_{\text{taper}}\left(\lambda_2=10\right)$ \\
    $d=3$: channel & $\boldsymbol{K}_{\boldsymbol{u}}^{(3)}$: $\boldsymbol{I}$ & $\boldsymbol{K}_{\boldsymbol{\mathcal{R}}}^{(3)}$: $\text{cov}(\boldsymbol{R}_{(3)})$ \\
    \bottomrule
    \end{tabular}
  \label{tab:image-kernel}
\end{table*}

\begin{table*}[t!]
  \centering
  \caption{Regularization parameters of GLSKF for Color image inpainting under RM.}
    \begin{tabular}{l|rr|rr|rr|rr|rr|rr|rr}
    \toprule
    \multicolumn{1}{c}{} & \multicolumn{2}{c}{Barbara} & \multicolumn{2}{c}{Baboon} & \multicolumn{2}{c}{House256} & \multicolumn{2}{c}{Peppers} & \multicolumn{2}{c}{Sailboat} & \multicolumn{2}{c}{House512} & \multicolumn{2}{c}{Airplane} \\
    \cmidrule(lr){2-15}
    SR & $\rho$ & $\gamma$ & $\rho$ & $\gamma$ & $\rho$ & $\gamma$ & $\rho$ & $\gamma$ & $\rho$ & $\gamma$ & $\rho$ & $\gamma$ & $\rho$ & $\gamma$ \\
    \cmidrule(lr){1-15}
    0.2 & 10 & 0.1 & 10 & 10 & 10 & 10 & 20 & 1 & 20 & 10 & 20 & 0.1 & 20 & 1 \\
    0.1 & 10 & 10 & 20 & 10 & 10 & 1 & 20 & 0.1 & 10 & 10 & 15 & 0.1 & 20 & 5 \\
    0.05 & 10 & 0.1 & 5 & 10 & 1 & 10 & 20 & 0.1 & 1 & 10 & 20 & 5 & 10 & 0.1 \\
    \bottomrule
    \end{tabular}
  \label{tab:image-config-regularization-RM}
\end{table*}

\begin{table*}[t!]
  \centering
  \caption{Regularization parameters of GLSKF for Color image completion under SM.}
    \begin{tabular}{l|rr|rr|rr|rr|rr|rr|rr}
    \toprule
    \multicolumn{1}{c}{} & \multicolumn{2}{c}{Barbara} & \multicolumn{2}{c}{Baboon} & \multicolumn{2}{c}{House256} & \multicolumn{2}{c}{Peppers} & \multicolumn{2}{c}{Sailboat} & \multicolumn{2}{c}{House512} & \multicolumn{2}{c}{Airplane} \\
    \cmidrule(lr){2-15}
    MS & $\rho$ & $\gamma$ & $\rho$ & $\gamma$ & $\rho$ & $\gamma$ & $\rho$ & $\gamma$ & $\rho$ & $\gamma$ & $\rho$ & $\gamma$ & $\rho$ & $\gamma$ \\
    \cmidrule(lr){1-15}
    Text & 1 & 5 & 1 & 1 & 1 & 0.2 & 5 & 5 & 10 & 5 & 5 & 5 & 20 & 0.2 \\
    Stripe & 1 & 5 & 1 & 5 & 10 & 0.2 & 1 & 0.2 & 1 & 10 & 1 & 10 & 5 & 10 \\
    Demosaic & 1 & 0.1 & 1 & 1 & 1 & 0.1 & 1 & 0.1 & 1 & 0.1 & 1 & 0.1 & 20 & 0.1 \\
    \bottomrule
    \end{tabular}
  \label{tab:image-config-regularization-SM}
\end{table*}

\paragraph{Video completion} We set the kernel function as Mat\'ern 3/2 for the row, column pixels, and frame dimensions of the color video data. The kernel function configuration is given in Table~\ref{tab:video-kernel}, and the regularization parameters are summarized in Table~\ref{tab:video-regularization}.

\begin{table*}[t!]
  \centering
  \caption{Kernel function settings for Color video completion.}
    \begin{tabular}{l|rr}
    \toprule
    Dimension (mode) & Global & Local  \\
    \midrule
    $d=1$: row pixel & $\boldsymbol{K}_{\boldsymbol{u}}^{(1)}$: $k_{\text{Mat\'ern 3/2}}\left(\ell_{\boldsymbol{u}}^{(1)}=30\right)$ & $\boldsymbol{K}_{\boldsymbol{\mathcal{R}}}^{(1)}$: $k_{\text{Mat\'ern 3/2}}\left(\ell_{\boldsymbol{r}}^{(1)}=5\right)k_{\text{taper}}\left(\lambda_1=10\right)$ \\
    $d=2$: column pixel & $\boldsymbol{K}_{\boldsymbol{u}}^{(2)}$: $k_{\text{Mat\'ern~3/2}}\left(\ell_{\boldsymbol{u}}^{(2)}=30\right)$ & $\boldsymbol{K}_{\boldsymbol{\mathcal{R}}}^{(2)}$: $k_{\text{Mat\'ern~3/2}}\left(\ell_{\boldsymbol{r}}^{(2)}=5\right)k_{\text{taper}}\left(\lambda_2=10\right)$ \\
    $d=3$: channel & $\boldsymbol{K}_{\boldsymbol{u}}^{(3)}$: $\boldsymbol{I}$ & $\boldsymbol{K}_{\boldsymbol{\mathcal{R}}}^{(3)}$: $\text{cov}(\boldsymbol{R}_{(3)})$ \\
    $d=4$: frame & $\boldsymbol{K}_{\boldsymbol{u}}^{(4)}$: $k_{\text{Mat\'ern 3/2}}(\ell_{\boldsymbol{u}}^{(4)}=5)$ & $\boldsymbol{K}_{\boldsymbol{\mathcal{R}}}^{(4)}$: $k_{\text{Mat\'ern 3/2}}(\ell_{\boldsymbol{r}}^{(4)}=5)k_{\text{taper}}(\lambda_4=10)$  \\
    \bottomrule
    \end{tabular}
  \label{tab:video-kernel}
\end{table*}

\begin{table*}[t!]
  \centering
  \caption{Regularization parameters for Color video completion.}
    \begin{tabular}{l|rr|rr|rr}
    \toprule
    \multicolumn{1}{c}{} & \multicolumn{2}{c}{Hall} & \multicolumn{2}{c}{Foreman} & \multicolumn{2}{c}{Carphone} \\
    \cmidrule(lr){2-7}
    SR & $\rho$ & $\gamma$ & $\rho$ & $\gamma$ & $\rho$ & $\gamma$ \\
    \midrule
    0.2 & 10 & 1 & 0.2 & 5 & 0.1 & 5 \\
    0.1 & 1 & 10 & 0.2 & 10 & 0.2 & 5 \\
    0.05 & 0.1 & 10 & 0.2 & 10 & 0.1 & 5  \\
    \bottomrule
    \end{tabular}
  \label{tab:video-regularization}
\end{table*}

\paragraph{MRI completion} The kernel functions and regularization parameters configuration for MRI completion are summarized in Table~\ref{tab:MRI-config}.

\begin{table*}[t!]
  \centering
  \caption{Configuration of GLSKF for MRI image reconstruction.}
    \begin{tabular}{l|l|lrr}
    \toprule
    \multicolumn{2}{c|}{Kernel settings for covariance matrices} & SR & $\rho$ & $\gamma$ \\
    \cmidrule(lr){1-2} \cmidrule(lr){3-5}
    $\boldsymbol{K}_{\boldsymbol{u}}^{(1)}$: $k_{\text{Mat\'ern~3/2}}\left(\ell_{\boldsymbol{u}}^{(1)}=30\right)$ & $\boldsymbol{K}_{\boldsymbol{\mathcal{R}}}^{(1)}$: {$k_{\text{Mat\'ern~3/2}}\left(\ell_{\boldsymbol{r}}^{(1)}=5\right)k_{\text{taper}}\left(\lambda_1=10\right)$} & 0.2 & 0.1 & 0.1   \\
    \cmidrule(lr){1-2} \cmidrule(lr){4-5}
    $\boldsymbol{K}_{\boldsymbol{u}}^{(2)}$: $k_{\text{Mat\'ern~3/2}}\left(\ell_{\boldsymbol{u}}^{(2)}=30\right)$ & $\boldsymbol{K}_{\boldsymbol{\mathcal{R}}}^{(2)}$: $k_{\text{Mat\'ern~3/2}}\left(\ell_{\boldsymbol{r}}^{(2)}=5\right)k_{\text{taper}}\left(\lambda_2=10\right)$ & 0.1 & 1 & 0.1   \\
    \cmidrule(lr){1-2} \cmidrule(lr){4-5}
    $\boldsymbol{K}_{\boldsymbol{u}}^{(3)}$: $k_{\text{Mat\'ern~3/2}}\left(\ell_{\boldsymbol{u}}^{(3)}=5\right)$ & $\boldsymbol{K}_{\boldsymbol{\mathcal{R}}}^{(3)}$: $k_{\text{Mat\'ern~3/2}}\left(\ell_{\boldsymbol{r}}^{(3)}=5\right)k_{\text{taper}}\left(\lambda_3=10\right)$ & 0.05, 0.01 & 0.1 & 0.2   \\
    \bottomrule
    \end{tabular}
  \label{tab:MRI-config}
\end{table*}

\section{Supplementary Experimental Results} \label{appendix-sec:results}
\subsection{Traffic Speed Imputation} \label{appendix_traffic_results}
\begin{figure*}[t!]
\centering
\includegraphics[width=1\textwidth]{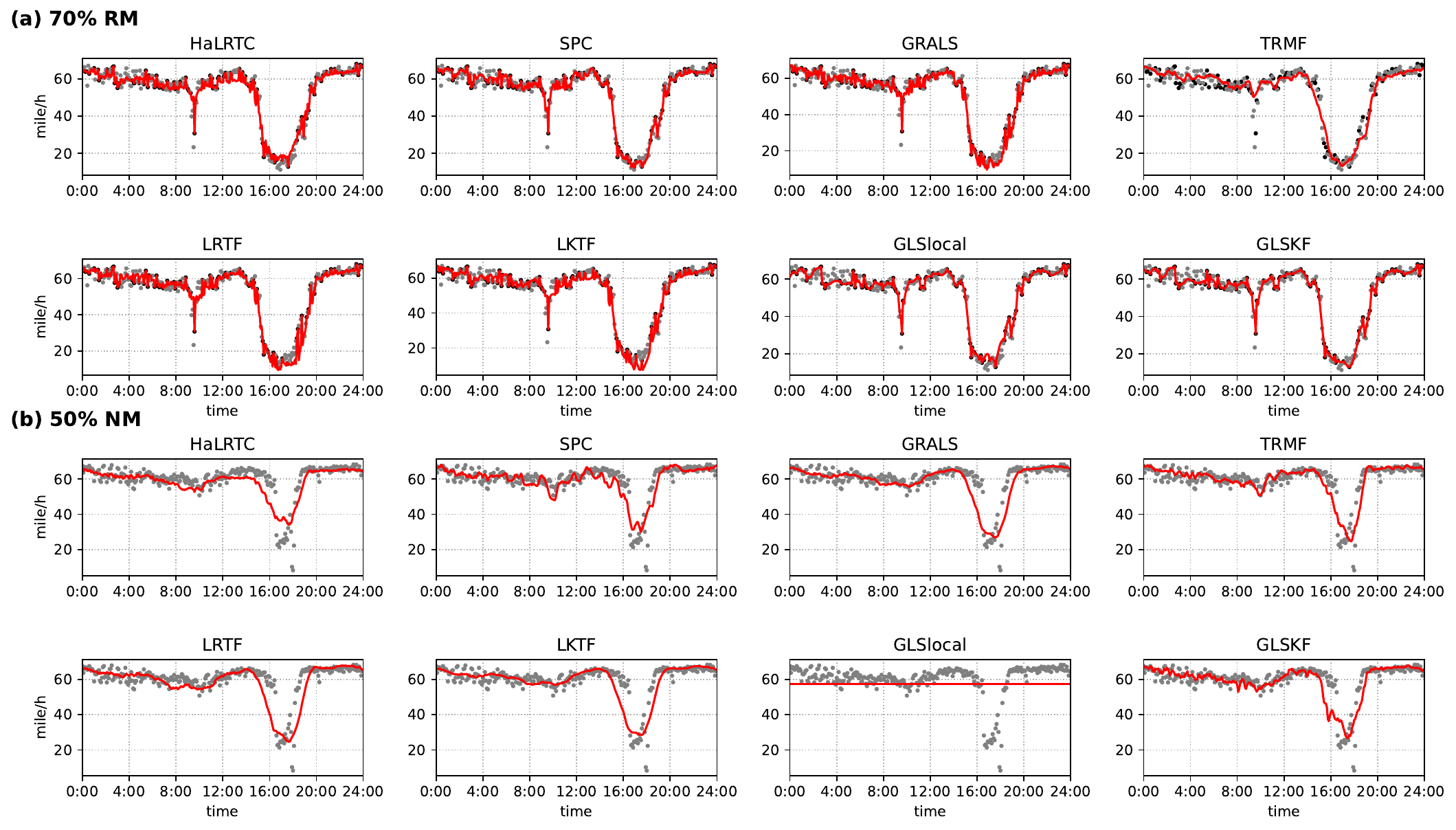}
\caption{Imputation results for the Seattle traffic speed dataset (S) under (a) 70\% RM and (b) 50\% NM, showing day 15 for RM and day 6 for NM at location 200. Black and gray dots represent the observed and missing entries, respectively.}
\label{fig:traffic-completion_appendix}
\end{figure*}

Example imputation results for the Seattle traffic speed dataset under 70\% RM and 50\% NM are shown in Figure~\ref{fig:traffic-completion_appendix}.

\subsection{Color Image Completion} \label{appendix_image_results}
\begin{figure*}[t!]
\centering
\includegraphics[width=0.9\textwidth]{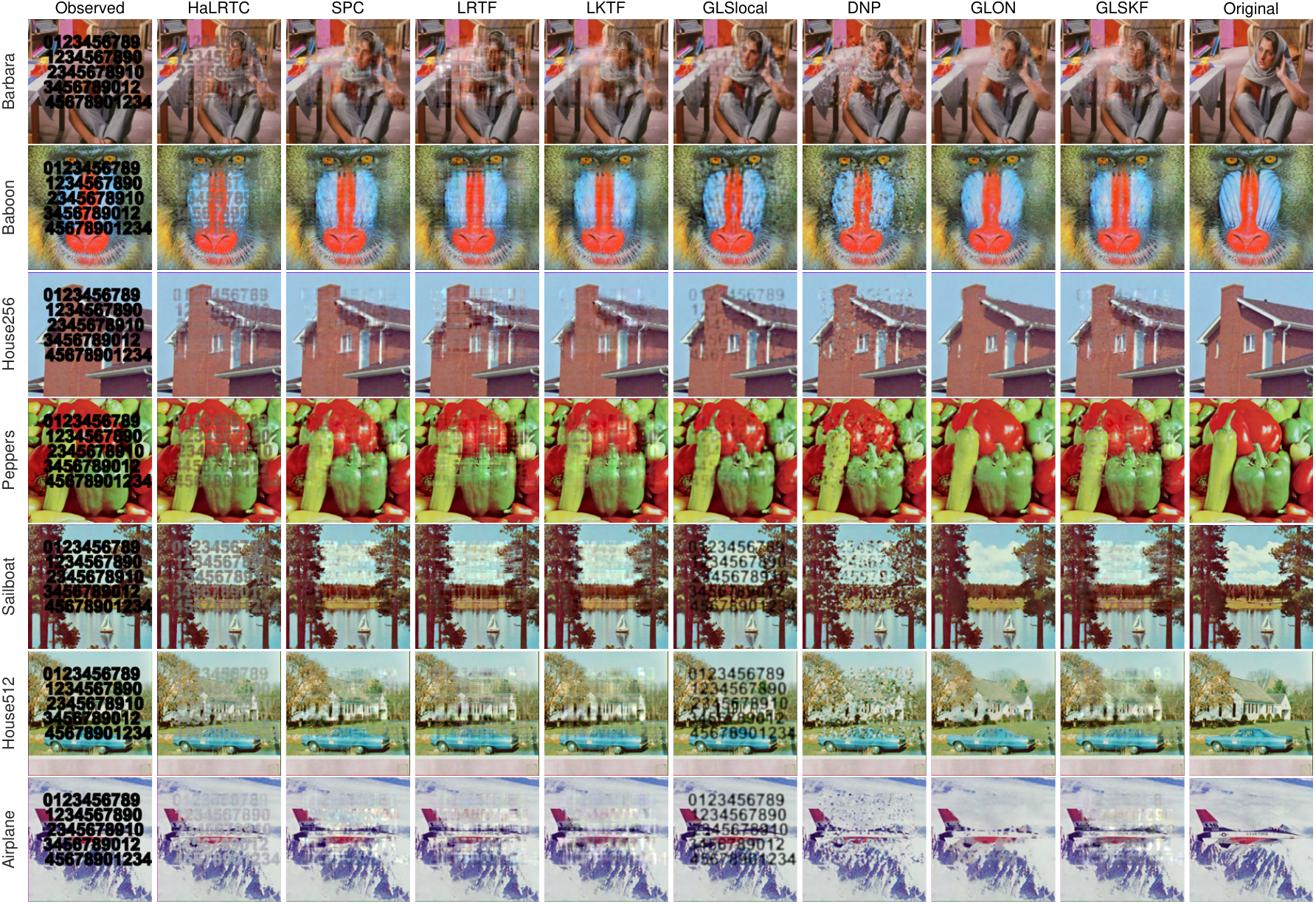}
\caption{Color image completion under structured text missing.}
\label{fig:image_text}
\end{figure*}

\begin{figure*}[t!]
\centering
\includegraphics[width=0.9\textwidth]{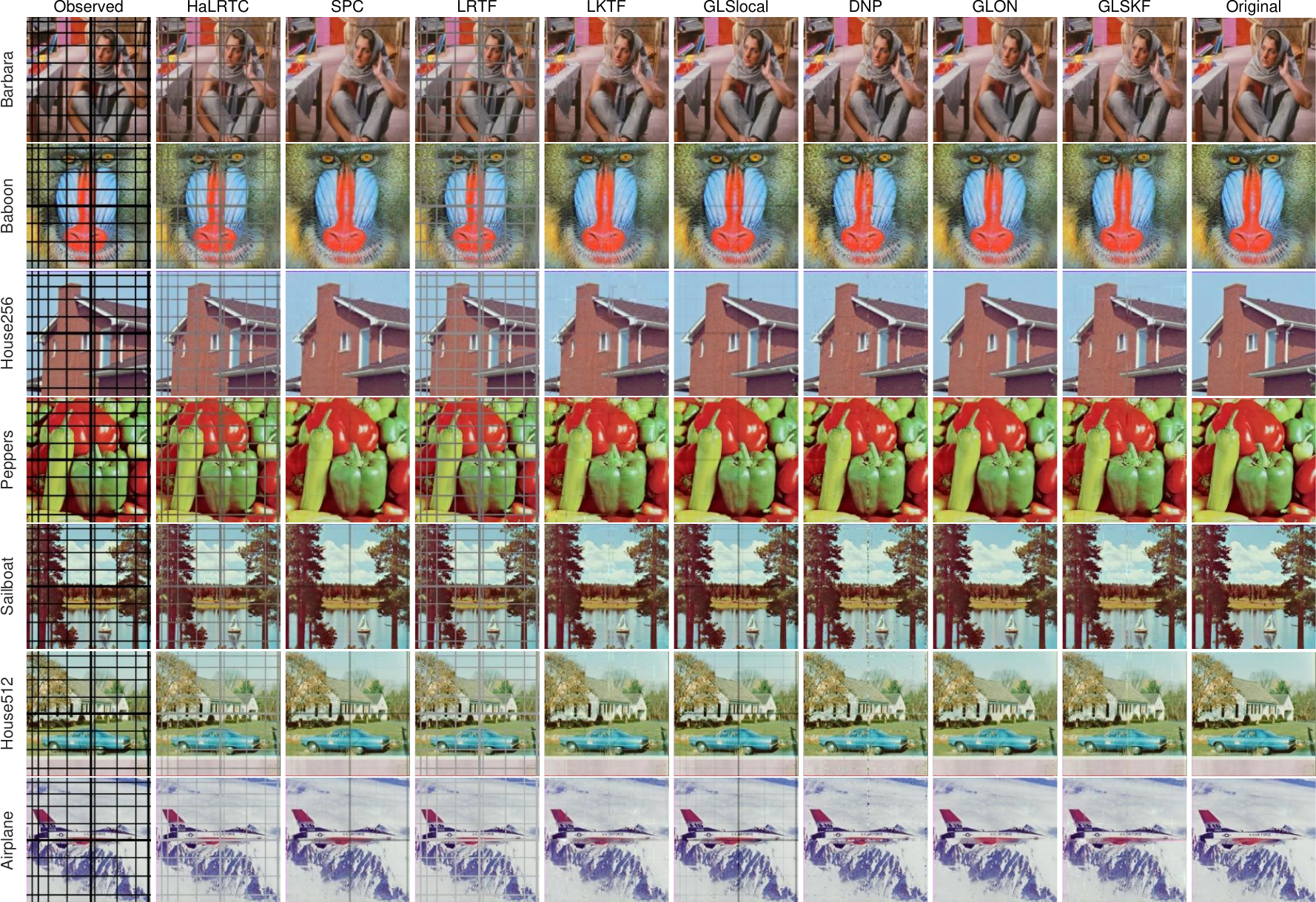}
\caption{Color image completion under structured stripe/fiber missing.}
\label{fig:image_stripes}
\end{figure*}

\begin{figure*}[t!]
\centering
\includegraphics[width=0.9\textwidth]{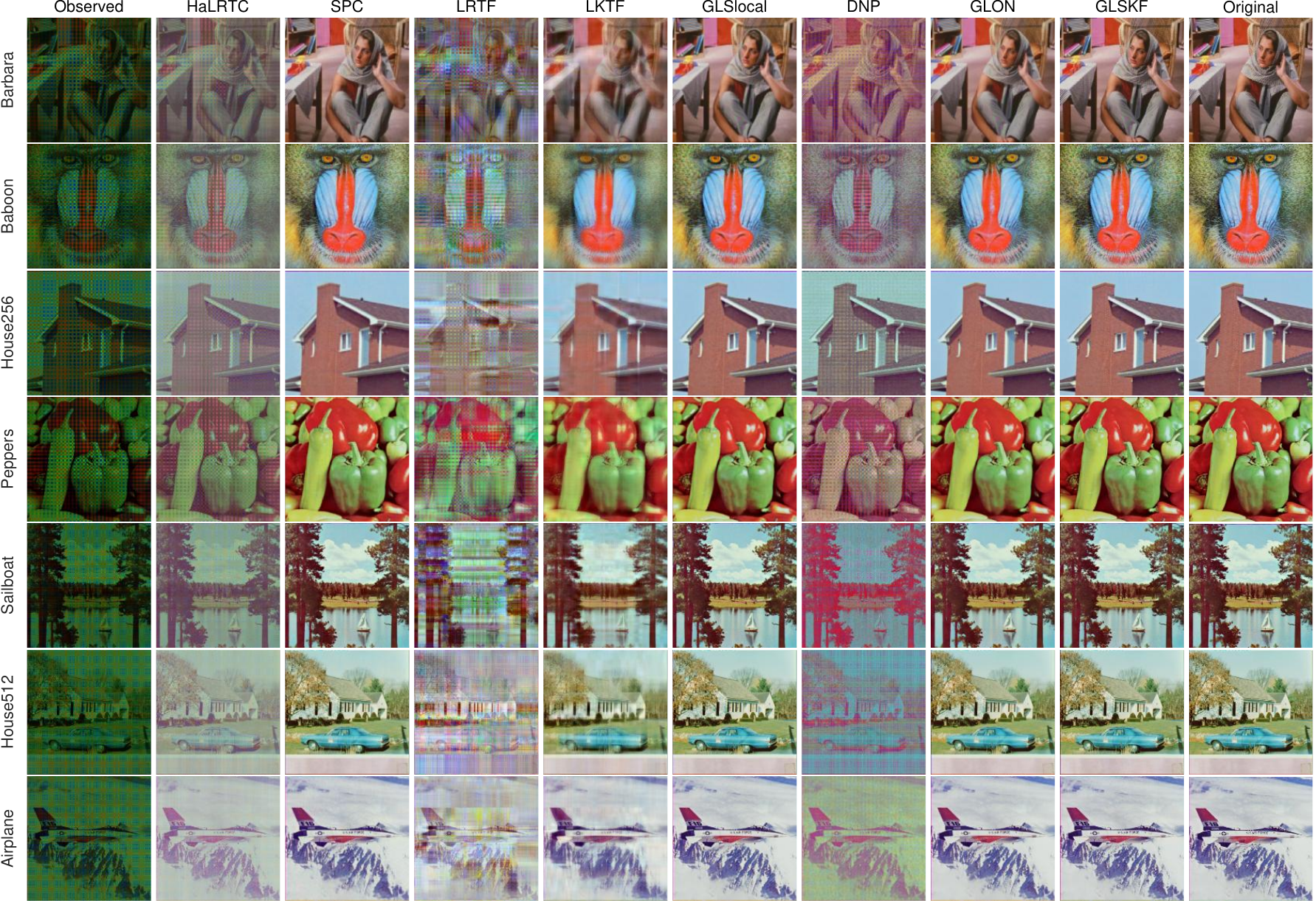}
\caption{Comparison results for color image demosaicing.}
\label{fig:image_demosaic}
\end{figure*}

The quantitative comparison under structured missingness (SM) is given in Table~\ref{tab:Image-comTable-SM}. The completion results under structured text missingness, structured stripe missingness, and demosaicing are shown in Figures~\ref{fig:image_text},~\ref{fig:image_stripes}, and~\ref{fig:image_demosaic}, respectively.

\begin{table*}[t!]
  \centering
  \caption{Comparison of color image completion under SM (PSNR/SSIM).}
    \begin{tabular}{ll|rrrrrrrr}
    \toprule
     & MS & {HaLRTC} & {SPC} & LRTF & LKTF & GLSlocal & DNP & GLON & GLSKF  \\
    \midrule
    \multirow{3}{*}{\rotatebox{90}{Barbara}} & text & 22.57/0.756 & 24.82/0.814 & 22.10/0.733 & 23.31/0.780 & 25.87/0.846 & 23.83/0.770 & \underline{26.56}/\underline{0.859} & \textbf{26.29}/\textbf{0.858} \\
    & strip & 20.77/0.724 & 29.69/0.920 & 20.68/0.720 & 26.60/0.856 & 30.20/0.933 & 27.75/0.873 & \underline{30.78}/\underline{0.935} & \textbf{30.10}/\textbf{0.937} \\
    & dem & 16.18/0.341 & \underline{29.86}/\underline{0.880} & 14.59/0.239 & 22.40/0.630 & 29.16/0.880 & 16.14/0.280 & 28.78/0.864 & \textbf{31.71}/\textbf{0.920} \\
    \cmidrule(r){2-10}
    \multirow{3}{*}{\rotatebox{90}{Baboon}} & text & 21.07/0.724 & 24.50/0.789 & 23.31/0.754 & 23.95/0.778 & 23.86/0.795 & 21.29/0.722 & \underline{25.24}/\underline{0.812} & \textbf{25.00}/\textbf{0.816} \\
    & strip & 20.61/0.758 & 27.66/0.886 & 20.31/0.745 & 26.40/0.863 & 27.35/0.888 & 24.15/0.812 & \underline{28.22}/\underline{0.898} & \textbf{27.49}/\textbf{0.893}  \\
    & dem & 15.91/0.378 & \underline{24.95}/\underline{0.764} & 13.14/0.218 & 21.83/0.584 & 24.29/0.736 & 15.46/0.324 & 24.41/0.733 & \textbf{24.90}/\textbf{0.791} \\ 
    \cmidrule(r){2-10}
    \multirow{3}{*}{\rotatebox{90}{House256}} & text & 23.69/0.784 & 26.10/0.847 & 22.94/0.764 & 24.78/0.820 & 25.90/0.834 & 25.59/0.805 & \underline{27.22}/\underline{0.886} & \textbf{28.49}/\textbf{0.886} \\
    & strip & 20.57/0.666 & 29.71/0.914 & 20.72/0.655 & 27.75/0.883 & 31.36/0.931 & 29.98/0.898 & \underline{31.89}/\underline{0.949} & \textbf{32.91}/\textbf{0.947} \\
    & dem & 16.26/0.309 & 29.71/0.850 & 12.88/0.123 & 23.67/0.700 & \underline{31.58}/0.875 & 19.95/0.382 & 31.10/\underline{0.881} & \textbf{32.91}/\textbf{0.947} \\
    \cmidrule(r){2-10}
    \multirow{3}{*}{\rotatebox{90}{Peppers}} & text & 20.60/0.733 & 23.98/0.815 & 20.98/0.727 & 21.71/0.767 & 25.16/0.855 & 23.74/0.790 & \underline{26.20}/\underline{0.895} & \textbf{26.44}/\textbf{0.891} \\
    & strip & 18.69/0.668 & 28.27/0.926 & 17.72/0.640 & 24.37/0.837 & \underline{29.99}/0.942 & 28.66/0.907 & 29.41/\underline{0.946} & \textbf{30.40}/\textbf{0.955} \\
    & dem & 13.49/0.234 & 29.18/0.882 & 11.63/0.128 & 20.63/0.590 & \underline{30.15}/\underline{0.926} & 13.53/0.199 & 29.84/0.912 & \textbf{31.87}/\textbf{0.944} \\
    \midrule
    \multirow{3}{*}{\rotatebox{90}{Sailboat}} & text & 18.49/0.690 & 23.05/0.791 & 21.89/0.750 & 22.45/0.777 & 16.80/0.711 & 18.09/0.687 & \underline{24.21}/\underline{0.869} & \textbf{24.35}/\textbf{0.842} \\
    & strip & 19.75/0.710 & 26.20/0.909 & 19.33/0.695 & 26.22/0.890 & 25.98/0.900 & 24.62/0.851 & \underline{28.95}/\underline{0.954} & \textbf{29.27}/\textbf{0.948} \\
    & dem & 14.11/0.694 & \underline{28.34}/{0.933} & 11.32/0.451 & 21.29/0.765 & 28.21/\underline{0.935} & 12.03/0.500 & 28.04/0.931 & \textbf{28.73}/\textbf{0.926} \\
    \cmidrule(r){2-10}
    \multirow{3}{*}{\rotatebox{90}{House512}} & text & 21.41/0.760 & 23.35/0.803 & 21.95/0.770 & 22.46/0.785 & 16.36/0.686 & 20.09/0.720 & \underline{23.46}/\underline{0.846} & \textbf{24.32}/\textbf{0.842} \\
    & strip & 21.43/0.742 & 25.66/0.903 & 20.93/0.743 & 26.71/0.901 & 24.95/0.884 & 25.86/0.857 & \underline{29.21}/\underline{0.953} & \textbf{29.51}/\textbf{0.951} \\
    & dem & 15.94/0.733 & 28.69/0.944 & 11.92/0.588 & 22.06/0.799 & \underline{29.61}/\underline{0.967} & 11.62/0.500 & 29.00/0.955 & \textbf{32.35}/\textbf{0.979} \\
    \cmidrule(r){2-10}
    \multirow{3}{*}{\rotatebox{90}{Airplane}} & text & 21.70/0.784 & 22.75/0.805 & 21.10/0.770 & 22.18/0.804 & 14.15/0.630 & 21.46/0.771 & \underline{23.72}/\underline{0.894} & \textbf{25.07}/\textbf{0.883} \\
    & strip & 22.14/0.749 & 25.45/0.903 & 21.94/0.771 & 27.20/0.911 & 24.30/0.864 & 27.30/0.905 & \underline{30.51}/\underline{0.966} & \textbf{31.31}/\textbf{0.967} \\
    & dem & 17.03/0.810 & 27.95/0.940 & 13.96/0.638 & 22.58/0.831 & \underline{31.09}/\underline{0.980} & 12.42/0.559 & 30.40/0.970 & \textbf{34.36}/\textbf{0.986} \\
    \bottomrule
    \multicolumn{10}{l}{Best results are highlighted in bold, and second-best results are underlined.}
    \end{tabular}
  \label{tab:Image-comTable-SM}
\end{table*}

\subsection{MRI Completion} \label{appendix_MRI}
\begin{figure*}[t!]
\centering
\includegraphics[width=0.9\textwidth]{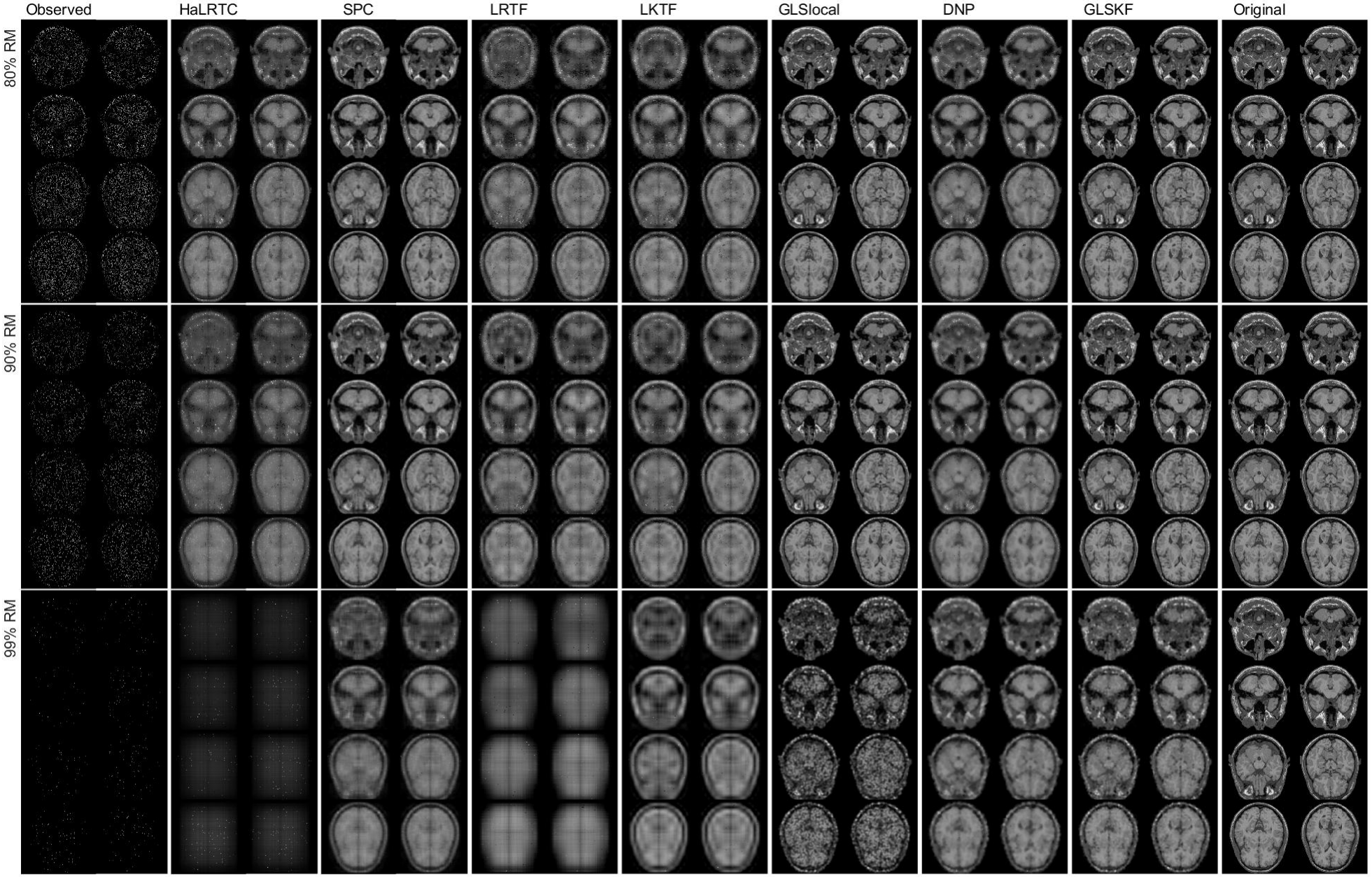}
\caption{Illustration of MRI image completion for eight frontal slices (10th, 20th, 30th, 40th, 50th, 60th, 70th, and 80th) under 80\%, 90\%, and 99\% missingness.}
\label{fig:MRIcompletion_appendix}
\end{figure*}

The RMSE comparison for MRI completion is summarized in Table~\ref{tab:MRI-comTable-RMSE}, and representative completion results under 80\%, 90\%, 99\% random missingness are shown in Figure~\ref{fig:MRIcompletion_appendix}.

\begin{table*}[t!]
  \centering
  \caption{Comparison of MRI image completion (RMSE).}
    \begin{tabular}{ll|rrrrrrr}
    \toprule
    MS & {SR} & {HaLRTC} & {SPC} & LRTF & LKTF & GLSlocal & DNP & {GLSKF}  \\
    \midrule
    RM & 0.2 & 0.0725 & 0.0470 & 0.0882 & 0.0884 & \underline{0.0374} & 0.0686 & \textbf{0.0371}  \\
    & 0.1 & 0.0963 & 0.0523 & 0.0897 & 0.0886 & \underline{0.0434} & 0.0689 & \textbf{0.0428} \\
    & 0.05 & 0.1213 & 0.0592 & 0.0902 & 0.0898 & \underline{0.0513} & 0.0696 & \textbf{0.0496} \\
    & 0.01 & 0.1992 & \underline{0.0848} & 0.1203 & 0.0924 & 0.1011 & 0.1618 & \textbf{0.0685}  \\
    \cmidrule{2-9}
    PM & 0.2 & 0.1242 & \underline{0.0770} & 0.0958 & 0.0948 & 0.1499 & 0.1007 & \textbf{0.0720} \\
    & 0.1 & 0.1652 & \underline{0.0955} & 0.1534 & 0.1196 & 0.1963 & 0.1383 & \textbf{0.0906} \\
    \bottomrule
    \multicolumn{9}{l}{{Best results are highlighted in bold, and second-best results are underlined.}}
    \end{tabular}
  \label{tab:MRI-comTable-RMSE}
\end{table*}

\subsection{Complementary Global and Local Components} \label{appendix_subsec:components}

\begin{figure*}[t!]
\centering
\includegraphics[width=0.65\textwidth]{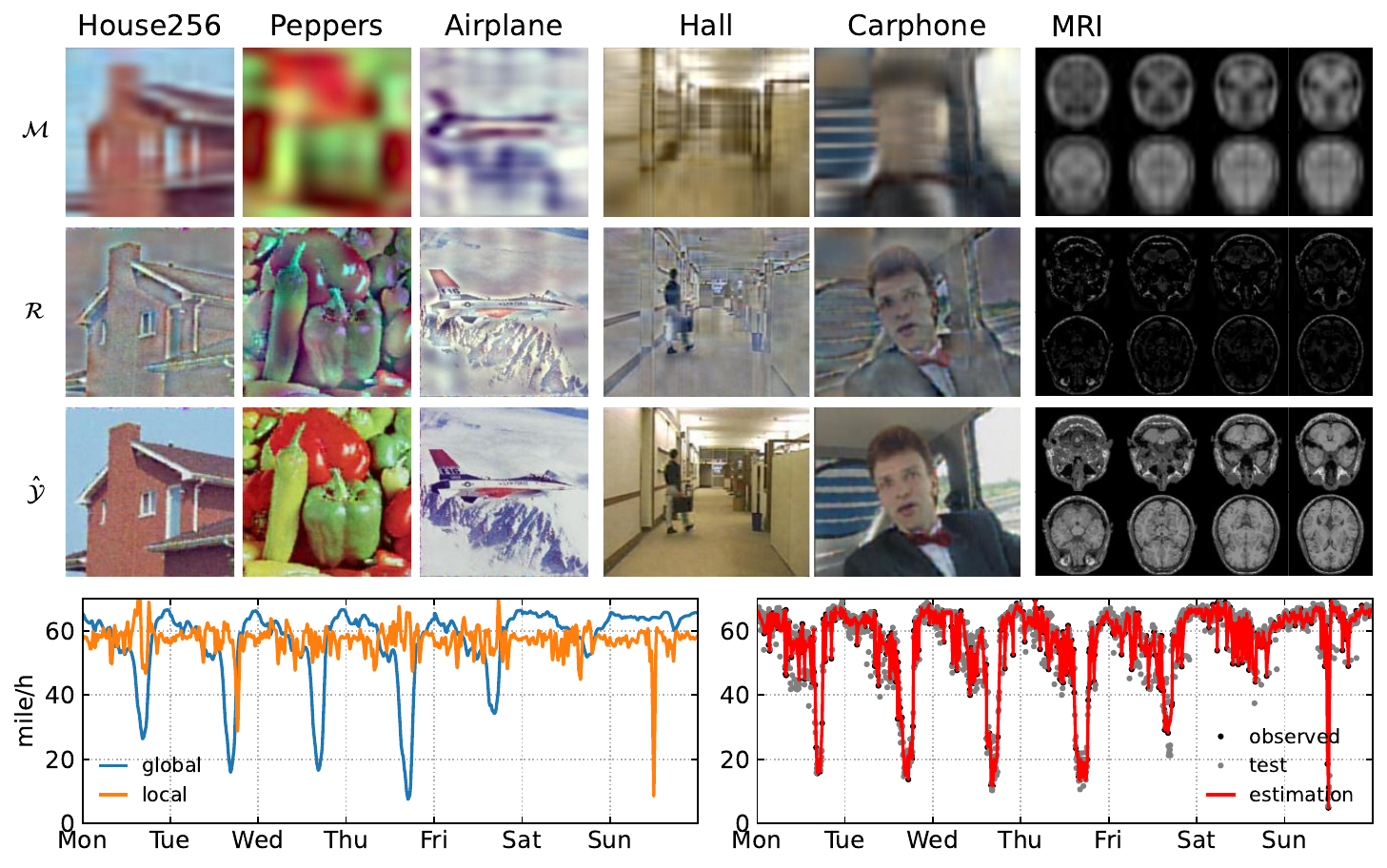}
\caption{Visualization of the global component $\boldsymbol{\mathcal{M}}$, local component $\boldsymbol{\mathcal{R}}$, and the completed tensor $\hat{\boldsymbol{\mathcal{Y}}}$ under random missingness scenarios. Results are shown for 90\% RM color image inpainting on \{House256, Peppers, Airplane\}, video completion at 80\% RM on {Hall} (frame 40) and 95\% RM on {Carphone} (frame 40), 90\% RM MRI completion on eight frontal slices, as well as 70\% RM traffic speed (S) imputation at location 10 for one week.}
\label{fig:Component-RM}
\end{figure*}

\paragraph{Random missing} 
The global and local components under random missingness across different datasets are shown in Figure~\ref{fig:Component-RM}.

\begin{figure*}[t!]
\centering
\includegraphics[width=0.75\textwidth]{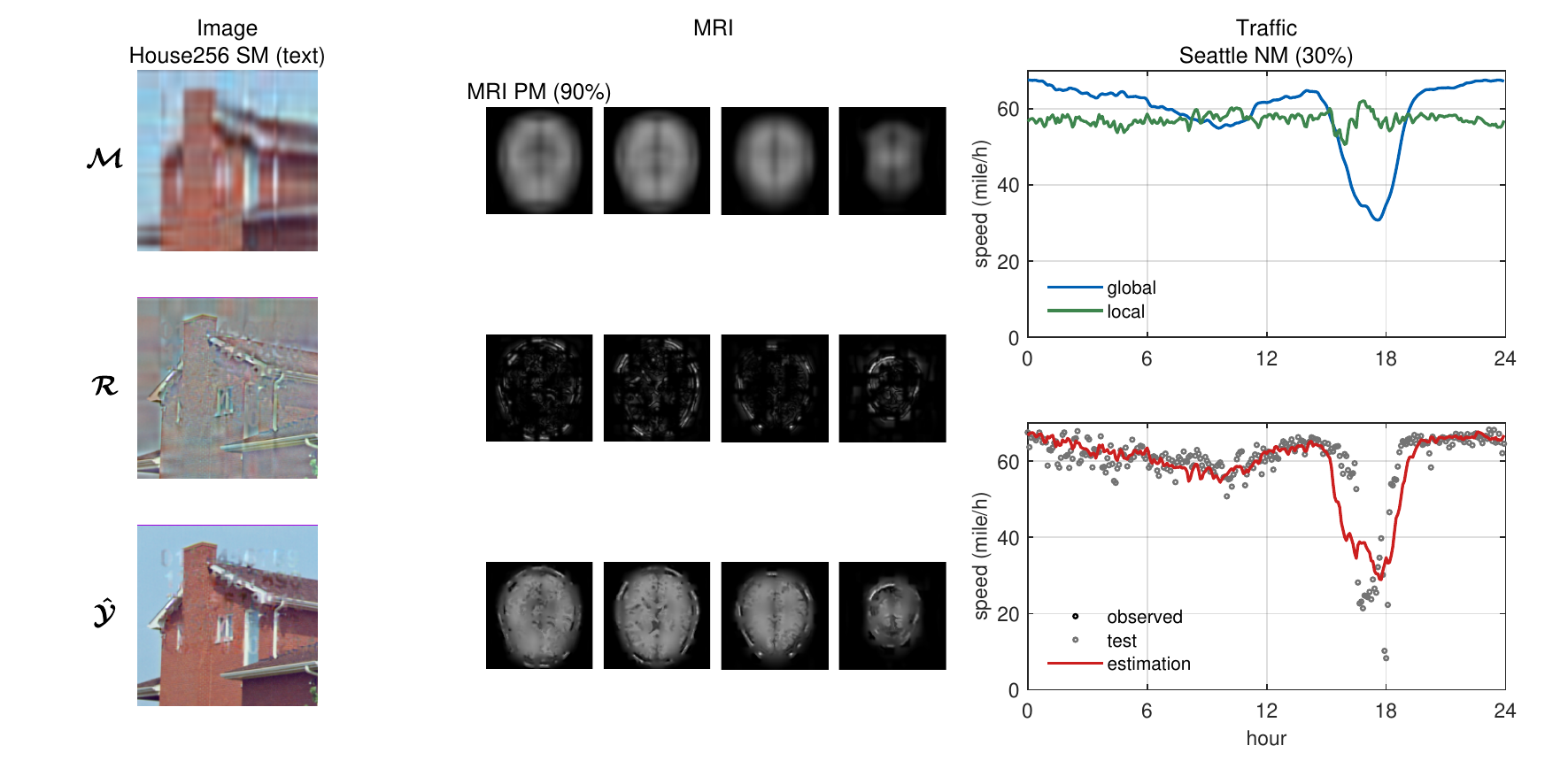}
\caption{Visualization of the global component $\boldsymbol{\mathcal{M}}$, local component $\boldsymbol{\mathcal{R}}$, and the completed tensor $\hat{\boldsymbol{\mathcal{Y}}}$ under nonrandom missingness scenarios. Results are shown for color image inpainting on \textit{House256} under structured text missing, 90\% PM MRI completion on four frontal slices, as well as 30\% NM traffic speed (S) imputation at location 10 for one day.}
\label{fig:Component-NM}
\end{figure*}

\paragraph{Nonrandom missingness}
The global and local components under nonrandom missing scenarios, including NM (nonrandom fiber missing along the time-of-day mode) for traffic data, SM (structured missingness) for image data, and PM (patch missingness) for MRI, have been summarized in Figure~\ref{fig:Component-NM}.

\end{document}